\documentclass[twoside]{article}

\usepackage[accepted]{aistats2022}

\usepackage[utf8]{inputenc} 
\usepackage[T1]{fontenc}    
\usepackage{hyperref}       
\usepackage{url}            
\usepackage{booktabs}       
\usepackage{amsfonts}       
\usepackage{nicefrac}       
\usepackage{xcolor}         
\usepackage{amsmath}
\usepackage{dsfont}

\usepackage{macros}
\usepackage{thm-restate}

\usepackage[round, authoryear]{natbib}

\usepackage{hyperref}
\usepackage[algo2e, inoutnumbered, algoruled,vlined]{algorithm2e}
\usepackage{makecell}
\usepackage{pgfplots}
\usepgfplotslibrary{fillbetween}
\usetikzlibrary{patterns}
\pgfplotsset{compat=1.5.1}
\usetikzlibrary{calc}

\newif\ifappendix
\appendixfalse

\begin{document}

%

%

\twocolumn[

\aistatstitle{Efficient Algorithms for Extreme Bandits}

\aistatsauthor{ Dorian Baudry$^1$ \And Yoan Russac$^2$ \And  Emilie Kaufmann$^1$ }

\aistatsaddress{$1.$ CNRS, Univ. Lille, Inria, Centrale Lille, UMR 9189 - CRIStAL, F-59000, Lille, France \\
$2.$ DI ENS, CNRS, ENS, Université PSL, Paris, France} ]

\begin{abstract}
In this paper, we contribute to the Extreme Bandit
problem, a variant of Multi-Armed Bandits
in which the learner seeks to
collect the largest possible reward. We first
study the concentration of
the maximum of i.i.d random variables under
mild assumptions on the tail of the rewards distributions. 
This analysis motivates the introduction of Quantile
of Maxima (QoMax). 
The properties 
of QoMax are sufficient
to build an Explore-Then-Commit (ETC)
strategy, QoMax-ETC, achieving
strong asymptotic guarantees despite its simplicity. 
We then propose and analyze 
a more adaptive, anytime algorithm, QoMax-SDA, 
which combines
QoMax with a subsampling method
recently introduced by \cite{lbsda}.  
Both algorithms are more efficient than existing approaches
in two aspects (1) they lead to better empirical performance
(2) they enjoy a significant reduction 
of the memory and time complexities.
\end{abstract}

\section{INTRODUCTION}
\label{sec::intro}

Multi-Armed Bandits (MAB) provide a powerful framework for
balancing exploration and exploitation in sequential 
decision making tasks. In a MAB model, a learner is interacting with 
$K$ unknown distributions (called arms) generating rewards, that we denote by $\nu_1,\dots,\nu_K$. In the most 
classical problem formulation, the learner sequentially samples the arms in order to
maximize her expected sum of rewards. In this paper, we consider a different 
setting in which the learner seeks 
to collect the largest possible reward. This problem, first introduced by \citet{cicirello2005max}, is 
referred to as \textit{Extreme Bandits} 
or \textit{max K-armed bandit}.
Obtaining the largest possible reward can 
be of interest for practical scenarios including financial
\citep{gilli2006application}, medical
\citep{neill2010multivariate} or online marketing 
\citep{skiera2010analysis} applications.

Letting $X_{k,t}$ be the reward obtained from arm $k$ at time $t$, a bandit algorithm (or policy) selects an arm 
$I_t$ using past observations and receives the reward $X_{I_t,t}$. The rewards stream $(X_{k,t})$ is drawn i.i.d. from $\nu_k$ and independently from other rewards streams. In this work, we assume that all arms have an unbounded support (the finite support case is studied by \cite{nishihara2016no}). In this context, \cite{carpentierExtreme} define the extreme regret of a policy as
\begin{equation}
\label{eq:extreme_regret_main}
\mathcal{R}_T^{\pi} = \max_{k \leq K} \bE[ \max_{t \leq T} X_{k,t} ]
- \bE_\pi[ \max_{t \leq T} X_{I_t, t} ] \;.
\end{equation}
Two types of performance guarantees have been derived in previous works. Using the terminology of \cite{bhatt2021extreme}, we say that $\pi$ has a vanishing regret in the \textit{weak} sense if
\begin{equation}
\label{eq:vanishing_regret_weak_main}
\mathcal{R}_T^{\pi} = \underset{T\rightarrow \infty}{o}\left(\max_{k \leq K} \bE[ \max_{t \leq T} X_{k,t} ]\right)
\end{equation}
and $\pi$ has a vanishing regret in the \textit{strong} sense if
\begin{equation}
\label{eq:vanishing_regret_strong_main}
\lim_{T \rightarrow \infty}
\cR_T^{\pi}  = 0 \;.
\end{equation}
%
While classical bandit algorithms aim for 
the arm with the largest expected reward, 
a good 
 algorithm for extreme 
bandit should intuitively discover the arm with the heaviest tail. 
Existing algorithms for this problem can be divided into 
three categories:
(1) Fully-parametric 
approaches
\citep{cicirello2005max,streeter2006asymptotically} 
where the distributions are assumed 
to be known (Frechet, Gumbel).
(2) Semi-parametric approaches
\citep{carpentierExtreme,achab2017max}
where distributions satisfy a second-order Pareto assumption.
In \cite{carpentierExtreme}, weak vanishing 
regret is obtained for second-order Pareto 
distributions assuming that a lower bound 
on a parameter of the distribution is 
known to the algorithm.
\cite{achab2017max} refine this 
analysis and obtain strong vanishing 
regret when this lower-bound is large enough.
(3) Distribution-free approaches \citep{streeter2006simple,
 bhatt2021extreme} which do not leverage any assumption
on the reward distributions. 
A simple algorithm, ThresholdAscent, was proposed in 
\cite{streeter2006simple}, but without theoretical guarantees. 
\cite{bhatt2021extreme} recently proposed Max-Median, 
an algorithm based on robust statistics that can 
be employed for any kind of distribution. Max-Median is proved to have 
weak vanishing regret for polynomial-like arms and 
strongly vanishing regret for exponential-like arms. 


In this work, we revisit the extreme bandit problem with the idea 
of designing algorithms based on \emph{pairwise comparisons of tails} 
with provable guarantees under minimal assumptions on the arms. 
The motivation stems from a recent line of work on subsampling algorithms for classical bandits \citep{baudry2020sub} which performs ``fair'' pairwise comparisons of empirical means based on an equal sample size and attains good performance for several types of distributions.  

In Section~\ref{sec::max_comp}, 
we highlight the limitation of comparing directly the 
maxima of $n$ i.i.d. samples and introduce 
the \textit{Quantile of Maxima} (QoMax) estimator. 
Instead of computing the maximum of $n$ samples, 
the learner separates the collected data into
\textit{batches} of equal size 
and compute the quantile of order $q$ of the maxima over the different batches. 
QoMax is inspired by 
the Median of Means estimator \citep{alon1999space} 
that was used for heavy-tail bandits
\citep{bubeck_heavy}. We derive upper bounds on the probability that 
one QoMax exceeds another, that are instrumental to design our algorithms. 
In Section~\ref{sec::qomax_etc},  we first propose an Explore-Then-Commit 
algorithm using QoMax, for which we establish vanishing regret in the strong sense under the mild assumption that the bandit model has a dominant arm. Albeit simple, this approach requires some tuning which depends on the horizon $T$. To overcome this limitation, we propose in Section~\ref{sec::qomax_sda} the QoMax-SDA algorithm which combines QoMax with the subsampling strategy from \cite{lbsda}. We prove that it achieves vanishing regret for arms with exponential or polynomial tails and also provide some elements of analysis under the weaker dominant arm assumption. In Section~\ref{sec::xp}, we highlight the efficiency of our algorithms which allow for a significant reduction of the storage and computational cost while outperforming  existing approaches empirically.


\section{COMPARING TAILS}\label{sec::max_comp}

In this section, we motivate our new QoMax estimator 
used for comparing 
the tails of two distributions based on $n$ i.i.d.\ samples of each. We first present the assumptions under which we are able to analyze QoMax and the resulting extreme bandit algorithms. \\

We define the \textit{survival function} 
$G$ of a distribution $\nu$  as $G(x)=\bP_{X\sim \nu}(X > x)$ for all $x\in \R$.
We shall consider two different assumptions for arms' distributions.

\begin{definition}[Exponential or polynomial tails]\label{def::tails}
Let $\nu$ be a distribution of 
survival function $G$. \textbf{(1)} If there exists $C>0$ and $\lambda > 1$ such that $G(x) \sim Cx^{-\lambda}$ we say that $\nu$ has a \textbf{polynomial tail}.
\textbf{(2)} If there exists $C>0, \lambda \in \R^+$ such that $G(x) \sim C \exp(-\lambda x)$ we say that $\nu$ has an \textbf{exponential tail}.
\end{definition}

These \textit{semi-parametric} assumptions (which says nothing about the lower part of the distribution) have been introduced by \cite{bhatt2021extreme}.
We remark that a polynomial tail is a weaker condition than the second-order Pareto assumption from \cite{carpentierExtreme}. Now, we introduce a general assumption which allows to compare two (arbitrary) tails.
\begin{definition}[Dominating tail]
\label{def:asympt_dominating}
Let $G_1$ and $G_2$ be the 
survival functions of two distributions $\nu_1$ and $\nu_2$.
We say that the tail of $\nu_1$ \textbf{dominates} the tail of $\nu_2$ (we write $\nu_1 \succ \nu_2$ 
)
if there exists $C>1$ and
$x \in \R$ such that for all $y > x$, 
$
G_1(y) > C G_2(y).$
\end{definition}

In the rest of the paper, we will consider a bandit model that has a dominating arm, denoted by 1 without loss of generality: $\nu_1 \succ \nu_k$ for all $k \neq 1$. Under this assumption, arm 1 is optimal in the sense that for $T$ large enough 
an oracle strategy would select this arm only. To the best of our knowledge, this is the weakest assumption introduced so far for extreme bandits.

\subsection{Comparing Maxima}\label{sec::prop_max}

Let $\nu_1$ and $\nu_2$ be two distributions from which we observe $n$ i.i.d.\ samples denoted by $X_{1,1},\dotsc,X_{1,n}$ and $X_{2,1},\dots,X_{2,n}$ respectively. 
A natural idea to compare their tails is to use the samples' maxima,
$X_{k,n}^+ = \max \{X_{k,1}, \dots, X_{k,n} \}$
for $k \in \{1,2\}$.  
For these estimators to serve as a proxy for comparing the tails, we need the 
probability $\bP(X_{1,n}^+ < X_{2,n}^+)$ to decay fast enough when $\nu_1 \succ \nu_2$. To upper bound this probability, we note that for any sequence $(x_n)$,
\[\bP(X_{1,n}^+ < X_{2,n}^+) \leq \bP(X_{1,n}^+ \leq x_n) + \bP(X_{2,n}^+ > x_n) \;.\]
Using first that $\bP(X_{1,n}^+ \leq x) \leq \exp(-n G_1(x))$ and then $\bP(X_{2,n}^+ > x) \leq n G_2(x)$,  and optimizing for $x_n$ yields the following result, proved in Appendix~\ref{app::prop_maximax}.

\begin{restatable}[Comparison of Maxima]{lemma}{compmaxima}
\label{lem::max_comp}
Assume that both $\nu_1$ and $\nu_2$ have either polynomial or exponential tails, with respective second parameter $\lambda_1$ and $\lambda_2$, with $\lambda_1<\lambda_2$ (so that $\nu_1\succ \nu_2$). Define $\delta = \frac{\lambda_2}{\lambda_1} - 1 > 0$, then there exists a sequence $(x_n)$ such that 
\[
\max\{\bP(X_{1,n}^+\leq x_n), \bP(X_{2,n}^+ \geq x_n)\}
= \cO \left(\frac{(\log n)^{\delta+1}}{n^\delta}\right).
\]
\end{restatable}

Lemma~\ref{lem::max_comp} shows that even under the stronger semi-parametric assumption, $\bP(X_{1,n}^+\leq X_{2,n}^+)$ does not decay exponentially fast, unlike what happens when we compare the empirical means of light-tailed distributions. Furthermore, the rate $\delta$ is problem-dependent and can be arbitrarily small. As pointed out by \cite{carpentierExtreme} it can actually be seen as the Extreme Bandits equivalent of the \textit{gap} in bandits, we therefore call $\delta$ the \textbf{\textit{tail gap}}. Besides, we prove in Lemma~\ref{lem::prob_lower} (in Appendix) a lower bound of order $\cO(n^{-(1+\delta)})$, which motivates using more robust statistics on the distributions' tails.

\subsection{Quantile of Maxima (QoMax)}

Results similar to those of Section~\ref{sec::prop_max} have been previously encountered in the bandit literature. In \citep{bubeck_heavy}, the authors study the problem of bandit with heavy tails, prove a concentration inequality in $n^{-\delta}$ for some $\delta>0$  and use this result to build several estimators with faster convergence. 
Among them, they consider
the Median-of-Means (MoM) introduced by \cite{alon1999space}. 
We build a natural variant of MoM, that we call Quantile of Maxima (QoMax). The principle of QoMax is simple: the learner chooses a quantile $q$, and has access to $N = b \times n$ data $\mathcal{X}= (X_{m,i})_{m\leq n, i \leq b}$. It then allocates the data in $b$ batches of size $n$ and: (1) find the maximum of each batch, (2) compute the quantile of order $q$ over the $b$ maxima. We summarize QoMax in Algorithm~\ref{alg::qomax}.

\begin{algorithm}[hbtp]
	\SetKwInput{KwData}{Input}
	\KwData{quantile $q$, $b$ batches of size $n$, observations $(X_{m,i})_{m \leq n, i \leq b}$}
	\For{$i=1, \dotsc, b$}{Compute $(X_{n}^+)^{(i)}=\max \{X_{1,i}, \dots, X_{n,i} \}$}
	\SetKwInput{KwResult}{Return}
	\KwResult{quantile of order $q$ of $\{(X_{n}^+)^{(1)}, \dots, (X_{n}^+)^{(b)}\}$}
	\caption{Quantile of Maxima (QoMax)}
	\label{alg::qomax}
\end{algorithm}


For a finite set of size $b$, we simply define the quantile $q$ as the observation of rank $\lceil bq \rceil$ in the list of sorted data (in increasing order). In the sequel we denote by $\bar{X}^{q}_{k,n,b}$ the QoMax of order $q$ computed from $b$ batches of size $n$ of i.i.d. replications from arm $k$. 

We are now ready to state the crucial property of QoMax estimators that will be used in our two analyses.

\begin{restatable}[Comparison of QoMax]{theorem}{compqomax}
\label{th::qomax_comp} 
Let $\nu_1$ and $\nu_2$ be two distributions satisfying $\nu_1 \succ \nu_2$ and $q \in (0,1)$. 
Then, \textbf{there exists} a sequence $x_n$, a constant $c > 0$, and an integer $n_{\nu_1, \nu_2, q}$ such that for $n\geq n_{\nu_1, \nu_2, q}$,
	\[
	\max\left\{\bP(\bar X_{1,n, b}^q \leq x_n), \bP(\bar X_{2,n, b}^q \geq x_n)\right\} \leq \exp(-cb) \;. 
	\]
If the tails are furthermore either polynomial or exponential with a \textbf{positive tail gap}, then the result holds \textbf{for any} $c > 0$ and $n$ larger than some $n_{c, \nu_1, \nu_2,q}$.
\end{restatable}

It follows from Theorem~\ref{th::qomax_comp} that $\bP(\bar X_{1,n, b}^q \leq \bar X_{2,n,b}^q) \leq 2\exp(-cb)$ for $n$ large enough. Strikingly, this result tells us that, under the simple assumption that one tail dominates the comparison of QoMax computed with the same parameters will not be in favor of the dominating arm with a probability that \textbf{decreases exponentially with the batch size}. 
\begin{remark}
In general QoMax is \textbf{not} an estimate of the expectation of the maximum. We will use it to \textbf{compare two tails}, in order to find the heavier.
\end{remark}

\begin{remark}[Choice of quantile level $q$] Note that Theorem~\ref{th::qomax_comp} holds for any value of $q\in (0,1)$, but the impact of $q$ is materialized in the (problem-dependent) sample size $n_{\nu_1,\nu_2,q}$ needed for the inequality to hold. 
For the practitioner, we think that in most cases choosing $q=1/2$ is appropriate.
Still, in Section~\ref{sec::xp} we exhibit a difficult setting where a choice of $q$ close to $1$ is helpful. 
\end{remark}

\subsection{Proof of Theorem~\ref{th::qomax_comp}} 

We let $\kl(x, y)=x\log(x/y) + (1-x)\log((1-x)/(1-y))$ denote the binary relative entropy. Just like for the analysis of Median-of-Means, the starting point is to relate deviations inequalities for a QoMax to deviation inequalities for binomial distributions. Letting $(X_{1,n}^{+})^{(i)}$ (resp. $(X_{2,n}^{+})^{(i)}$) denote the maximum over the $i$-th batch of observations from $\nu_1$ (resp. $\nu_2$), 

\begin{align*}
\bP(\bar X_{1,n, b}^q \leq x) &\leq 
\bP\left(\sum_{i=1}^b \ind\!\left((X_{1,n}^{+})^{(i)} \leq x\right) \geq bq  \right) \\
& \leq \exp(-b \times \kl(q, \bP(X_{1,n}^{+} \leq x))) \;.
\end{align*}

The last step applies the Chernoff inequality to a binomial distribution with parameters $b$ and $p = \bP(X_{1,n}^{+} \leq x)$, and holds whenever $\bP(X_{1,n}^{+} \leq x) \leq q$. Similarly, if $\bP(X_{2,n}^{+} \geq x) \leq 1-q-1/b$, we have

\begin{align*}
\bP(\bar X_{2,n, b}^q \geq x) &\leq \bP\left(\sum_{i=1}^b \!\ind \!\left((X_{2,n}^{+})^{(i)} \geq x\right) \! \geq \! b-bq-1  \!\right) \\
& \leq \exp(-b  \kl(1-\hspace{-0.02cm}q-1/b, \bP(X_{2,n}^+ \geq x)))
\end{align*}
For exponential and polynomial tails, thanks to Lemma~\ref{lem::max_comp} there exists a sequence $(x_n)$ such that both $\bP(X_{1,n}^+\leq x_n)$ and $\bP(X_{2,n}^+ \geq x_n)$ converge to zero, and the result follows easily. 
Under the dominance assumption, the following result controls the deviations of the maxima and is proved in Appendix~\ref{app::prop_maximax}.

\begin{restatable}{lemma}{lemmsecond}
\label{lem::max_general} Assume that $\nu_1\succ \nu_2$. Then, for any $q\in (0, 1)$ there exists $n_{\nu_1, \nu_2, q}\in \N$, a sequence $x_n$ and some $\epsilon>0$ such that for all $n \geq n_{\nu_1, \nu_2,q}$ and $b$ large enough,
\[\bP(X_{1,n}^+\leq x_n) \leq q-\epsilon \;,\text{ and} \quad \bP(X_{2,n}^+ \leq x_n)\geq q+\epsilon\;.\]
\end{restatable}

With the notation of Lemma~\ref{lem::max_general}, Theorem~\ref{th::qomax_comp} then holds for $c = \min\left(\kl(q,q-\epsilon),\kl(1-q-\varepsilon/2,1-q-\varepsilon)\right)$ provided that the batch size is larger than $2/\varepsilon$.


%
\section{QoMax-ETC}
\label{sec::qomax_etc}

In this section, we propose QoMax-ETC, a simple Explore-Then-Commit algorithm using QoMax estimators.
The algorithm is reported in Algorithm~\ref{alg:qomax-etc} and works as follows. First, the learner selects a quantile $q$, and given the time horizon $T$ picks a batch size $b_T$ and a sample size $n_T$. Then, the exploration phase starts where every arm is pulled $N_T = b_T \times n_T$ times allocated in $b_T$ batches of size $n_T$. At the end of this step, the learner computes a $q$-QoMax estimator from the history of each arm using the different batches. Next comes the exploitation phase where the algorithm pulls the arm $I_T$ with the largest QoMax until time $T$.

\begin{algorithm}[hbtp]
\SetKwInput{KwData}{Input}
\KwData{$K$ arms, horizon $T$, quantile $q$, number of batches $b_T$, number of samples per batch $n_T$}
\For{$k=1, \dots, K$}{
\hspace{0.08cm} Pull arm $k$, $b_T \times n_T$ times \\
Allocate the data in $b_T$ batches of size $n_T$ \\
Compute their QoMax, $\bar X_{k,n_T, b_T}^q$ (Algorithm~\ref{alg::qomax})}
\For{$t=K \times n_T \times b_T +1, \dots, T$}{
\hspace{0.08cm} Pull arm $I_T = \aargmax_k 
\bar X_{k,n_T, b_T}^q$}
\caption{QoMax-ETC}
\label{alg:qomax-etc}
\end{algorithm}

We remark that an ETC algorithm has already been proposed by \cite{achab2017max} for extreme bandits. Their algorithm differs from ours by the choice of the arm $I_T$ drawn in the exploitation phase: they build an upper confidence bound on the maximum under the assumption that the distributions are second-order Pareto and select $I_T$ as the arm with largest upper confidence bound. In contrast, QoMax-ETC does not assume anything about the arms distributions.


We now analyze QoMax-ETC under a bandit model $\nu = (\nu_1,\dots,\nu_K)$ such that $\nu_1 \succ \nu_k$ for all $k\neq 1$. 

\begin{restatable}[Regret of QoMax-ETC]{proposition}{regretetc}
\label{prop::regret_ETC_main}
	Let $\pi$ be an ETC policy sampling $N_T = n_T\times b_T$ times each arm during the exploration phase. 
	 If $T\geq KN_T$, 
	\ifappendix
	\begin{align*}
	\cR_{T}^\pi 
	& \leq \underbrace{\bE\left[\max_{t \leq T} X_{1,t} \right]  -
	\bE\left[\max_{t \leq T-KN_T} X_{1,t}
	\right]}_{\text{Exploration cost}}
	+
	\underbrace{\bP(I_T\neq 1)\bE\left[\max_{t
	\leq T} X_{1,t} \right]}_{\text{Cost of picking a wrong
	arm}} \;.
	\end{align*}
	\else
	\begin{align*}
	\cR_{T}^\pi 
	& \leq \underbrace{\bE\left[\max_{t \leq T} X_{1,t} \right]  -
	\bE\left[\max_{t \leq T-KN_T} X_{1,t}
	\right]}_{\text{Exploration cost}}
	\\
	& \quad
	+ \;
	\underbrace{\bP(I_T\neq 1)\bE\left[\max_{t
	\leq T} X_{1,t} \right]}_{\text{Cost of picking a wrong
	arm}} \;.
	\end{align*}
	\fi
\end{restatable}

We prove this result in Appendix~\ref{app::proof_etc}. This proposition
shows that the regret of the ETC algorithm can be properly controlled by two factors \textbf{(1)} the probability of picking a wrong arm for the exploitation phase, \textbf{(2)} the gap between the growth rate of the maximum over $T$ or $T-KN_T$ observations of the dominant arm, that we call "exploration cost" as it is fully determined by the length of the exploration phase and the arms' distributions. In the rest of the paper we will assume that the distribution of the dominant arm satisfies the following assumption.

\begin{restatable}{assumption}{assespmax}
\label{ass:esp_max}
\ifappendix
$\bE\left[ X_T^+ \right]=o(T)$, and for any $\gamma<1$ if $N_T = o(T^\gamma)$ then
\[
	\bE\left[X_T^+ \right]  -
		\bE\left[X_{T-N_T}^+
		\right] \xrightarrow[T \to +\infty]{} 0\;. 
		\]
\else
$\bE\left[\max_{t \leq T} X_{1,t} \right]=o(T)$, and for any $\gamma<1$ if $N_T = o(T^\gamma)$ then
\[
	\bE\left[\max_{t \leq T} X_{1,t} \right]  -
		\bE\left[\max_{t \leq T-KN_T} X_{1,t}
		\right] \xrightarrow[T \to +\infty]{} 0\;. 
		\]
\fi
\end{restatable}
This condition is satisfied for nearly all distributions encountered in practice (e.g polynomial, exponential or gaussian tails) as discussed in Appendix~\ref{app::prop_maximax}, in which we provide explicit upper bounds on the exploration cost. \black
We now state our main theoretical claim for QoMax-ETC.

\begin{restatable}[Vanishing regret of QoMax-ETC]{theorem}{vanishingqomax}
\label{th::vanishing_qomax_etc}
Consider a bandit $\nu=(\nu_1, \dots, \nu_K)$ with $\nu_1 \succ \nu_k$ for $k\neq 1$. 
Under Assumption~\ref{ass:esp_max}, for any quantile $q \in (0,1)$ and any sequence $(b_T, n_T)$ satisfying 
\[
\frac{b_T}{\log(T)} \rightarrow + \infty \text{ and } n_T \rightarrow +\infty\;, 
\] 
the regret of QoMax-ETC with parameters $(q, b_T, n_T)$ is \textbf{vanishing in the strong sense}. Furthermore, for polynomial/exponential tails with positive tail gaps this result also holds for $b_T = \Omega(\log T)$.
\end{restatable}

\begin{proof} From Theorem~\ref{th::qomax_comp}, there exists constants $c_k$ for $k\geq 2$ such that for $T$ large enough (such that $n_T$ becomes larger than $n_{\nu_1,\nu_k,q}$), it holds that
\[
\bP(I_T \neq 1) \leq \sum_{k =2}^{K} \bP(\bar X^{q}_{k,n_T,b_T} >\bar X^{q}_{1,n_T,b_T}) \leq \sum_{k=2}^{K}e^{-c_k b_T}
\]
It follows that $\bP(I_T \neq 1) = o(T^{-1})$ if $b_T/\log(T)\rightarrow \infty$ and we conclude with Proposition~\ref{prop::regret_ETC_main} and Assumption~\ref{ass:esp_max}. For polynomial or exponential tails, as the above inequality holds for any value of $c_k$, $b_T = \Omega (\log T)$ is sufficient to obtain $\bP(I_T \neq 1) = o(T^{-1})$.
\end{proof}


Even if Theorem~\ref{th::vanishing_qomax_etc} is stated in an asymptotic way, we emphasize that its proof provides a finite-time upper bound on the probability of picking a wrong arm, $\bP(I_T\neq 1)$, that is valid provided that $T$ is larger than some (problem-dependent) constant. In particular, $T$ needs to be large enough so that $n_T \geq \max_{k \neq 1} n_{\nu_1, \nu_k, q}$ where $n_{\nu_1, \nu_k, q}$ is the number of samples need in Theorem 1 for the concentration of QoMax. This number is not always large. For example if we have two Pareto distributions with parameters $\lambda_1 = 1.5$ and $\lambda_2 = 2$, $n_T=3$ is enough. Using our regret decomposition, this result would lead to a finite-time upper bound on the extremal regret for distributions for which a finite-time bound on the exploration cost is available.

For satisfying the theoretical requirements while obtaining good empirical performance, we recommend using $b_T =  (\log(T))^2$ and $n_T = \log(T)$ when running the algorithm. All the experiments reported in Section~\ref{sec::xp} use these values. 
QoMax-ETC is computationally appealing and has strong asymptotic guarantees. However in practice we found that its performance can vary significantly depending 
on the choices of $b_T$ and $n_T$, which should in particular use a reasonable guess for the horizon $T$.  
For this reason, in the next section we propose QoMax-SDA, which is still based on QoMax comparisons but is anytime (i.e. independent on $T$) and requires less parameter tuning.


%
%
\section{QoMax-SDA}
\label{sec::qomax_sda}

In this section we present QoMax-SDA, an algorithm using a subsampling mechanism based on LB-SDA \citep{lbsda}. We detail the key principles of the algorithm and propose a theoretical analysis.

\subsection{Algorithm and Implementation}

From a high level QoMax-SDA follows the structure of the subsampling duelling algorithms introduced in \cite{baudry2020sub}. The algorithm operates in successive rounds composed of (1) the selection of a leader, (2) the different duels between the leader and the challengers and (3) a data collection phase. We develop each of those steps in the sequel.

At the beginning of a round $r$, the learner has access to the history of the different arms denoted $\cX_k^r$. For the needs of the QoMax, the collected rewards for arm $k$ are gathered within $b_k(r)$ batches of equal size $n_k(r)$ such that $|\cX_k^r| = b_k(r) n_k(r)$. $n_k(r)$ is called the \textit{number of queries} and corresponds to the number of times the arm $k$ has been selected by the learner at the end of round $r$. The leader at round $r$, denoted by $\ell(r)$, is the arm that has been queried the most up to round $r$. The $K-1$ remaining arms are called \textit{challengers}. In case of equality, ties are broken according to any fixed rule (e.g at random). Formally, $\ell(r) = \aargmax_{k \leq K} \; n_k(r)$.

Once the leader is selected, $K-1$ duels with the different challengers are performed. We denote $\cA_{r+1}$ the set of arms that will be pulled at the end of round $r$. An arm $k$ is added to $\cA_{r+1}$ in two cases (1) if it wins its duel or (2) if its number of queries is too small: $n_k(r) \leq f(r)$ for a fixed function $f(r)$ representing the \textit{sampling obligation}. If no challenger is added to $\cA_{r+1}$ the leader is pulled. We now detail the duel procedure that is reported in Algorithm~\ref{alg::duel_sda}. We assume that an infinite stream of rewards is available for each arm, in the form of an array with an infinite number of rows and columns, so that we denote the rewards of arm $k$ by $(X_{k,n,b})_{n \in \mathbb{N}, b \in \mathbb{N}}$, where $X_{k,n,b}$ corresponds to the $n$-th sample of $b$-th batch from arm $k$. We further assume that the number of batches available for an arm $k$ depends only on its number of queries $n_k(r)$ so that $b_k(r) = \lceil B(n_k(r))\rceil$ for some function $B$. The duel is a comparison of the QoMax of the challenger using its entire history and the QoMax of the leader on a subsample of its history.
\begin{algorithm}[hbtp]
	\SetKwInput{KwData}{Input}
	\KwData{$q$, arm $k$, leader $\ell$, current history, batch count and batch size: $(\cX_m, b_m, n_m)$ for $m \in \{k, \ell\}$}
	\SetKwInput{KwProg}{QoMax computation}
\KwProg{
\begin{enumerate}
\item  Compute $I_k=\text{QoMax}(q, b_k, n_k, \cX_k)$ (Alg.~\ref{alg::qomax})\\
\item Collect the \textbf{subsample} $\cY_\ell=(X_{\ell, i,j})_{i \in \cN, j \in \cB} \subset \cX_\ell$ for $\cN = [n_\ell-n_k+1, n_\ell]$ and $\cB = [1, b_k]$.
\item Compute $I_\ell = \text{QoMax}(q, b_k, n_k, \cY_\ell) $ (Alg.~\ref{alg::qomax})
\end{enumerate}}
\SetKwInput{KwResult}{Return}
\KwResult{$\aargmax_{m \in \{k, \ell\}} I_m$}
	\caption{Duel ($q$-QoMax comparison)}
	\label{alg::duel_sda}
\end{algorithm}

Our subsampling mechanism is inspired by LB-SDA and works as follows. When comparing the leader $\ell(r)$ with a challenger $k$: (1) we only consider the rewards collected from the $n_k(r)$ \textbf{last} \textit{queries} of arm $\ell(r)$ (as in LB-SDA), and (2) we only keep the $b_k(r)$ \textbf{first} batches
for $\ell(r)$. This way, the QoMax from the leader and the challenger is computed using the same amount of data. Taking the last queries introduces some diversity in the subsamples encountered when $\ell$ is often pulled (we refer to \cite{lbsda} for details) and using the first batches allows for a reduction of the storage need (see Implementation tricks).

\begin{figure}[hbt]
	\centering
	\includegraphics[width=0.47\textwidth]{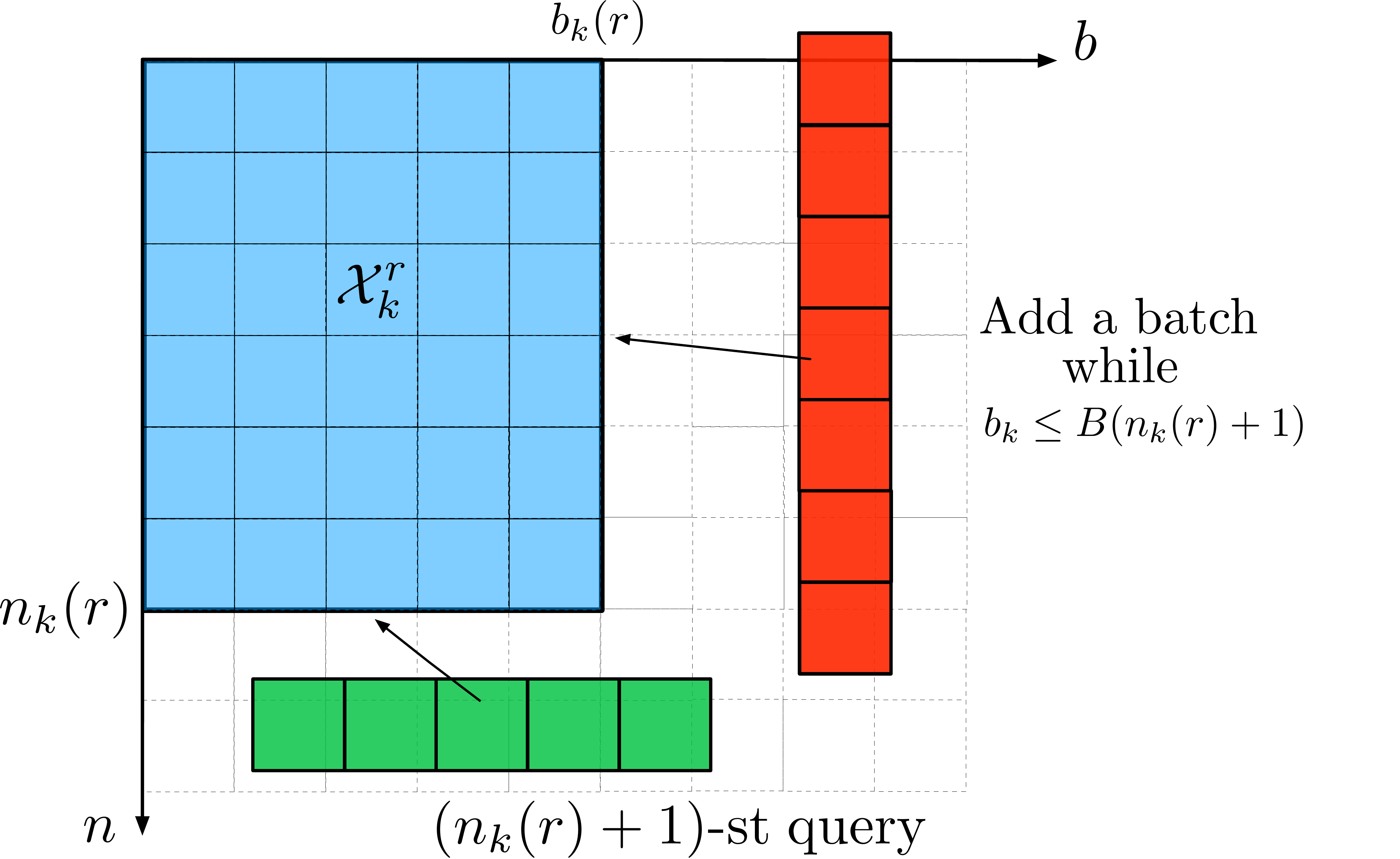}
	\caption{Illustration of the CollectData procedure at round $r$ for a challenger $k \in \mathcal{A}_{r+1}$.}
	\label{fig::collectdata}
\end{figure}

We now detail the data collection procedure that is used by QoMax-SDA and illustrated on Figure~\ref{fig::collectdata}.
If we query arm $k$ with parameters $(\cX_k, n_k, b_k, B, \ell)$ at round $r$,
(1) we update existing batches: collect a $(n_k+1)$-st query for all existing batches
$(X_{k,n_{k}+1},b)_{b \leq b_k}$. (2) Create new batches: \textit{while} $b_k < B(n_k+1)$, collect the $n_k+1$ rewards $(X_{k,n,b})_{n \leq n_k+1,b_k < b \leq B(n_k +1)}$.

Combining all those elements gives QoMax-SDA reported in Algorithm~\ref{alg:qomax-sda}. 

\begin{algorithm}[hbtp]
	\SetKwInput{KwData}{Input}
	\KwData{$K$ arms, quantile level $q$ \\  exploration function $f$, batch function $B$}
	\SetKwInput{KwResult}{Initialization}
	\KwResult{$r\leftarrow 0$ \\
		$\forall k \in \{1, ..., K \}$:  $n_k \leftarrow 0$, $b_k\leftarrow0$, $\cX_k\leftarrow $\text{emptyarray2d}}
	\For{$r \geq 1$}{
		$r \leftarrow r +1$,
		$\mathcal{A} \leftarrow \{\}$,  $\ell \leftarrow \text{leader}(n_k, \cX_k)$ \\
		\If{$r=1$}{
			$\cA \leftarrow \{1, \dots, K\}$ (Draw each arm once)}
		\Else{
			\For{$k \neq \ell \in \{1,...,K\}$}{
				\If{$n_{k} < f(r)$ or \emph{Duel}$(k, \ell)=k$}{$\mathcal{A} \leftarrow \mathcal{A} \cup \{ k \}$}
				\If{$| \mathcal{A} | = 0$}{$\mathcal{A} \leftarrow \{l \}$}
			}
		}
		\For{$k \in \mathcal{A}$}{
	   CollectData$(\cX_k, n_k, b_k, B, \ell)$ , update $\cX_k$, $b_k$\\
		$n_k \leftarrow n_k+1$
	}
}
	\caption{QoMax-SDA}
	\label{alg:qomax-sda}
\end{algorithm}

\paragraph{Implementation tricks}
Our algorithm can enjoy a significant reduction of storage with two different tricks. (1) An efficient storing of the maxima: for arm $k$ in the batch $b$ every time a new sample $x$ is collected, all stored values smaller than $x$ (if any) are deleted.
The new sample $x$ and the round where $x$ was received are then stored.
(2) An efficient CollectData procedure. We could use the same procedure for all the arms and obtain a number of batch for the leader that scales as $n^\gamma$.
If the algorithm ends up pulling an arm most of the time (which is expected), this will create new batches for the leader that are never used in the duels because
with our subsampling mechanism, only the first $b_k$ batches are used when the leader competes with arm $k$. Instead, the CollectData procedure is only applied to the challengers (see Algorithm~\ref{alg:collectdatasda}) and a batch is added to the leader only when it has to match the number of batches of the second most pulled arm. Those tricks are detailed in Appendix~\ref{sec:implem_tricks_app}.


Note that a sampling obligation, through the exploration function $f$ (independent on $T$), is necessary under general assumptions as in all existing algorithms.



\subsection{Extreme Regret Analysis}
We now provide an analysis of QoMax-SDA under the same assumption as before: $\nu_1 \succ \nu_k$ for all $k\neq 1$.  Let $N_k(t)$ denote the number of pulls of arm $k$ at time $t$. We start with a generic regret decomposition.
\begin{restatable}[Regret decomposition with a low probability event]{proposition}{regretsda}
\label{prop::regret_decompo_SDA_main}
Define the event
\[\xi_T := \{N_1(T) \leq T-KM_T\} \;,\]
where $(M_T)_{T \in \N}$ is a fixed sequence. Then, for $T\geq KM_T$, \black 
for any constant
$x_T \in \mathbb{R}$,  it holds that,
\ifappendix
\begin{align*}
\mathcal{R}_T^\pi &\leq \underbrace{\bE\left[\max_{t \leq T} X_{1,t}\right] -
\bE\left[\max_{t \leq T- K M_T} X_{1,t} \right]}_{\text{Exploration cost}}
+ \underbrace{x_T \bP(\xi_T)
+ \bE\left[\max_{t\leq T} X_{1,t}
\ind \left(\max_{t\leq T} X_{1,t} \geq x_T \right)\right]}_{\text{Cost
incurred by } \xi_T} \;.
\end{align*}
\else
\begin{align*}
\mathcal{R}_T^\pi &\leq \underbrace{\bE\left[\max_{t \leq T} X_{1,t}\right] -
\bE\left[\max_{t \leq T- K M_T} X_{1,t} \right]}_{\text{Exploration cost}}
\\
&
+ \underbrace{x_T \bP(\xi_T)
+ \bE\left[\max_{t\leq T} X_{1,t}
\ind\left(\max_{t\leq T} X_{1,t} \geq x_T\right)\right]}_{\text{Cost
incurred by } \xi_T} \;.
\end{align*}
\fi
\end{restatable}

The proof of this result follows the analysis from \cite{carpentierExtreme} and is given in Appendix~\ref{app::proof_sda}, which contains the proofs of all results from this section.
The ``cost incurred by $\xi_T$'' features two terms. Interestingly, only the first term depends on the algorithm. We upper bound it below.


\begin{restatable}[Upper bound on $\bP(\xi_T)$]{lemma}{lemmaprobxisda}
\label{lem::prob_xi_sda}
For any $q \in (0,1)$, any $M_T$ and any $\gamma>0$, under QoMax-SDA with parameters $B(n)=n^{\gamma}$ and $f(r)=(\log r)^\frac{1}{\gamma}$,
	\[
	\bP(\xi_T) = \cO\left(
	(\log T)^{\frac{1}{\gamma}} M_T^{-\frac{1}{1+\gamma}}
	\right)\;.
	\]
Moreover, for all $k\neq 1$, $\bE[n_k(T)] = \mathcal{O}((\log T)^{1/\gamma})$.
\end{restatable}

\begin{proof}[Sketch of proof]
We first prove that $\bP(\xi_T)$ is upper bounded by $\sum_{k=2}^K \bP(N_k(T)\geq M_T)$. Using that $N_k(T)=b_k(T)\times n_k(T)=n_k(T)^{1+\gamma}$ and Markov inequality we obtain
\[\bP(\xi_T) \leq M_T^{-\frac{1}{1+\gamma}} \sum_{k=2}^K \bE[n_k(T)] \;.\]
It remains to study the expected number of queries of sub-optimal arms $k \geq 2$. This can be done following the outline of \cite{lbsda} and using the deviation inequalities from Theorem~\ref{th::qomax_comp}. 
\end{proof}

The second term in the ``cost incurred by $\xi_T$'' only depends on the distribution of the optimal arm and can be further upper bounded assuming exponential and polynomial tails, leading to the following result.

\begin{restatable}[Upper bound on the regret of QoMax-SDA]{theorem}{thmregretsda}
\label{th::vanishing_qomax_sda}
For any quantile $q$, any $\gamma >0$, defining the parameters of QoMax-SDA as $B(n) = n^\gamma$ and $f(r) = (\log r)^{\frac{1}{\gamma}}$.
\ifappendix

\begin{enumerate}
\item The regret of QoMax-SDA is vanishing in the strong sense for exponential tails
\item The regret of QoMax-SDA is vanishing in the weak sense for polynomial tails.
\end{enumerate}
\else
The regret of QoMax-SDA is \textbf{(1)} vanishing in the strong sense for exponential tails
\textbf{(2)} vanishing in the weak sense for polynomial tails.
\fi
\end{restatable}

\begin{proof}[Sketch of proof]
For parametric tails, we can calculate the growth rate of $\bE[\max_{t \leq T} X_{1,t}]$ with respect to $T$.
This permits to tune the values of $M_T$ and $x_T$ to properly balance the terms in the regret decomposition.
The difference in the convergence for \textbf{(1)} and \textbf{(2)} comes from the fact that the exploration cost scales logarithmically with the time horizon when using exponential tails, whereas the dependency is polynomial with polynomial tails.
\end{proof}

We note that there is no hope to upper bound the last term in our current regret decomposition assuming only that arm 1 dominates the others, so we could not establish vanishing regret for QoMax-SDA under this assumption. As can be seen in the proof of Proposition~\ref{prop::regret_decompo_SDA_main}, the ``cost incurred by $\xi_T$'' is actually an upper bound on $\bE\left[\ind(\xi_T)\max_{t \leq T} X_{1,t}\right]$. If this term were upper bounded by $\bP(\xi_T)\bE\left[\max_{t \leq T} X_{1,t}\right]$\footnote{This is intuitively true as under $\xi_T$, arm 1 underperforms, hence $\ind(\xi_T)$ and $X_{1,T}^+$ are expected to be negatively correlated}, we would get a regret decomposition closer to that in Proposition~\ref{prop::regret_decompo_SDA_main}, leading to a strongly vanishing regret for QoMax-SDA using Lemma~\ref{lem::prob_xi_sda}. Even if we were not able to prove this, we note that (1) QoMax-SDA achieves state-of-the-art performance for exponential and polynomial tails (2) Lemma~\ref{lem::prob_xi_sda} provides a strong indicator of the good performance of QoMax-SDA under more general assumptions, as it shows that the algorithm queries each sub-optimal arm $\mathcal{O}((\log T)^\frac{1}{\gamma})$ times.

We now turn our attention to the practical benefits of using our QoMax-based algorithms.

\section{PRACTICAL PERFORMANCE}
\label{sec::xp}

In all of our experiments, we compare QoMax-SDA and QoMax-ETC with ThresholdAscent \citep{streeter2006simple}, ExtremeHunter \citep{carpentierExtreme}, ExtremeETC \citep{achab2017max} and MaxMedian \citep{bhatt2021extreme}.
We use the parameters suggested in the original
papers (see Appendix~\ref{app::add_xp} for details and remarks
on the tuning).
Namely, $b=1$ for ExtremeHunter/ETC, $s=100, \delta=0.1$ for
ThresholdAscent, $\epsilon_t=(t+1)^{-1}$ for MaxMedian.
For QoMax-ETC, we use $b_T= (\log T)^2$ batches of $n_T=\log T$
samples. This matches the size of the exploration phase
of ExtremeETC and allows for a fair comparison.
For QoMax-SDA, we choose $\gamma=2/3$, which seems to work well across all examples. All the results presented in this section are obtained with these values.


\subsection{Time and Memory Complexity}

We summarize in Table~\ref{tab::costs} the storage and computational time required by the different adaptive and ETC algorithms that we consider, with the aforementioned parameters. The smallest values in each category are colored in blue.
We do not include ThresholdAscent in the table because the comparison is unfair, as it uses a fixed number of data but is not theoretically grounded. We refer the reader to \cite{bhatt2021extreme} for the complexities of the baselines, and we give a few insights on how we obtained the results for QoMax algorithms
(details can be found in Appendix~\ref{app::storage_comp}). 

For QoMax-ETC, the memory needed is $K b_T$ and the time complexity is in $\cO(\max(n_T \;b_T, b_T\log b_T))$ due to the collection phase and the quantile computation. Plugging the values of $b_T$ and $n_T$ gives the result. The time complexity of QoMax-SDA is in $\cO(KT\log T)$ as its main cost consists in sorting data online, just like MaxMedian. The storage of QoMax-SDA is obtained thanks to the two tricks: one allows to keep $\cO(\log T)$ batches, the other $\cO(\log T)$ samples per batch \textit{for the leader}. On the contrary, the complexity for the challengers remains in
$\mathcal{O}(\log T \log\log T)$,
therefore the dependency in $K$ only appears as a second order term.


\begin{table}[h]
	\caption{Average time and storage complexities
		of Extreme Bandit algorithms for a time horizon $T$.
	}
	\label{tab::costs}
	\begin{center}
		\begin{small}
			\begin{tabular}{ccc}
				\toprule
				Algorithm & Memory & Time\\
				\midrule
				Extreme Hunter 
				& $T$ & {\color{red} $\cO(T^2)$} \\
				\midrule
				MaxMedian
				& $T$ & \textcolor{blue}{$\cO(K T \log T)$} \\
				\midrule
				\textbf{QoMax-SDA} &
				$\textcolor{blue}{\cO((\log T)^2)}$
				& \textcolor{blue}{$\cO(KT\log T)$} \\
				\midrule
				\midrule
				Extreme ETC
				& \textcolor{blue}{$\cO\left(K(\log T)^{3}\right)$} & $\cO\left(K(\log T)^{6}\right)$\\
				\midrule
				\textbf{QoMax-ETC} &
				{\color{blue}\textbf{$\cO(K (\log T)^2)$}}
				& {\color{blue}$\cO(K (\log T)^3)$}  \\
				\bottomrule
			\end{tabular}
		\end{small}
	\end{center}
\end{table}

QoMax-SDA offers an exponential reduction of the storage cost compared to ExtremeHunter and MaxMedian,
while being as computationally efficient as MaxMedian.
On the other hand, choosing the same length for the exploration phase of the two ETC leads to a significantly smaller time complexity for QoMax-ETC. Hence, both QoMax-SDA and QoMax-ETC present a substantial improvement over their counterparts.



\subsection{Empirical Performance}

\begin{figure*}[hbt]
	\centering
	\includegraphics[width=0.45\textwidth]{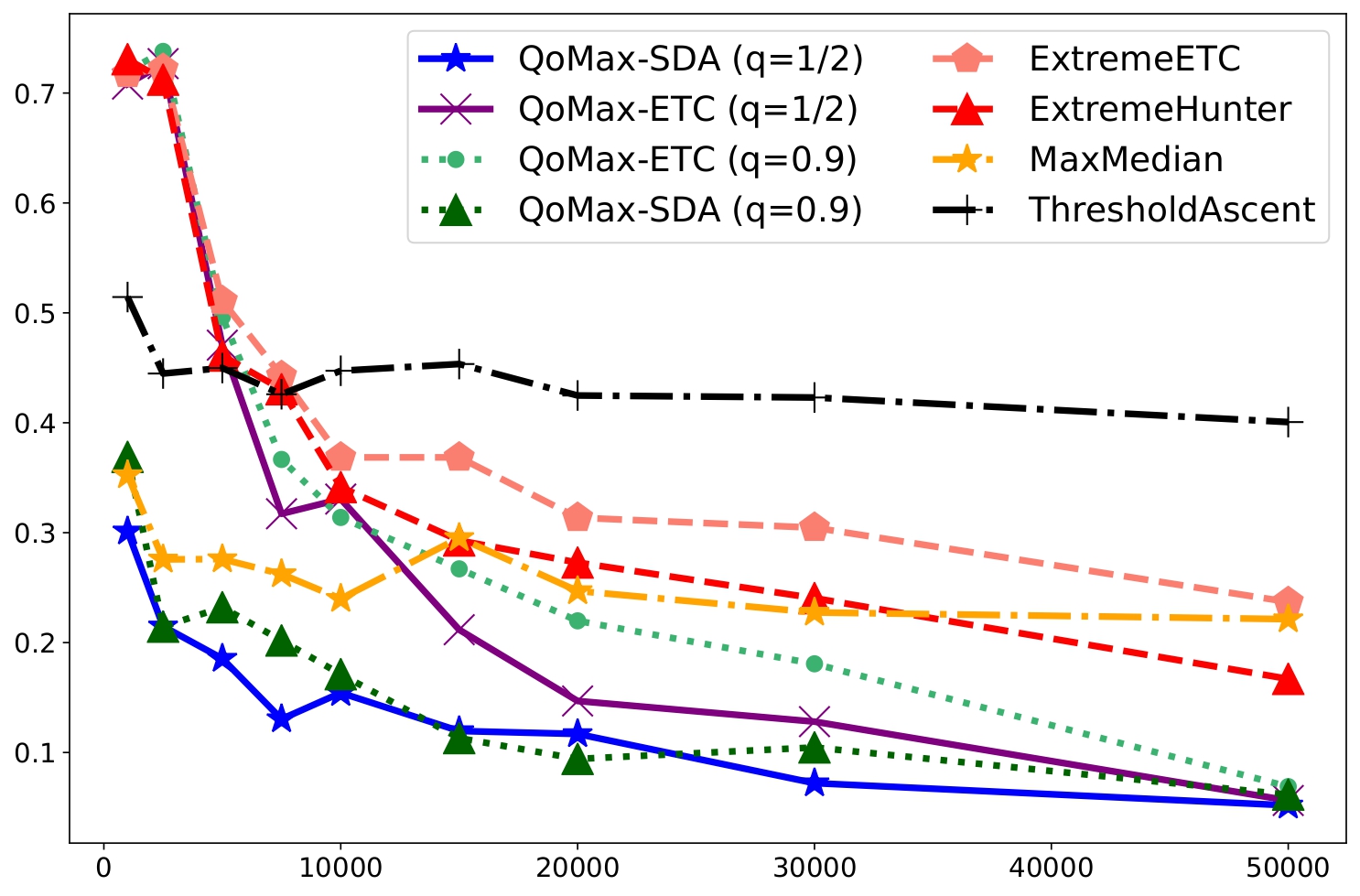} \qquad \qquad \includegraphics[width=0.45\textwidth]{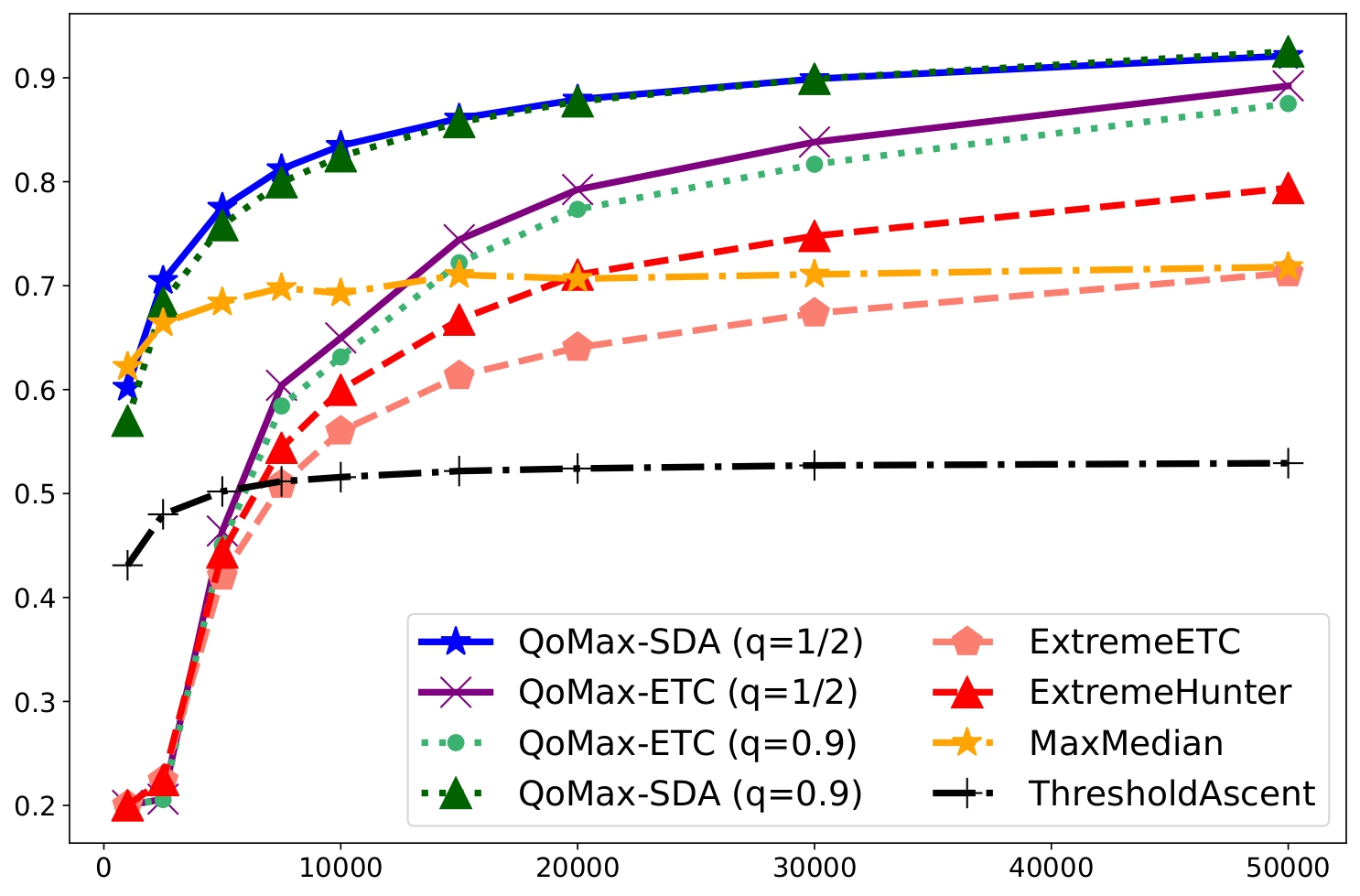}
	\caption{Proxy Empirical Regret \textbf{(I)} and Percentage of best arm pulls \textbf{(II)} averaged over $10^4$ independent trajectories for $T \in \{10^3, 2.5\times 10^3, 5\times 10^3, 7.5\times 10^3, 9\times 10, 10^4, 1.5\times 10^4, 2\times 10^4, 3\times 10^4, 5 \times 10^4 \}$.}
	\label{fig::xp1}
\end{figure*}

We compare the empirical performance of the QoMax algorithms with the different competitors on synthetic data. We reproduced 6 experiments from previous works\footnote{Our code is available \href{https://github.com/ExtremeBandits/ExtremeBandits_submission}{here}}: all experiments from \cite{bhatt2021extreme} (Experiments 1-4 for us), and the experiments 1 and 2 from \cite{carpentierExtreme} (5-6 here).
We also implement new experiments with other families of distributions to highlight the generality of our approach. Due to space limitation, we present in this section
(1) our methodology for evaluating Extreme Bandits algorithms, and (2) the results for the experiment 1 of \cite{bhatt2021extreme}, as it illustrates well our findings across all the settings we tested. We analyze the results for the other experiments 
in Appendix~\ref{app::add_xp}.

\paragraph{Empirical evaluation} We consider 4 performance criteria: \textbf{(I)} an \textit{empirical evaluation of the extreme regret}, \textbf{(II)} the \textit{fraction of pulls of the optimal arm}, \textbf{(III)} the \textit{empirical distribution of the number of pulls of the optimal arm} and \textbf{(IV)} the \textit{empirical distribution of the maximal reward}, estimated over $N = 10^4$ independent trajectories for different values of the horizon $T$. Most works report only \textbf{(I)}, and \textbf{(II)} was first proposed by \cite{bhatt2021extreme}. Our analysis shows that the extreme regret of a strategy is closely related to its capacity to sample the optimal arm $T-o(T)$ times, so we think that \textbf{(II)} is indeed a good performance indicator. Criterion \textbf{(III)} completes it by displaying the following quantiles of the empirical distribution of best arm pulls: $q \in [1\%, 10\%, 25\%, 50\%, 75\%, 90\%, 99\%]$. 
Regarding \textbf{(I)}, we note that estimating the expectation $\bE[\max_{t\leq T}X_{I_t,t}]$ featured in the extreme regret is very hard, and that approximations of $\bE[X_{1,T}^+] \coloneqq \bE[\max_{t\leq T} X_{1,t}]$ are known only for a few families. Standard Monte-Carlo estimators will have a very large variance due to the heavy tails of the distributions (see illustrations in Appendix~\ref{app::add_xp}). Hence, we propose the following estimation strategy \textit{when a tight approximation of $\bE[X_{1, T}^+]$ is known}. We first find $\tilde{q} = \tilde{q}_{\nu_1,T}$ such that $\bE[X_{1,T}^+]$ is equal to the quantile of order  $\tilde{q}_{\nu_1,T}$ of $\nu_{1, T}^+$. 
We then compute the empirical quantile of order $\tilde{q}$ of the collected rewards, denoted by $\widehat X_T(q)$, as an estimator of their expected maximum. This allows to compute what we call Proxy Empirical Regret (PER), $\cR_{T}^{\text{proxy}} = \frac{\bE[X_{1,T}^+]-\widehat X_T(q)}{\bE[X_{1,T}^+]}$, where the normalization facilitates the check of a \textit{weakly vanishing} regret. We are able to compute \textbf{(I)} for experiments 1-6. When \textbf{(I)} is not available we recommend looking at \textbf{(IV)} with the same quantiles as for \textbf{(III)}.

\paragraph{Experiment 1} We consider $K=5$ Pareto distributions with parameters $[1.1, 1.3, 1.9, 2.1, 2.3]$. We chose this experiment because it enters in the theoretical guarantees of most baselines. 
The QoMax-based algorithms outperform their competitors in this problem, both in terms of \textbf{(I)} and \textbf{(II)} (see Figure~\ref{fig::xp1}). QoMax-SDA learns faster, but at horizon $T=5\times 10^4$ the ETC are close. The quantile $q=0.5$ performs (very) slightly better than $q=0.9$. 
Strikingly, QoMax algorithms surpass the two baselines designed for this parametric setting (ExtremeHunter, ExtremeETC). 
We also observe that the performance of ThresholdAscent and MaxMedian stops improving early, even if MaxMedian is competitive for $T\leq10^4$. 
To understand this phenomenon, we look at \textbf{(III)} (Table~\ref{tab::nb_pulls_xp1} in Appendix~\ref{app::add_xp}). 
Surprisingly, for at least 25\% of the trajectories MaxMedian ended up playing the optimal arm less than 35 times over $5\times 10^4$ pulls\footnote{Furthermore, we discuss in Appendix~\ref{app::proof_sda} a potential issue in the analysis of MaxMedian}. On the other hand, QoMax-SDA ($q=0.5$) selects the optimal arm at least $2\times 10^4$ times for $99\%$ of them. It also obtains much higher statistics on the empirical distribution of the maxima (see Table~\ref{tab::maxima_xp1} in Appendix~\ref{app::add_xp}, criterion \textbf{(IV)}). 

\paragraph{Other Experiments} The benefits of QoMax are also clear from experiments 2 to 5: we verify that they work well for Exponential (experiment 3), Gaussian tails (experiment 4), as well as for other Pareto examples (experiments 2 and 5) including one where the tail gap is $0$ (experiment 2). The impact of the number of arms is discussed, showing that for reasonable time horizons QoMax-SDA should be preferred over QoMax-ETC (experiment 4). Experiment 6 allows to discuss the limits of QoMax in a difficult scenario, in which the 2nd-order Pareto assumption allows ExtremeHunter and ExtremeETC to outperform all other algorithms. In this example, setting $q=0.9$ has a benefit as well as enforcing the sample obligation. Finally, in additional experiments we consider different families of heavy-tail distributions stressing out the generality of the dominance assumption under which QoMax algorithms are efficient. 

\paragraph{Conclusion} Overall, QoMax-based algorithms seem to be solid choices for the practitioner, as demonstrated in a variety of examples. Their strong theoretical guarantees and implementation tricks reducing the time and space complexities make them an efficient solution for the Extreme Bandits problem.

\newpage

\subsubsection*{Acknowledgements}
The PhD of Dorian Baudry is funded by a CNRS80 grant. This work has been supported by the French Ministry of Higher Education and Research, Inria, Scool,
and the French Agence Nationale de la Recherche (ANR) under grant ANR-19-CE23-0026-04 (BOLD project).

Experiments presented in this
paper were carried out using the Grid’5000 testbed, supported by a scientific interest group hosted by
Inria and including CNRS, RENATER and several Universities as well as other organizations (see
https://www.grid5000.fr).

The authors want to thank Mastane Achab, Alexandra Carpentier and Michal Valko for carefully answering our questions regarding the ExtremeHunter and ExtremeETC algorithms, and helping us for their implementation.


\bibliographystyle{abbrvnat}
\bibliography{biblio}

\clearpage
\appendix

\thispagestyle{empty}

\onecolumn \makesupplementtitle

%

\section{PROPERTIES OF MAXIMA}
\label{app::prop_maximax}

We first recall the notation from Section~\ref{sec::max_comp}. We consider i.i.d. samples $(X_{1,i})$ and $(X_{2,i})$ from two distributions $\nu_1$ and $\nu_2$ and denote by $X_{1,i}^+$ and $X_{2,i}^+$ their maxima. Our goal is to upper bound 
\[\max\left\{\bP\left(X_{1,n}^+ \leq x_n\right),
\bP\left(X_{2,n}^+ \geq x_n\right)\right\}\]
for a well chosen sequence $(x_n)$ under exponential and polynomial tails (Lemma~\ref{lem::max_comp}) and under the weaker assumption that $\nu_1\succ \nu_2$ (Lemma~\ref{lem::max_general}). In both cases, we start by writing 
\begin{eqnarray}
 \bP\left(X_{1,n}^+ \leq x_n\right) & = & (1-G_1(x_n))^{n} \leq \exp(-n G_1(x_n)) \label{UBX1}\\
 \bP\left(X_{2,n}^+ \geq x_n\right) & \leq &\sum_{i=1}^{n} \bP(X_{2,i} \geq x_n) = n G_2(x_n),\label{UBX2}
\end{eqnarray}
where $G_1$ and $G_2$ are the survival functions of $\nu_1$ and $\nu_2$ respectively.

\subsection{Proof of Lemma~\ref{lem::max_comp}}

\compmaxima*

\begin{proof}
	The key of the proof is to consider $x_n$ "slightly" 
	below $G_1^{-1}(1/n)$. Consider the exponential tails first, for which  $G_1(x) \sim C_1 \exp(-\lambda_1 x)$ and $G_2(x)\sim C_2 \exp(-\lambda_2 x)$, for some $(C_1, \lambda_1)$ and $(C_2, \lambda)$ with $\lambda_1 < \lambda_2$. Hence, for any $\epsilon>0$ it holds that for $x$ large enough, $G_1(x)\geq(1-\epsilon)C_1 \exp(-\lambda_1 x)$ and $G_2(x)\leq C_2(1+\epsilon)\exp(-\lambda_2 x)$. So, we prove without loss of generality the result by continuing the proof as if the survival functions were exactly equal to their equivalents, as we don't assume anything on $C_1$ and $C_2$. 
	
	We let $\delta = \frac{\lambda_2}{\lambda_1}-1$ and 
	choose 
	\[
	x_n = \frac{1}{\lambda_1}\left(\log n + \log(C_1) 
	- \log(\delta \log n)\right) \;.
	\]
	
	We now simply compute $G_1(x_n)$ and $G_2(x_n)$. First,	
	\begin{align*}
	G_1(x_n) &= C_1 \exp(-(\log n + \log C_1 - \log(\delta\log n))\\
	& = \frac{\delta(\log n)}{n} \;.
	\end{align*}
	Then, 
	\begin{align*}
	G_2(x_n) &= 
	C_2 \exp\left(-\frac{\lambda_2}{\lambda_1}
	\left(\log n + \log C_1 - \log(\delta\log n\right)\right) 
	\\
	& = \frac{1}{n^{\frac{\lambda_2}{\lambda_1}}} 
	\times (\delta \log n)^{\frac{\lambda_2}{\lambda_1}}
	\times\frac{C_2}{C_1^\frac{\lambda_2}{\lambda_1}}
	\end{align*}
	
	So finally, using Equation~\eqref{UBX1} and \eqref{UBX2} we obtain 
	\[
	\bP\left(Y_n^+ \geq x_n\right) 
	= \cO\left(\frac{(\log n )^{\delta+1}}
	{n^{\delta}}\right) \quad \text{and} \quad 
	\bP\left(X_n^+ \leq x_n\right) \leq \frac{1}{n^{\delta}} \;, 
	\]
	which gives the result.

	Now we consider polynomial tails, for which $G_1(x)=C_1 x^{-\lambda_1}$ and $G_2(x)=C_2 x^{-\lambda_2}$ for $x$ large enough. This time we define the sequence
	\[
	x_n = (C_1 n )^\frac{1}{\lambda_1} 
	\times (\delta \log n)^{-\frac{1}{\lambda_1}} \;,
	\]
	with $\delta = \frac{\lambda_2}{\lambda_1}-1$, as above. We obtain $\exp(-nG_1(x_n))=n^{-\delta}$, and $nG_2(x_n) = \cO\left(\frac{(\log n )^{\delta+1}}{n^{\delta}}\right)$, giving the result.
\end{proof}

\subsection{Proof of Lemma~\ref{lem::max_general}}

\lemmsecond*

\begin{proof}

	Let $q \in (0,1)$. We define the sequence $(x_n)$ by  
	\[G_1(x_n)=1-q^{\frac{1}{n}},\]
	so that $\bP(X_{1,n}^+ \leq x_n) = q$. 
	
	As $\nu_1\succ \nu_2$, there exists a constant $C>1$ such that $G_1(x)\geq C G_2(x)$ for $x$ large enough. 
	Hence, as $x_n \rightarrow +\infty$ it holds that $G_1(x_n)>CG_2(x_n)$ for $n$ large enough. 
	
	For such large enough $n$ we have
	\begin{align*}
	\bP(X_{2,n}^+ \leq x_n)& =(1-G_2(x_n))^n \\
	& \geq (1- \frac{1}{C} G_1(x_n))^n \\
	& = \left(1-\frac{1}{C} \left(1-q^\frac{1}{n}\right)\right)^n.
	\end{align*}
	Now we can consider the asymptotic behavior of this quantity, using first that $1- q^\frac{1}{n} \sim \frac{- \log q}{n}$, and deducing that \[\lim_{n \rightarrow \infty} \left(1-
	\frac{1}{C} \left(1-q^\frac{1}{n}\right)\right)^n = q^{\frac{1}{C}} \;.\]
	
Hence, for any $\epsilon_0>0$ when $n$ is large enough we have for this specific choice of $x_n$ \[\bP(X_{n,1}^+ \leq x_n)=q \ \ \text{ and} \ \ \ \bP(X_{2,n}^+\geq x_n) < 1 - q^\frac{1}{C}+\epsilon_0.\] It is clear from this point that if we change a bit $x_n$ in order to have $\bP(X_{n,1}^+\leq x_n) \leq q-\epsilon$, for some $\epsilon >0$, then $\bP(X_{2,n}^+\geq x_n) \leq 1-(q-\epsilon)^\frac{1}{C} + \epsilon$ holds for $n$ large enough. Taking $\epsilon$ small enough to obtain $1-(q-\epsilon)^\frac{1}{C} + \epsilon < 1-q-\epsilon$ concludes the proof.
\end{proof}

\subsection{Lower Bound}

A natural question is whether the rate obtained in Lemma~\ref{lem::max_comp} can be improved, and if it is really impossible to achieve an exponentially decreasing probability as for the comparison of empirical means. We show that this is not the case even under semi-parametric assumptions with the following result.

\begin{lemma}[Lower bound]\label{lem::prob_lower}
Assume that both $\nu_1$ and $\nu_2$ have either polynomial or exponential tails, with respective second parameter $\lambda_1$ and $\lambda_2$, with $\lambda_1<\lambda_2$ (so that $\nu_1\succ \nu_2$).
	\[ 
	\bP\left(X_{1,n}^+ \leq X_{2,n}^+\right) = \Omega\left(n^{-\frac{\lambda_2}{\lambda_1}}\right) \;.\]
\end{lemma}


\begin{proof} Letting $f_k$ and $F_k$ be the pdf and cdf of the distribution $\nu_k$ for $k\in \{1,2\}$. We lower bound the probability of interest as follows:
\begin{align*}
\bP(X_{1,n}^+ \leq X_{2,n}^+)  & \geq \bP(X_{2,1}\geq \max_{1\leq i \leq n} X_{1,i}) \\
&= \bE_{X \sim \nu_2}[F_1(X)^n] 
= \int_{\R} f_2(x) F_1(x)^n dx \\
&\geq \int_{m_n}^{+ \infty} f_2(x) F_1(x)^n dx 
 \geq F_1(m_n)^n G_2(m_n) \;, \\
\end{align*}
for any choice of $m_n$. If we choose $m_n = F_1^{-1}(1-\frac{1}{n})$, we have for exponential tails $m_n = \frac{1}{\lambda_1} (\log C_1 + \log n)$. If we plug this into $G_2$ we obtain a lower bound in $e^{-1} \frac{C_2}{C_1^{\lambda_2/\lambda_1}} \frac{1}{n^{\frac{\lambda_2}{\lambda_1}}}$. The same can be done for polynomial tails.
\end{proof}

\subsection{Maxima of (semi)-parametric distributions}\label{app::maxprop}

We first introduce a few notation to ease the presentation. In this section, we let $(X_{i})_{i\in\N}$ be an i.i.d. sequence from the distribution $\nu_1$ whose survival function is denoted by $G_1$. For any integers $n,m$ with $n < m$, we let $X_{n:m}^{+} = \max_{i \in \{n,\dots,m\}} X_i$ and use the shorthand $X_n^{+} = X_{1:n}^{+}$. 

We first recall known results about the rate of growth of the expected maximum for distributions that have exponential or polynomial tails.

\begin{proposition}\label{prop::calcul_max} If $\nu_1$ has an exponential tail with parameters $C_1$ and $\lambda_1$, it holds that   
\[\bE\left[X_{T}^+\right]\underset{T \rightarrow \infty}{\sim} \frac{1}{\lambda_1}\log(T)\] 
If $\nu_1$ has a polynomial tail with parameters $C_1$ and $\lambda_1>1$, it holds that 
\[\bE\left[X_T^+\right]\underset{T \rightarrow \infty}{\sim} T^{\frac{1}{\lambda_1}}C_1^{\frac{1}{\lambda_1}} \Gamma\left(1 - \frac{1}{\lambda_1}\right)\] 
\end{proposition}

\begin{proof}
For exponential tails, we refer the reader to Appendix A.1 of \cite{bhatt2021extreme}. For polynomial tails, we can use Theorem 1 of \cite{carpentierExtreme} which applies to second-order Pareto distributions, and in particular Pareto distributions, for which $G(x) =\frac{C}{x^{\lambda}}$ for $x$ large enough, with exact equality. To handle our semi-parametric assumption, we first note that for all $B$, 
\begin{eqnarray*}
\bE\left[X_{T}^{+}\right]  & = &  \bE\left[X_{T}^{+}\ind\left(X_T^+ \leq M \right)\right]  + \bE\left[X_{T}^{+}\ind\left(X_T^+ > M\right)\right]\\
 & = & \bE\left[X_{T}^{+}\ind\left(X_T^+ \leq M\right)\right] + \int_{M}^{\infty}\left(1 - (1-G_1(x))^T\right)dx
\end{eqnarray*}
The first terms tends to zero when $T$ goes to infinity for any distribution that has an unbounded support, so if $G_{C,\lambda}(x) =\frac{C}{x^{\lambda}}$ is the survival function of an exact Pareto distribution, it follows that for all 
$M>0$
\[\int_{M}^{\infty}\left(1 - (1-G_{C,\lambda}(x))^T\right)dx \sim T^{\frac{1}{\lambda_1}}C_1^{\frac{1}{\lambda}} \Gamma\left(1 - \frac{1}{\lambda}\right).\]
Now assume that $G_1(x) \sim C_1x^{-\lambda_1}$ when $x$ tends to infinity. For all $\varepsilon>0$ there exists $M>0$ such that for $x > M$, 
\[(1-\varepsilon)\frac{C}{x^{\lambda}}\leq G_1(x) \leq (1 +\varepsilon)\frac{C}{x^{\lambda}}\]
and 
\begin{eqnarray*}
 \bE\left[X_{T}^{+}\right]  & \leq  & \bE\left[X_{T}^{+}\ind\left(X_T^+ 
 \leq M \right)\right] + \int_{M}^{\infty}\left(1 - (1-G_{(1+\varepsilon)C_1,\lambda_1}(x))^T\right)dx \sim  T^{\frac{1}{\lambda_1}}((1+\varepsilon)C_1)^{\frac{1}{\lambda_1}} \Gamma\left(1 - \frac{1}{\lambda_1}\right)\\
 \bE\left[X_{T}^{+}\right]  & \geq  & \bE\left[X_{T}^{+}\ind\left(X_T^+ \leq M
 \right)\right] + \int_{M}^{\infty}\left(1 - (1-G_{(1-\varepsilon)C_1,\lambda_1}(x))^T\right)dx \sim  T^{\frac{1}{\lambda_1}}((1-\varepsilon)C_1)^{\frac{1}{\lambda_1}} \Gamma\left(1 - \frac{1}{\lambda_1}\right),
\end{eqnarray*}
which permits to conclude the proof.
\end{proof}

We now recall Assumption~\ref{ass:esp_max}, under which we analyse QoMax-ETC and QoMax-SDA. 

\assespmax*

In order find a sufficient condition for Assumption~\ref{ass:esp_max} to be satisfied, for any constant $U >0$ we write
\begin{align*}
 \bE[X_{T}^{+}] - \bE[X_{T-N_T}^+]  &=  \bE\left[X_{T-N_T+1:T}^{+}\ind\left(X_T^+ = X_{T-N_T+1:T}^{+}\right)\right] \\
 & \leq  \bE\left[X_{T-N_T+1:T}^{+}\ind\left(X_T^+ = X_{T-N_T+1:T}^{+}\right)\ind\left(X_{T-N_T+1:T}^{+} \leq B \right)\right] \\
& \quad  + \bE\left[X_{T-N_T+1:T}^{+}\ind\left(X_{T-N_T+1:T}^{+} > B \right)\right] \\
 & \leq  B \bP\left(X_T^+ = X_{T-N_T+1:T}^{+}\right) + \int_{B}^{\infty}\bP\left(X_{T-N_T+1:T}^{+} > x\right)dx \\
 & =  B \frac{N_T}{T} + \int_{B}^{\infty}\bP\left(X_{N_T}^{+} > x\right)dx \\
 & \leq  B \frac{N_T}{T} + N_T\int_{B}^{\infty}\bP(X_1 > x)dx \\
 & \leq  N_T \left( \frac{B}{T} + \int_{B}^{\infty}G_1(x)dx\right).
\end{align*}
where we have used the fact that that maximum has the same probability to be attained in each batch of size $N_T$ and the union bound $\bP\left(X_{N_T}^{+} > x\right) \leq \sum_{i=1}^{N_T} \bP(X_i > x)$. 

To prove that Assumption~\ref{ass:esp_max} is satisfied for exponential and polynomial tails, in each case we exhibit a value of $B$ such that the resulting upper bound tends to 0.

\paragraph{Exponential tails} In that case $G_1(x) = O\left(C_1e^{-\lambda_1 x}\right)$ and there exists a constant $C >0$ such that 
\begin{align*}
\bE[X_{T}^{+}] - \bE[X_{T-N_T}^+] &
\leq N_T\left( \frac{B}{T} 
+ C\int_{B}^{\infty} e^{-\lambda_1 x} dx\right) \\
&= N_T\left( \frac{B}{T} + \frac{C}{\lambda_1} e^{-\lambda_1 B}\right) \;.
\end{align*}
If \emph{there exists} $\gamma \in (0,1)$ such that $N_T = o(T^\gamma)$, choosing $B = \frac{\log(T)}{\lambda_1}$ yields $\lim_{T\rightarrow \infty}  \bE[X_{T}^{+}] - \bE[X_{T-N_T}^+]  = 0$.

\paragraph{Polynomial tails}  In that case $G_1(x) = O\left(C_1x^{-\lambda_1}\right)$ for $\lambda_1 >1$ and there exists a constant $C >0$ such that 
\begin{align*}
\bE[X_{T}^{+}] - \bE[X_{T-N_T}^+] 
&\leq N_T\left( \frac{B}{T} + C\int_{B}^{\infty} \frac{1}{x^{\lambda_1}} dx\right) 
\\
&= N_T\left( \frac{B}{T} + \frac{C}{\lambda_1} B^{1-\lambda_1}\right)
\end{align*}
Choosing $B = T^{1/\lambda_1}$ yields 
\[ \bE[X_{T}^{+}] - \bE[X_{T-N_T}^+] \leq \left(1 + \frac{C}{\lambda_1}\right)\frac{N_T}{T^{1-\frac{1}{\lambda_1}}}\]
If \emph{for all} $\gamma \in (0,1)$, $N_T = o(T^\gamma)$ then in particular $N_T = o(T^{1-\frac{1}{\lambda_1}})$ and $\lim_{T\rightarrow \infty}  \bE[X_{T}^{+}] - \bE[X_{T-N_T}^+]  = 0$.



\clearpage
\section{PROOFS OF SECTION~\ref{sec::qomax_etc} (ETC)}
\label{app::proof_etc}

\regretetc*

\begin{proof}
We recall that $N_T = b_T \times n_T$ is the number of pulls of each arm during the exploration phase of the ETC algorithm (see Algorithm~\ref{alg:qomax-etc})
and that $X_{k,t}$ corresponds to the observation of arm $k$ at time $t$ (if any).
The ETC simplifies a lot the study of the extremal regret, as we can separate the explore and commit phase in the analysis. First, an exact decomposition of the expected value of the policy is 
	\[
	\bE\left[\max_{t \leq T} X_{I_t, t}\right] 
	= \bE\left[\max \left\{ \max_k \max_{t \leq K N_T} X_{k,t}
	, \max_{t=[ K N_T + 1,T]} X_{I_T,t}\right\}\right] \;.
	\]
	We obtain the lower bound by simply ignoring the exploration phase.
	\begin{align*}
\bE\left[\max_{t \leq T} X_{I_t, t} \right] 
&\geq \bE\left[\max_{t=[K N_T + 1,T]} X_{I_t,t}\right]  \\
&= 
\bE\left[\max_{t=[K N_T + 1,T]} X_{I_T,t}\right] \\
&= \bE\left[\max_{t=[K N_T + 1,T]} X_{I_T,t}\sum_{k=1}^K \ind(I_T=k)\right] \\
&= \sum_{k=1}^K \bE\left[\max_{t=[K N_T + 1,T]} X_{I_T,t} \ind(I_T=k)\right] \\
& = \sum_{k=1}^K \bP(I_T=k) \bE\left[\max_{t=[KN_T+1 ,T]} X_{k,t}\right] \\
& \geq \bP(I_T=1) \bE\left[\max_{t=[1,T-KN_T]} X_{1,t} \right] \\
& = (1-\bP(I_T\neq 1)) \bE\left[\max_{t=[1,T-KN_T]} X_{1,t} \right] \\
& \geq \bE\left[\max_{t \leq T-KN_T} X_{1,t}
\right] - \bP(I_T\neq 1)  
\bE\left[\max_{t \leq T} X_{1,t}
\right] \;.
\end{align*}

The fourth equality holds because the fact that arm $k$ is chosen by the algorithm after the exploration phase is independent of the rewards that are available for arm $k$ in the exploitation phase. We also used that as the distributions are supported on $\R$ the expectation of their maximum is positive for $T$ large enough. This concludes the proof.

\end{proof}
\clearpage
\section{PROOFS OF SECTION~\ref{sec::qomax_sda} (SDA)}
\label{app::proof_sda}

\subsection{Proof of Proposition~\ref{prop::regret_decompo_SDA_main}}
\regretsda*

\begin{proof}
We recall that $N_k(t)$ denotes the number of pulls of arm $k$ at time $t$.
\begin{align*}
\bE \left[\max_{t \leq T} X_{I_t, t} \right] 
& \geq \bE\left[\max_{n \leq N_1(T)} X_{1,n} \right] 
\quad (\textnormal{keeping observations from a single arm})
\\
& \geq  \bE \left[\max_{t \leq N_1(T)} X_{1,t} \ind(\xi_T^c)\right] \\
& \geq  \bE \left[\max_{t \leq T-KM_T} X_{1,t} \ind(\xi_T^c)\right]
 \\
& = \bE \left[\max_{t \leq T-KM_T} X_{1,t}\right] - 
\bE \left[\max_{t \leq T-KM_T} X_{1,t} \ind(\xi_T)\right] \\
& \geq \bE\left[\max_{t \leq T-KM_T} X_{1,t} \right] - 
\bE \left[\max_{t \leq T} X_{1,t} \ind(\xi_T)\right] \;.
\end{align*}

At this step the decomposition is very 
similar to the one of the ETC proof. However, this time 
the event $\xi_T$ is not independent on the 
maximum on the available rewards so we need 
to control the expectation more precisely. We use 
the notation $X_T^+ = \max_{t\leq T} X_{1,t}$ for simplicity, and
then consider a constant $x_T \in \R$ and write

\begin{align*}
\bE\left[\max_{t \leq T} X_{1,t} \ind(\xi_T)\right] 
& = 
\bE[X_T^+ \ind(\xi_T)]
\leq  \bE[X_T^+ \ind(\xi_T) \ind(X_T^+ \leq x_T)] 
+ \bE[X_T^+ \ind(\xi_T) \ind(X_T^+ \geq x_T)] 
\\
& \leq x_T \bP(\xi_T) + \bE[X_T^+ \ind(X_T^+ \geq x_T)] \;.
\end{align*}
This concludes the proof.
\end{proof}

Before going further with this result, we can make a few remarks.

\begin{remark}[Comparison with the regret bound for ETC strategies]
	The expression we obtain can be compared with the result for the ETC strategies. The first part (exploration cost) is similar, with $M_T$ as the total number of samples collected during the exploration phase. The second term is more complicated as we simply had $\bP(\xi_T) \bE[\max_{t\leq T} X_{1,t}]$ for the ETC strategy. We now require this decomposition because the event $\xi_T$ is correlated with all rewards from arm $1$. However, the upper bound $\bP(\xi_T) \bE[\max_{t\leq T} X_{1,t}]$ should hold because intuitively $\xi_T$ and the maximum should be negatively correlated, as $\xi_T$ corresponds to arm $1$ under-performing. This seems however very intricate to prove.
\end{remark}

\subsection{Proof of Lemma~\ref{lem::prob_xi_sda}}
\lemmaprobxisda*

\begin{proof}
We recall $\xi_T := \{N_1(T) \leq T-KM_T\}$. 
First, using $\sum_{k=1}^K N_k(T) = T$, we remark that 
\[
\bP(N_1(T) \leq T- KM_T) \leq
\bP(\exists k \geq 2, N_k(T) \geq M_T) 
\leq \sum_{k=2}^K \bP(N_k(T)\geq M_T)\;,
\]


We denote by $r_T$ the index of the round for which the number of observations equals or exceeds $T$. As at least one observation is collected at the end of the round it holds that $r_T\leq T$. Hence, we can obtain 
\begin{align*}
\bP(N_1(T) \leq T- KM_T) &
\leq \sum_{k=2}^K \bP(n_k(r_T) b_k(r_T) \geq M_T)  \leq \sum_{k=2}^K \bP(n_k(T) b_k(T) \geq M_T)\;.
\end{align*}

Using $b_k(T) = n_k(T)^\gamma$ and Markov inequality gives
\begin{align*}
\bP(N_1(T) \leq T- KM_T) 
& \leq \sum_{k=2}^K \bP(n_k(T)^{1+\gamma} \geq M_T) 
 \leq \sum_{k=2}^K \bP(n_k(T) \geq M_T^\frac{1}{1+\gamma})
 \leq \sum_{k=2}^K \frac{\bE[n_k(T)]}{M_T^\frac{1}{1+\gamma}} \;.
\end{align*}

For all $k\geq 2$, Lemma~\ref{lemm:pullsuboptimal} (proved in Section~\ref{app::lemma5proof}) shows that $\bE[n_k(T)] = \cO\left((\log T)^\frac{1}{\gamma}\right)$, with the tuning we choose for the algorithm. This concludes the proof.
\end{proof}

\subsection{Proof of Lemma~\ref{lemm:pullsuboptimal}}\label{app::lemma5proof}
The remaining part consists in upper bounding the expectation of the number of queries of each suboptimal arm for $T$ rounds of QoMax-SDA, for which we will apply techniques similar to the proof of LB-SDA (see \cite{lbsda}).

\begin{restatable}{lemma}{pullsuboptimal}
	\label{lemm:pullsuboptimal}
	Under QoMax-SDA with parameters $B(n)=n^{\gamma}$ and $f(r)=(\log r)^\frac{1}{\gamma}$, for all $k \geq 2$, there exists a constant $C_k$ such that the number of pulls of arm $k$ at time $T$ satisfies
	\[
	\bE[n_k(T)] \leq C_k \left(\log(T) \right)^{\frac{1}{\gamma}} + \mathcal{O}(1) \;.
	\]
\end{restatable}

\begin{proof}
In the proof, we denote the $q$-QoMAx from arm $k$ using the samples between the sample $n_1$ and the sample $n_2$, $\bar{X}^q_{k, n_1:n_2}$ and the $q$-QoMax from arm $k$ using the first $n$ samples by $\bar{X}^q_{k, n}$.
Note that we omit the dependency in the batch size $b$ because this one is implicit through $B(n) = n^\gamma$.

\paragraph{\textbf{(1)} A first decomposition.} We start with a decomposition similar to the one proposed for LB-SDA, which is that for any function $n_0(T)$ we have
\begin{align*}
\bE[n_k(T)] 
&= \bE\left[ \sum_{r=0}^{T-1} \ind(k \in \cA_{r+1}) \right]
= \bE\left[\sum_{r=0}^{T-1} 
\ind(k \in \cA_{r+1}, \ell(r)= 1)\right] 
+ \bE\left[\sum_{r=1}^{T-1}\ind(k \in \cA_{r+1}, \ell(r)\neq 1)\right] 
\\ 
& \leq \bE\left[\sum_{r=0}^{T-1} \ind(k \in \cA_{r+1}, 
\ell(r)= 1, n_k(r) \leq n_0(T))\right] 
+ \bE\left[\sum_{r=0}^{T-1} \ind(k \in \cA_{r+1}, 
\ell(r)= 1, n_k(r) \geq n_0(T))\right] \\
& 
\quad 
+ \quad \bE\left[\sum_{r=1}^{T-1}\ind(\ell(r)\neq 1)\right] \\
& \leq n_0(T) + \bE\left[
\underbrace{\sum_{r=0}^{T-1} \ind(k \in \cA_{r+1}, \ell(r)= 1, n_k(r) \geq n_0(T)}_{A}
\right] + \bE\left[\sum_{r=1}^{T-1}\ind(\ell(r)\neq 1)\right] \;,
\end{align*}
where we used that 
\begin{align*}
\sum_{r=0}^{T-1} \ind(k \in \cA_{r+1}, n_k(r) \leq n_0(T)) & \leq \sum_{r=0}^{T-1} \ind(k \in \cA_{r+1}, n_k(r) \leq n_0(T)) \\
& \leq \sum_{r=0}^{T-1} \sum_{n = 1}^{n_0(T)} \ind(k \in \cA_{r+1}, \ell(r)= 1, n_k(r)=n) \\
&  \leq \sum_{n = 1}^{n_0(T)} \left(\sum_{r=0}^{T-1}\ind(k \in \cA_{r+1}, \ell(r)= 1, n_k(r)=n)\right) \\
&  \leq \sum_{n = 1}^{n_0(T)} 1 
= n_0(T) \;,
\end{align*}

as the event $\{k \in \cA_{r+1}, n_k(r) = n\}$ can only happen at one round.

\paragraph{\textbf{(2)} Upper bound for A.} Now, we can upper bound the counterpart with $n_k(r) \geq n_0(T)$, using the concentration from Theorem~\ref{th::qomax_comp}.
\begin{align*}
A &\coloneqq \bE\left[\sum_{r=0}^{T-1} 
\ind(k \in \cA_{r+1}, \ell(r)= 1, n_k(r) \geq n_0(T))\right] \\
&\leq \bE\left[\sum_{r=0}^{T-1} \ind(k \in \cA_{r+1}, \bar X_{k, n_k(r)}^q
 \geq \bar X_{1, n_1(r)-n_k(r)+1:n_1(r)}^q, 
 \ell(r)= 1, n_k(r) \geq n_0(T))\right] \\
& \leq \bE\left[\sum_{r=0}^{T-1} \ind 
\left(\bar X_{k, n_k(r)}^q \geq x_{n_k(r)}, n_k(r) \geq n_0(T), k \in \cA_{r+1} \right) \right]
\\
& \quad + \quad 
\bE\left[\sum_{r=0}^{T-1} \ind 
\left(\bar X_{1, n_1(r)-n_k(r)+1:n_1(r)}^q \leq x_{n_k(r)}, 
\ell(r)= 1, n_k(r) \geq n_0(T), k \in \cA_{r+1}  \right) \right] \;,
\end{align*}
where we used that if the QoMax of $k$ exceeds the QoMax of $1$, then it is either larger than $x_n$ or the QoMax of $1$ is smaller than $x_n$ for any arbitrary choice of $x_n$. In our case, we will choose a convenient value of $x_n$ to use Theorem~\ref{th::qomax_comp}. Using union bounds on the number of queries it then holds that
\begin{align*}
&A \leq \bE\left[\sum_{r=0}^{T-1}\sum_{n_k=n_0(T)}^{T-1}\ind(\bar X_{k, n_k}^q \geq x_{n_k}, k \in \cA_{r+1}, n_k(r)=n_k) \right]  
\\
& \quad  +\quad \bE\left[\sum_{r=0}^{T-1}\sum_{n_k=n_0(T)}^{T-1}\sum_{n=r/K}^{T-1} \ind(\bar X_{1, n-n_k+1:n}^q\leq x_{n_k}, \ell(r)= 1, n_k(r)=n_k, k\in \cA_{r+1}, n_1(r)=n) \right] \;. 
\end{align*}

We now use the same trick as before to reduce the double sum on $r$ and $n_k$ to only one sum, and write that

\begin{align*}
A & \leq \bE\left[\sum_{n_k=n_0(T)}^{T-1}\ind(\bar X_{k, n_k}^q 
\geq x_{n_k}) \sum_{r=0}^{T-1} \ind(k \in \cA_{r+1}, n_k(r)=n_k) \right] 
+ \bE\left[\sum_{n_k=n_0(T)}^{T-1}\sum_{n=r/K}^{T-1} 
\ind(\bar X_{1, n-n_k+1:n}^q \leq x_{n_k}) \right] 
\\
& \leq \sum_{n_k=n_0(T)}^{T-1}\bP(\bar X_{k, n_k}^q \geq x_{n_k}) 
+ \sum_{n_k=n_0(T)}^{T-1}\sum_{n=r/K}^{T-1} 
\bP(\bar X_{1, n-n_k+1:n}^q\leq x_{n_k}) 
\\
& \leq \sum_{n_k=n_0(T)}^{T-1}\bP(\bar X_{k, n_k}^q
\geq x_{n_k}) + 
T \sum_{n_k=n_0(T)}^{T-1} \bP(\bar X_{1, n_k}^q \leq x_{n_k}) \;.
\end{align*}

Plugging the concentration result from Theorem~\ref{th::qomax_comp}, one has
\begin{align*}
A \leq \sum_{n_k=n_0(T)}^{T-1}e^{-c_k b(n_k)} + T \sum_{n=n_0(T)}^{T-1} e^{-c_1 b(n)} \leq Te^{-c_k b(n_0(T))} + T^2 \exp(-c_1 b(n_0(T)))  \;.
\end{align*}

Let $n_0$ the integer for which Theorem~\ref{th::qomax_comp} can be applied between the arm 1 and any arm $k$ for $k \geq 2$.
Now we choose, 
$n_0(T)=\max\left(b^{-1}
\left(\frac{2 \log T}{c_1}\right), b^{-1}\left(\frac{\log T}{c_k}\right), 
n_0 \right)$. With this choice, we get
\begin{equation}
\label{eq:A_UB}
A = \bE\left[\sum_{r=0}^{T-1} 
\ind(k \in \cA_{r+1}, \ell(r)= 1, n_k(r) \geq n_0(T))\right]  = \cO(1) \;.
\end{equation}
If we define $b(n)=n^\gamma$, using Equation~\eqref{eq:A_UB} and the decomposition for $\bE[n_k(T)]$, it holds that for some constant $C$

\begin{equation}
\bE[n_k(T)] \leq C (\log(T))^{\frac{1}{\gamma}} + \bE\left[\sum_{r=0}^{T-1} \ind(\ell(r)\neq 1) \right] + \cO(1) \;.
\end{equation}

The next step of the proof is to have a deeper look at $\bE\left[\sum_{r=0}^{T-1} \ind(\ell(r)\neq 1) \right]$.

\paragraph{\textbf{(3)} Upper bound for $\bE\left[\sum_{r=0}^{T-1} \ind(\ell(r)\neq 1) \right]$.}

We provide a similar decomposition as in \cite{lbsda}, considering the case where arm $1$ has already been leader and the alternative. Before that we recall the following property obtained by the definition of the leader
\begin{align*}
\ell(r) = k \Rightarrow n_k(r) \geq \left\lceil \frac{r}{K}\right\rceil \;.
\end{align*}

We then define $a_r=\left\lceil \frac{r}{4}\right\rceil$, 
and write that
\begin{equation}
\label{dec:Z}
\bP\left(\ell(r) \neq 1\right)= 
\bP\left(\{\ell(r) \neq 1\} \cap \cD^r\right) + 
\bP\left(\{\ell(r) \neq 1\} \cap \bar\cD^r\right)
\;,
\end{equation}
where we define $\cD^r$ the event under which the 
asymptotically dominating 
arm has been leader at least once in $[a_r,r]$.
$$
\cD^r = \{\exists u \in  [a_r, r] \text{ such that } \ell(u) = 1 \}.
$$
We now explain how to upper bound the term in the left hand side of Equation~\eqref{dec:Z}. 
We look at the rounds larger than some round $r_0$ that will be specified later in the proof.

We introduce a new event 
\begin{align*}
&\cB^{u} = \{\ell(u)=1, k \in \mathcal{A}_{u+1}, n_k(u)= n_1(u) \text{ for some arm } k \} \;.&
\end{align*}
Under the event $\cD^r$, $\{\ell(r)\neq 1 \}$ 
can only be true if the leadership 
has been taken over by a suboptimal arm at some round between $a_r$ and $r$, that is 
\begin{equation}
\{\ell(r)\neq 1\} \cap \cD^r \subset \cup_{u=a_r}^{r-1}\{\ell(u)=1, \ell(u+1)\neq 1 \} \subset \cup_{u=a_r}^{r} \cB^u \;.
\end{equation}

This is because a leadership takeover can only happen after a challenger has defeated the leader while having the same number of observations. Moreover, each leadership takeover has been caused by either (1) a QoMax of a challenger is over-performing, or (2) a QoMax of the leader is under-performing, with a sample size in each case larger than $s_r = \lceil a_r/K\rceil$. In addition, each of these QoMax can \textbf{only cost one takeover} (thanks to the subsampling scheme), hence we can simply use an union bound on these events. In summary, after defining some $r_0>8$ we have that

\begin{align*}
\bE\left[\sum_{r=0}^{T-1}\ind(\ell(r) \neq 1, \cD_r)\right] & \leq r_0 + \bE\left[\sum_{r=r_0}^{T-1}\left( \sum_{u=a_r}^r \ind(\cB^u) \right)\right] \\
& \leq r_0 + \bE\left[\sum_{r=r_0}^{T-1} \left(\sum_{n=s_r}^r \left( \ind(\bar X_{1, n}^q \leq x_n) + \sum_{k=2}^K \ind(\bar X_{k, n}^q \geq x_n)\right) \right)\right] \\
& \leq r_0 + \sum_{r=r_0}^{T-1} \sum_{n=s_r}^r \left(\bP(\bar X_{1, n}^q\leq x_n) + \sum_{k=2}^K \bP(\bar X_{k, n}^q \geq x_n) \right) \\
& \leq r_0 + \sum_{k=1}^K \sum_{r=r_0}^{T-1} \sum_{n=s_r}^r \exp(-c_k b(n)) \\
& \leq r_0 + \sum_{k=1}^K \sum_{r=r_0}^{T-1} r \exp(-c_k b(s_r)) \\
& = \cO(1) \;,
\end{align*}

since $b(s_r) = s_r^\gamma = \Omega(r^\gamma)$. This is true if $s_{r_0} \geq n_0$, which is the condition that allows the use of the concentration inequality from Theorem~\ref{th::qomax_comp}. We consider $r_0$ large enough to satisfy this condition.

We now handle the case when the asymptotically dominant arm has never been leader between $a_r$ and $r$, which implies that it has lost a lot of duels 
against the respective leaders of many rounds. We introduce
$$
\cL^r= \sum_{u=a_r}^r \ind(\cC^u)\;,
$$
with $\cC^u = \{\exists k \neq 1, \ell(u)=k, 1 \notin \cA_{u+1} \}$.
It is proved in \cite{chan2020multi} that 
\begin{equation}
\label{eq:upper_z}
\bP(\ell(r) \neq 1 \cap \bar \cD^r) \leq \bP(\cL^r\geq r/4)\;. 
\end{equation}
and the author uses the Markov inequality to provide the upper bound
\begin{equation}
\label{eq:upper_z_2}
\bP(\cL^r\geq r/4) \leq \frac{\bE(\cL^r)}{r/4}= \frac{4}{r} \sum_{u=a_r}^r \bP(\cC^u)\;.
\end{equation}

From this step we can refactor $\sum_{r=r_0}^r \bP(\cL^r \geq r/4)$ using the following trick from \cite{baudry2020sub},

\begin{align*}
\sum_{r= r_0}^{T-1}  \frac{4}{r} \ind(u \in [a_r, r]) &= 
\sum_{r= r_0}^{T-1}  4 \frac{\ind(u\leq r)}{r} \ind(a_r \leq u) 
 \leq \frac{4}{u} \sum_{r= r_0}^{T-1} \ind(a_r \leq u) \\
&  \leq \frac{4}{u} \sum_{r= r_0}^{T-1} \ind(\lceil r/4 \rceil \leq u) 
  \leq \frac{4}{u} \sum_{r= r_0}^{T-1} \ind(r/4 \leq u+1) \\
& \leq  \frac{4}{u}\times 4(u+1)
 \leq 32 \;.
\end{align*}

With this result we obtain that 
\[
\sum_{r= r_0}^{T-1} \bP(\ell(r) \neq 1 \cap \bar \cD^r) \leq \sum_{r= r_0}^{T-1} \bP( \cL^r\geq r/4) \leq 32 \sum_{r=a_{r_0}}^{T-1} \bP(\cC^r) \;.
\] 

Now we can have a more precise look at $\bP(\cC^r)=\bP(\exists k \neq 1, \ell(r)=k, 1 \notin \cA_{r+1})$. We recall that we defined in the algorithm a forced exploration $f(r)$, ensuring that $n_k(r) \geq f(r)$ for any arm $k$ and any round $r$. 

\begin{align*}
\sum_{r=a_{r_0}}^{T-1} \bP(\cC^r) & 
\leq \sum_{r=a_{r_0}}^{T-1} \bP\left(\{\bar X_{1, n_1(r)}^q 
\leq x_{n_1(r)}\} \cup_{k=2}^K \left\{
\bar X_{k, n_k(r)-n_1(r)+1: n_1(r)}^q \geq x_{n_1(r)}, 
\ell(r)=k \right\}\right) 
\\
& \leq \sum_{r=a_{r_0}}^{T-1} \sum_{n = f(r)}^{r/2} 
\bP(\bar X_{1, n}^q \leq x_{n}) + \sum_{k=2}^K 
\sum_{r=a_{r_0}}^{T-1} \sum_{n = f(r)}^{r/2} 
\sum_{n_k=\lceil r/K \rceil}^r \bP(\bar X_{k, n_k-n+1:n_k}^q 
\geq x_{n}) 
\\
&  \leq \sum_{r=a_{r_0}}^{T-1} 
\sum_{n = f(r)}^{r/2} \exp(-c_1 b(n)) + 
\sum_{k=2}^K \sum_{r=a_{r_0}}^{T-1} \sum_{n = f(r)}^{r/2} 
r \exp(-c_k b(n)) \\
& \leq \sum_{r=a_{r_0}}^{T-1} \frac{r}{2} \exp(-c_1 b(f(r))) 
+ \sum_{k=2}^K \sum_{r=a_{r_0}}^{T-1} \frac{r^2}{2} \exp(-c_k b(f(r) ) ) 
\;,
\end{align*}


where the use of the concentration from Theorem~\ref{th::qomax_comp} is permitted only if $f(a_{r_0})\geq n_0$. Now, this result provides a sound theoretical tuning for the forced exploration parameter as a function of $b$, as choosing $f(r) \geq \max_k\left(\frac{4 \log r}{c_k}\right)^\frac{1}{\gamma}$ ensures 

\[\sum_{r=a_{r_0}}^{T-1} \bP(\cC^r) = \cO(1)\;. \]

Hence, we obtain the final result that for some constant $C_k$ it holds that

\[\bE[n_k(T)] \leq C_k(\log(T))^{\frac{1}{\gamma}} + \cO(1) \;,\]

under the assumptions that Theorem~\ref{th::qomax_comp} can be applied and that the forced exploration is of the same scaling as the regret, namely $f(r)=\Omega((\log r)^\frac{1}{\gamma})$.
\end{proof}

\subsection{Proof of Theorem~\ref{th::vanishing_qomax_sda}}
\thmregretsda*

\begin{proof} We instantiate the decomposition of Proposition~\ref{prop::regret_decompo_SDA_main} using the value of $\bP(\xi_T)$ obtained in Lemma~\ref{lem::prob_xi_sda}. Plugging all of these values and using similar tricks as those already used in Appendix~\ref{app::maxprop} to establish Assumption~\ref{ass:esp_max} for semi-parametric tails, we write for $\pi$ being any instance of QoMax-SDA with parameter $\gamma$,

\begin{align*}
\mathcal{R}_T^\pi &\leq \underbrace{\bE\left[\max_{t \leq T} X_{1,t}\right] -
	\bE\left[\max_{t \leq T- K M_T} X_{1,t} \right]}_{\text{Exploration cost}}
+ \underbrace{x_T \bP(\xi_T)
	+ \bE\left[\max_{t\leq T} X_{1,t}
	\ind\left(\max_{t\leq T} X_{1,t} \geq x_T\right)\right]}_{\text{Cost
		incurred by } \xi_T} \\
	& \leq \underbrace{K M_T \left[\frac{B_T}{T}+\int_{B_T}^{+\infty}G_1(x) dx\right]}_{(A_1)} + \underbrace{x_T \frac{C (\log T)^\frac{1}{\gamma}}{M_T^\frac{1}{1+\gamma}}}_{(A_2)} + \underbrace{T \int_{x_T}^{+\infty}G_1(x) dx}_{(A_3)} \;,
\end{align*}
for any values of $x_T, B_T, M_T$, that we now specify for each of the two families considered. 

\paragraph{Exponential tails} We recall that if $G_1(x) = \cO(\exp(-\lambda x))$, then for any $y\in\R$ we have
\[\int_{y}^{+\infty} G_1(x) dx = \cO\left(\exp(-\lambda y) \right) \;.\]

First, if we choose $B_T= \frac{1}{\lambda}\log(T)$ then $(A_1)$ 
vanishes for any choice of $M_T = T^\alpha$ with $0< \alpha< 1$. 
Similarly, choosing $x_T = \frac{2}{\lambda} \log T$ ensures that $(A_3)=\cO(1/T)$. Then, $(A_2)$ is in $\cO\left(\frac{(\log T)^{1+\frac{1}{\gamma}}}{M_T^{\frac{1}{1+\gamma}}}\right)$, which is vanishing for any choice of $M_T = T^\alpha$, $\alpha\in (0,1)$. We conclude that for exponential tails, $\lim_{T\rightarrow \infty} \cR_T^{\pi} = 0$.

\paragraph{Polynomial tails} Consider again $M_T = T^\alpha$, for some $\alpha\in (0, 1)$. This time,

\[\int_{y}^{+\infty} G_1(x) dx = \cO\left(\frac{1}{y^{\lambda-1}}\right)\;. \]

Plugging into $(A_3)$, we get a term of order $\cO(T\times x_T^{1-\lambda})$. Let's take $x_T=T^\beta$ for some $\beta \in (0, 1)$, we then have 
\[(A_3)=\cO(T^{1+\beta(1-\lambda)})\;. \]

Now consider $(A_2)$, omitting the polylog terms we obtain 
\[(A_2) =\cO(T^{\beta-\frac{\alpha}{1+\gamma}})\;.\]

Consider finally $(A_1)$. Choosing $B_T = T^{\frac{1}{\lambda}}$ (as in Appendix~\ref{app::maxprop}) we obtain the tightest upper bound on the exploration cost:
\[(A_1) = \cO\left(\frac{M_T}{T^{1-\frac{1}{\lambda}}}\right) = \cO(T^{\alpha-1+\frac{1}{\lambda}}) \;.\]

To get the smallest order with this proof technique we want to equalize all these three exponents, which gives

\[
\alpha-1+\frac{1}{\lambda} = \beta-\frac{\alpha}{1+\gamma} = 1+\beta(1-\lambda)  \;.
\]

For simplicity we write $\beta = \frac{1}{\lambda} + \eta$ and try to find $\eta$ instead. Re-writing the the three equalities yields

\[\alpha + \frac{1}{\lambda} -1 = \frac{1}{\lambda} + \eta -\frac{\alpha}{1+\gamma} = \frac{1}{\lambda}-(\lambda-1) \eta \;.
\]
This can be further simplified in 
\[\alpha -1 = \eta -\frac{\alpha}{1+\gamma} = -(\lambda-1) \eta \;.
\]

This gives in particular a system of two equations with two unknowns $\eta$ and $\alpha$. By substituing $\alpha$ we get
\begin{align*}
\eta - \frac{1- (\lambda-1) \eta }{1+\gamma} = - (\lambda-1) \eta \\ 
\Leftrightarrow \eta\left[1+\gamma + \lambda-1 + (\lambda-1)(1+\gamma) \right] = 1 \;,
\end{align*}
which gives $\eta = \frac{1}{\lambda(2+\gamma)-1}$ and $\alpha = \frac{\lambda(1+\gamma)}{\lambda(2+\gamma)-1}$. 

Plugging in these values, we obtain that $(A_1)$, $(A_2)$ and $(A_3)$ are all in $\cO\left(T^{\frac{1}{\lambda}-\frac{\lambda-1}{\lambda(2+\gamma)-1}}\right) = o(T^{1/\lambda})$. Recalling the rate of growth of the maximum for polynomial tails given in Proposition~\ref{prop::calcul_max} we get that for polynomial tails 
\[\cR_T^{\pi} = \underset{T\rightarrow \infty}{o}\left(\bE\left[\max_{t \leq T} X_{1,t}\right]\right).\]
\end{proof}

\subsection{Possible Mistake in the Analysis of Max-Median}

In the proof of Theorem 4.1 of \cite{bhatt2021extreme} the authors upper bound 
\[P\left(m(n) \geq v_n, W_i(n) \geq (1+\delta)\lambda_i^{-1}\log(m(n))\right)\]
where $v_n = \frac{1}{a}\sum_{d=1}^n \varepsilon_d$, $m(n) = \min_{k} N_k(n)$ and $W_i(n)$ is the index used by Max-Median for arm $i$, which is the order statics of order $\lfloor N_i(n)/m(n)\rfloor$. To do so, they use union bounds and concentration of a binomial random variable, which can be rewritten as follows: 
\begin{eqnarray*}
 P\left(m(n) \geq v_n, W_i(n) \geq (1+\delta)\lambda_i^{-1}\log(m(n))\right)& \leq & \sum_{m \geq v_n} \sum_{k \geq m} \bP\left(\cO_{i,k}\left(\left\lfloor\frac{k}{m}\right\rfloor\right) \geq (1+\delta)\lambda_i^{-1}\log(m)\right) \\
 & \leq & \sum_{m \geq v_n} \sum_{k \geq m} \bP\left(S_{k} \geq \frac{k}{m}\right) \;,
\end{eqnarray*}
where $S_k$ counts the number of observations among the $k$ first observations from arm $i$ that are exceeding $(1+\delta)\lambda_i^{-1}\log(m)$. From the tail assumption, $S_k$ is a binomial distribution with parameter $k$ and $p = a_i/m^{1+\delta}$. 

To upper bound this last probability, the authors use an exponential Markov inequality with a particular value of $\theta$. Using instead Chernoff inequality, which consists in optimizing over $\theta$ to get the smallest possible upper bound, one obtains 
\[\bP\left(S_{k} \geq \frac{k}{m}\right) = \bP\left(\frac{S_{k}}{k} \geq \frac{1}{m}\right) \leq \exp\left(-k \mathrm{kl}\left(\frac{1}{m}, \frac{a_i}{m^{1+\delta}}\right)\right),\]
provided that $k/m$ exceeds the mean $a_i/m^{1+\delta}$, where $\mathrm{kl}(x,y) = x\log(x/y) + (1-x)\log((1-x)/(1-y))$ is the binary relative entropy. Hence, for $n$ large enough, 
\[ P\left(m(n) \geq v_n, W_i(n) \geq (1+\delta)\lambda_i^{-1}\log(m(n))\right) \leq \sum_{m \geq v_n} \sum_{k \geq m}\exp\left(-k \mathrm{kl}\left(\frac{1}{m}, \frac{a_i}{m^{1+\delta}}\right)\right).\]
In the proof of Theorem 4.1, \cite{bhatt2021extreme} end up summing a quantity that does not depend on $m$, $\exp(-k^\delta/2)$, but it seems to be obtained by mistaking $k/m$ by $m$ in the tail probability of the binomial distribution. Without this mistake and with the tightest possible bound on the tail of a binomial distribution, the upper bound we obtain does depend on $m$. More precisely as 
\[
\mathrm{kl}\left(\frac{1}{m}, \frac{a_i}{m^{1+\delta}}\right) \sim \frac{\delta \log(m)}{m} \;,
\]
when $m$ tends to infinity, one obtains an upper bound of order 
\[
B_n = \sum_{m \geq v_n} \sum_{k \geq m}\exp\left(-k \frac{\delta \log(m)}{m}\right) = \sum_{m \geq v_n} \frac{\exp\left(-m \frac{\delta \log(m)}{m}\right)}{1 - \exp\left(-\frac{\delta \log(m)}{m}\right)} \;.
\]
Given that 
\[
\frac{\exp\left(-m \frac{\delta \log(m)}{m}\right)}{1 - \exp\left(-\frac{\delta \log(m)}{m}\right)} \sim \frac{m^{1-\delta}}{\delta \log(m)} \;,
\]
when $m$ tends to infinity, we don't see how we can get $\sum_{n}B_n< \infty$ (for any small enough $\delta$) which is needed in the rest of the proof of Theorem 4.1 in order to be able to apply Borel Cantelli's lemma.

\clearpage
\section{COMPLEMENTS OF SECTION~\ref{sec::xp}: PRACTICAL PERFORMANCE OF QOMAX ALGORITHMS} 
\label{app::add_xp}

\subsection{Implementation Tricks for QoMax-SDA}
\label{sec:implem_tricks_app}

 \begin{figure}[hbt]
	\centering
	\includegraphics[width=0.55\textwidth]{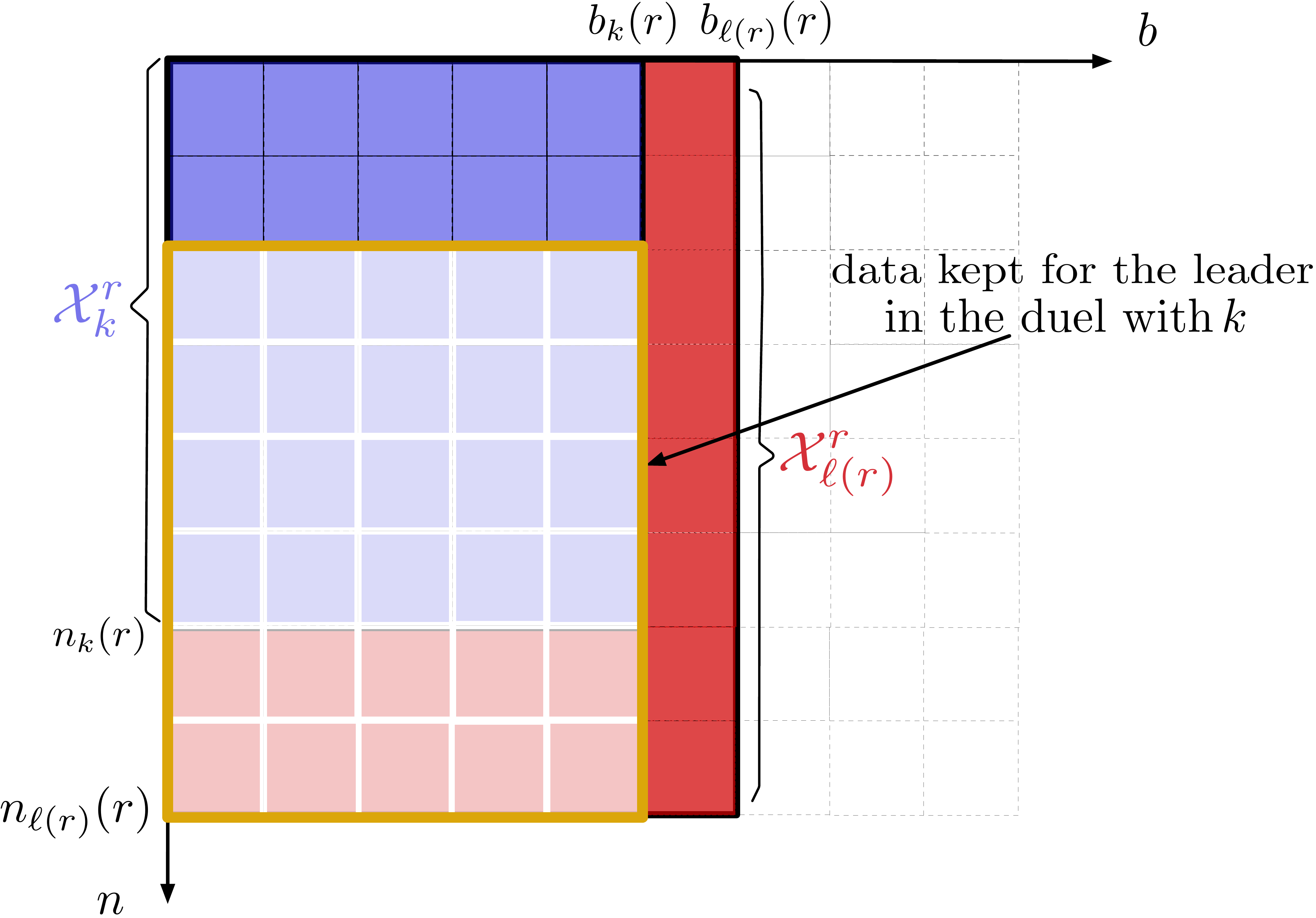}
	\caption{Illustration of the subsampling mechanism used by QoMax-SDA between the leader and a challenger
$k$ at round $r$.}
	\label{fig::subsampling}
\end{figure}

In this section we detail the CollectData procedure used for QoMax-SDA (see Algorithm~\ref{alg::qomax}) and briefly introduced in Section~\ref{sec::qomax_sda}. In particular, in Algorithm~\ref{alg:trick2} we describe the second implementation trick that allows to reduce significantly the memory complexity of QoMax-SDA. The principle is actually quite simple: the "last-block" subsampling that considers a subsample of the leader's history of the same size as the challenger's to compute their QoMax will take the \textit{last queries} for the leader as illustrated on Figure~\ref{fig::subsampling}.
Moreover, as we look for the maximum of this subsample on each batch, it is clear that we can remove a lot of information. For instance, imagine that the last data pulled for a batch is its global maximum.
Then all previously stored data in this batch is useless when looking at the last block and can be deleted.
If we apply this principle recursively, we have the following: (1) the newly pulled observation is necessarily kept (in case we consider a last block of size $1$), (2) we can remove \textit{all observations smaller than this last data}: again, looking at these samples in the past will not change anything for the value of the maximum. To implement this trick we store a list of data and a list of their indexes (i.e to which query of the arm they correspond) in order to know where to look at when performing the subsampling step for the leader. More than a storage trick, this is also time efficient: the global maximum of the batch is simply the first element of the list, and for a subsample of batches between queries $n_\ell-n_k$ and $n_\ell$ the subsample maximum is simply the observation corresponding to the first index in $\cI_\ell$ (defined in Algorithm~\ref{alg:collectdatasda}) larger than $n_\ell-n_k$.

\paragraph{Details of the procedure CollectData.} Considering Algorithm~\ref{alg:trick2} for the addition of new data, we can summarize the procedure in very few steps: (1) For each arm $k$ that is queried, \textbf{add one observation to each of its existing batches}. (2) For each queried arm $k$ that is \textit{not the leader}, \textbf{collect as many batches as necessary} to match the number of batches $B(n_k)$. (3) Collect as many batches as necessary for the leader to match the number of batches of the arm with the second largest number of queries.

\begin{algorithm}[H]
	\SetKwInput{KwData}{Input}
	\KwData{List of indices $\cI=\{i_1, \dots, i_L\}$, sorted list $\cX=\{X_1, \dots, X_L\}$, $X_1 > X_2 > \dots > X_L$,\\
		$\qquad \quad$ new index $i$, new data $X$}
	\SetKwInput{KwResult}{Return}
	\textbf{Search step: } Find the largest $j$ satisfying $X_j > X$ (Binary Search) \\
	\textbf{Update step:}\\
	{Set $\cX \leftarrow \{X_1, X_2, \dots,  X_j, X\}$ \color{purple}\small \tcp{Remove $X_{j+1}, \dots, X_L$ and add $X$}}
	{Set $\cI \leftarrow \{i_1, \dots, i_j, i \}$ \color{purple}\small \tcp{Remove $i_{j+1}, \dots, i_L$ and add $i$}}
	\KwResult{List of indices $\cI$, list of data $\cX$.}
	\caption{Efficient Update of a list of maxima for QoMax-SDA}
	\label{alg:trick2}
\end{algorithm}

\begin{algorithm}[h]
	\SetKwInput{KwData}{Input}
	\KwData{$K$ arms with data $\cX_k$ stored as $(\cI_k^{(j)},\cX_k^{(j)})_{j \in \{1, \dots, b_k\}}$, number of queries $n_k$ for each arm,
	distributions $(\nu_k)_{k \in K}$, $\cA$: set of arms chosen by QoMax-SDA, $\ell$ current leader, $B$ function controlling the batch size}
	\SetKwInput{KwResult}{Return}
	\For{$k \in \{1, \dots, K, \textcolor{blue}{\ell} \}$}{
	\If{$b_k>0$, \textcolor{blue}{$k \in \cA$}}{
		{$n_k \leftarrow n_k+1 $ {\color{purple} \small \tcp{Update the number of queries of arm $k$}}}
		\For{$j \in \{1,\dots, b_k\}$}{
			{Collect $X \sim \nu_k$ (Update batch $j$) \color{purple} \small \tcp{Add one observation in each existing batch $j$ of $\cX_k$}}
			$(\cI_k^{(j)}, \cX_k^{(j)}) \leftarrow
			\text{ EfficientUpdate}(\cI_k^{(j)}, \cX_k^{(j)}, n_k, X)$ (Alg. ~\ref{alg:trick2})
		}
	}
	\If{$k \neq \ell$}{{$B_{\text{new}} = B(n_k)$} \color{purple} \small \tcp{New batch size computed with $B$ if $k$ is a challenger.}}
\Else{$B_{\text{new}} = \max_{k \neq \ell} b_k$ \color{purple} \small \tcp{If $k$ is leader, align its batch size to the second most pulled challenger.}}
	\While{\textcolor{blue}{$b_k\leq B_{\text{new}}$}}{
		$\cI_k^{(b_k+1)}, \cX_k^{(b_k+1)} \leftarrow \{\}, \{\}$\\
		\For{$i \in \{1, \dots, n_k\}$}{
			{Collect $X \sim \nu_k$ \color{purple} \small \tcp{Collect $n_k$ data to form a new batch}}
			$(\cI_k^{(b_k+1)}, \cX_k^{(b_k+1)}) \leftarrow \text{ EfficientUpdate}(\cI_k^{(b_k+1)}, \cX_k^{(b_k+1)}, i, X)$ (Alg.~\ref{alg:trick2})\\
		}
		\hspace{0.06cm}
		Add $(\cI_k^{(b_k+1)}, \cX_k^{(b_k+1)})$ in $\cX_k$\\
		$b_k \leftarrow b_k+1$
	}
}
	\KwResult{$\cX_1, \dots, \cX_K$}
	\caption{Collect Data at the end of a round for QoMax-SDA}
	\label{alg:collectdatasda}
\end{algorithm}

\begin{figure}[hbt]
\centering
\includegraphics[scale=0.40]{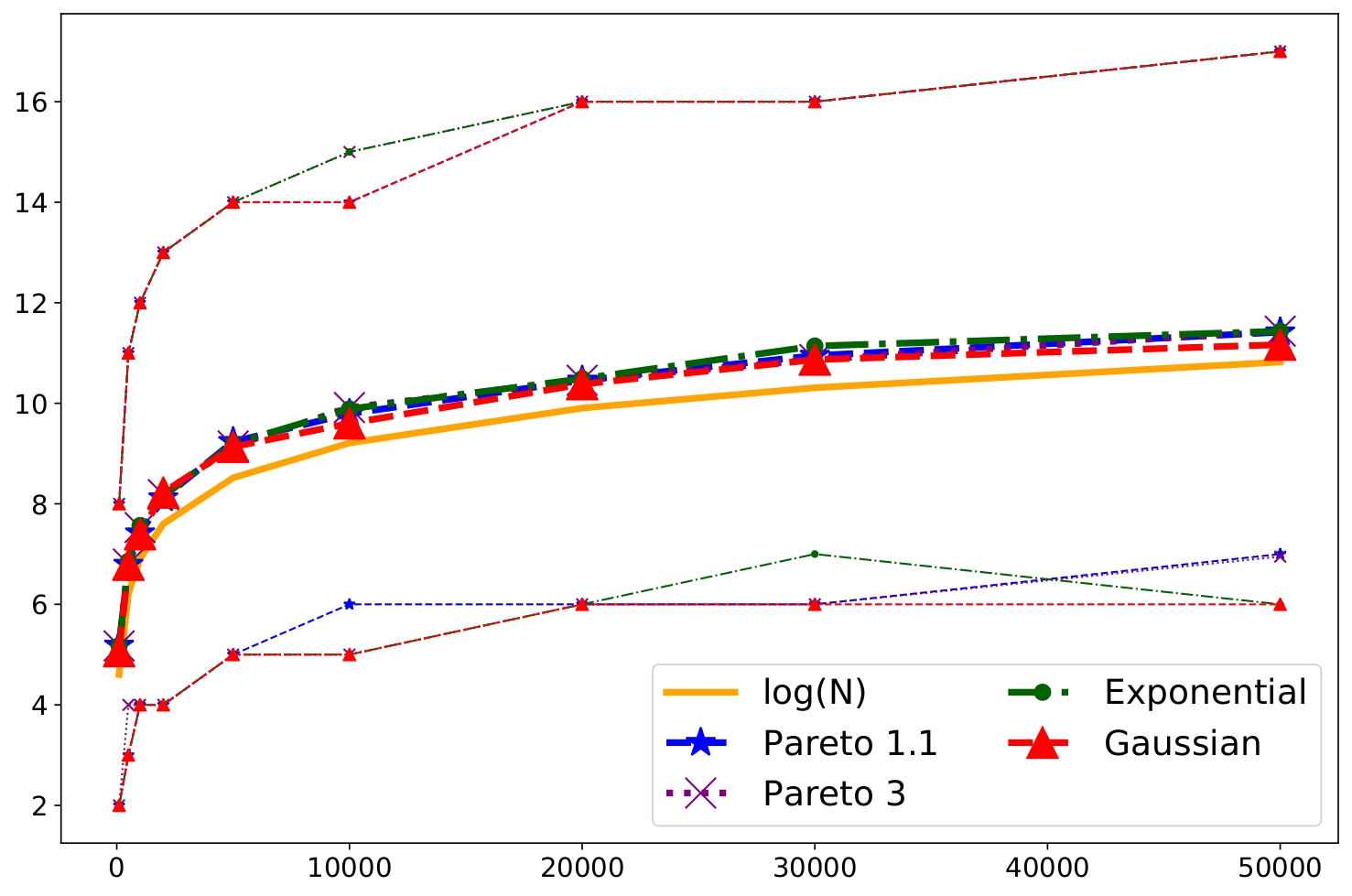} \qquad \caption{Average number of data kept in memory with the efficient storage of maxima, for $1000$ simulations with sample size $N \in [10^2, 5\times 10^2, 10^3, 2\times 10^3, 5\times 10^3, 10^4, 2\times 10^4, 3\times 10^4, 5\times 10^4]$ and the empirical $5\%$ and $95\%$ quantiles.}
\label{fig::storage_trick}
\end{figure}

\paragraph{Empirical evidences of the efficiency of the storage trick (Algorithm~\ref{alg:trick2}).} We propose simulations to verify that the solution we propose to store the data used by QoMAx-SDA is indeed efficient. We performed $1000$ simulations for each sample size $N \in [10^2, 5\times 10^2, 10^3, 2\times 10^3, 5\times 10^3, 10^4, 2\times 10^4, 3\times 10^4, 5\times 10^4]$), and for 4 distributions: (1) a Pareto distribution with tail parameter $1.1$, (2) a Pareto distribution with tail parameter $3$, (3) an exponential distribution with parameter $1$, (4) a standard normal distribution. We report in Figure~\ref{fig::storage_trick} the average number of data stored by the algorithm for each sample sizes, along with the empirical $5\%$ and $95\%$ quantiles on the $1000$ simulations and the curve $y=\log(N)$ for comparison. We observe that:
(1) The results do not depend on the distribution.
(2) All 4 curves are very close to exactly $\log(N)$, which is as small as $\approx 10$ for a sample size of $5\times 10^4$.
(3) $90\%$ of the simulations have no more than $17$ data stored, and the maximum we observe on all 4 experiments is actually $23$ which is very small compared to $N=5\times 10^4$.

Therefore, we conclude that the trick we introduced is indeed efficient and our experiments corroborate the intuition that it allows to store $\cO(\log N)$ data out of $N$ on average. We now prove it formally in Lemma~\ref{lem::storage_trick_log}

\begin{lemma}[Expected memory with the efficient storing of maxima] \label{lem::storage_trick_log}
Denote by $C_N$ the random variable denoting the memory usage of a random i.i.d sample of size $N$ drawn from any distribution with the implementation trick from Alg.~\ref{alg:trick2}. For any $\nu$, it holds that
\[\bE[C_N] = \sum_{n=1}^N \frac{1}{n} \sim \log(N)\;. \]
\end{lemma}
\begin{proof}
Denote the sorted random samples by $X_1 > \dots > X_N$. As the observations are i.i.d, all of them are equally likely to be in the last position. We consider $I_N$ the random variable denoting the index of the last observation, it holds that
\[	\bE[C_N] = \frac{1}{N} \sum_{j=1}^N \bE[C_N | I_N=j] \;.\]
Then, we remark that if $I_N=j$, all observations of higher order $X_{j+1}, \dots, X_N$ are removed from the history. Hence, it only remains to count the average amount of data considering only $X_1, \dots, X_{j-1}$, which is equal to $\bE[C_{j-1}]$ and add $1$ for the last observation.
Using that $\bE[C_1] = 1$,

\begin{align*}
&\bE[C_N] = \frac{1}{N} \sum_{j=1}^N \bE[C_N | I_N=j] = \frac{1}{N} \sum_{j=1}^N (1+\bE[C_{j-1}]) \\
&\Rightarrow (N+1) \bE[C_{N+1}]- N \bE[C_N] = \sum_{j=1}^{N+1} (1+\bE[C_j]) - \sum_{j=1}^{N} (1+\bE[C_j])
 = 1 + \bE[C_N] \\
&\Rightarrow (N+1) (\bE[C_{N+1}] - \bE[C_N]) = 1 \\
&\Rightarrow \bE[C_{N+1}]  = \bE[C_N] + \frac{1}{N+1} \\
&\Rightarrow \bE[C_N] = \sum_{n=1}^N \frac{1}{n} \;.
\end{align*}
\end{proof}


\subsection{More on the Storage/Computation Time}\label{app::storage_comp}

In this section we detail the computation of the storage constraints and time complexities reported in Table~\ref{tab::costs}. We restate them in Table~\ref{tab::costs_app} and express them as a function of their parameters.

\begin{table}[h]
	\caption{Average time and storage complexities
		of Extreme Bandit algorithms according to their parameters for a time horizon $T$.
	}
	\label{tab::costs_app}
	\begin{center}
		\begin{small}
			\begin{tabular}{ccc}
				\toprule
				Algorithm & Memory usage & Time complexity\\
				\midrule
				ThresholdAscent
				& $s$ & $\cO(KT)$ \\
				\midrule
				Extreme Hunter
				& $T$ & $\cO(T^2)$ \\
				\midrule
				MaxMedian
				& $T$ & $\cO(K T \log T)$ \\
				\midrule
				\textbf{QoMax-SDA} &
				$\cO((\log T)^2 + (K-1) \log (T) \log \log(T))$
				& $\cO(KT\log T)$ \\
				\midrule
				\midrule
				Extreme ETC
				& $K(\log T)^{2+\frac{1}{b}}$ & $\cO\left(K(\log T)^{2\times \left(2+\frac{1}{b}\right)}\right)$\\
				\midrule
				\textbf{QoMax-ETC} &
				\textbf{$K b_T$}
				& $\cO\left(\max\left\{K b_T  n_T, K b_T \log(b_T) \right\}\right)$  \\
				\bottomrule
			\end{tabular}
		\end{small}
	\end{center}
\end{table}

\paragraph{ThresholdAscent \citep{streeter2006simple}.} After simplifying the statement of the algorithm presented in their paper (beginning of Section 3) for continuous distributions, we find out that the algorithm actually considers the $s$ \textit{largest observations observed so far} where $s$ is a parameter of the algorithm. Indeed as long as the threshold is larger than $s$ observations, the threshold is increased.
The authors suggest taking $s=100$, implying a memory complexity as small as $100$ observations. After remarking this, the implementation of the algorithm is simplified largely (see our code): we can drop all data that are not in the $s$ largest observed so far, and the index needs to be re-computed only if this list changes. Asymptotically we expect this list to change very rarely, hence the time complexity of the algorithm is dominated by the check that an observation is larger than the $s$-th largest reward collected so far.


\paragraph{ExtremeHunter/ExtremeETC \citep{carpentierExtreme, achab2017max}.} The values we provide for these algorithms in Table~\ref{tab::costs_app} come directly from \cite{achab2017max}. We recall that in Table~\ref{tab::costs} we considered $b=1$, but we remark that even with $b$ very large ($b=+\infty$ corresponds to exact Pareto distributions) the memory and time complexities cannot go below $K(\log T)^2$ and $K(\log T)^4$ respectively.

\paragraph{MaxMedian \citep{bhatt2021extreme}.} In short, MaxMedian essentially tracks the quantile of order $1/m(t)$ for each arm, where $m(t)$ is the number of samples from the arm that has been pulled the least. Technically, even if it is unlikely, all observations could be used in the future (if we continuously collect data that are smaller than all values we obtained so far).
For this reason, all $T$ observations have to be stored.

\paragraph{QoMax-ETC (this paper).} The storage required is simply $K b_T$, which corresponds to storing online the maximum of each batch ($b_T$ batches) for each arm. Collecting the $b_T$ maxima takes a $\cO(K \times n_T\times b_T)$ time (collecting an observation and comparing it with a current maximum costs $\cO(1)$). At the final step of the exploration phase, 
computing the QoMax takes an additional $\cO(K b_T \log b_T)$, which is the cost of sorting $K$ lists of size $b_T$. So, as a function of $(n_T, b_T)$, the complexity of the algorithm is $\cO\left(\max\left\{b_T\times n_T, b_T \log b_T \right\}\right)$.

\paragraph{QoMax-SDA (this paper).} We recall that QoMax-SDA uses a batch size $n^\gamma$ for an arm that has been queried $n$ times, a forced exploration $(\log r)^{1/\gamma}$, and that the total number of queries of every sub-optimal arm is provably $\cO((\log T)^{1/\gamma})$ (see Theorem~\ref{th::vanishing_qomax_etc}). We start with the computation of the memory capacity and detail how the two implementation tricks presented in~\ref{sec:implem_tricks_app} work: (1) indexing the number of batches of the leader to the second most pulled arm allows to reduce the number of batches of the leader from $T^\gamma$ 
to
$\cO\left(\left((\log T)^{1/\gamma} \right)^\gamma\right)=\cO(\log T)$, for any $\gamma$. Then, we look at how many data are stored in each batch, and (2) according to Lemma~\ref{lem::storage_trick_log} the efficient storage of maxima allows to store only $\cO(\log N)$ observations out of $N$ on average.
This gives $\cO(\log T)$ for the leader, and $\cO(\log \log T)$ for the challengers. This explains why the dependency of $K$ in the memory becomes a second order term for $T$ large enough.
Then, we consider the computational time, which can be divided into two steps that are executed at each round: (a) updating the lists of values (each batch of the $K$ arms), and (b) computing the $K-1$ QoMax for the challengers and the $K-1$ QoMax for the leader. Operation (a) requires to find the index from which previous data can be erased. As the list is sorted (by construction), this can take up to $(\log N)$ with $N$ the sample size of a batch using a binary search. Hence, this gives $\cO(\log \log T)$ for each batch of the leader and $\cO(\log \log \log T)$ for each batch of the challengers. 
On the other hand, for step (b) the efficient storage ensures that we have access to the maximum of each batch at constant cost (first observation of the list), and we only need to find the quantiles over the different batches, giving $\cO(2(K-1)\log T)$. Hence, we can report an overall $\cO(K T \log T)$ time complexity, or a $\cO( T \log T \log \log T)$ when $T$ is very large. We report the first, because if $K=5$ then $\log \log T > K$ only for $T>10^{65}$, which is unreasonably large.

\subsection{Supplementary for Section~\ref{sec::xp} : Additional Experiments}

In this section we provide the complete results for all the experiments we performed and that were advertised in Section~\ref{sec::xp}. We first reproduce the experiments from previous papers, and then consider a few new settings. Before that, we detail the parameters used for each algorithm.

The code to reproduce the experiments is available on \href{https://github.com/ExtremeBandits/ExtremeBandits_submission}{Github}.

\paragraph{Parameters for All Experiments.} We recall
the parameters we used for the different experiments. 
For each experiment, we run $N=10^4$ independent trajectories for $10$ time horizons $T \in [1000, 2500, 5000, 7500, 9000, 10000, 15000, 20000, 30000, 50000]$.
This methodology is computationally expensive
but
allows for a fair comparison between ETC and more adaptive strategies. Furthermore, it is also a way to stabilize the results because if the same trajectories were used to plot the results for different time horizons then a few extreme trajectories for some algorithms would have too much influence on our conclusions. This is not a problem as all runs for $T\leq 20000$ are actually quite fast with parallel computing, and the total computation time of our experiments is largely dominated by the experiment with $T=50000$. 

The parameters we used are the following:

\begin{itemize}
	\item ThresholdAscent: $s=100, \delta=0.1$, as suggested in \cite{streeter2006simple}.
	\item ExtremeETC/ExtremeHunter: $b=1$, as in \cite{carpentierExtreme}. As the authors, we use $\delta=0.1$ for the experiments instead of the theoretical value that is too large for the time horizons considered, and $D=E=10^{-3}$ for the UCB. Other theoretically-motivated parameters are $r=T^{-1/(2b+1)}$ (fraction of samples used for the tail estimation),  $N=(\log T)^{\frac{2b+1}{b}}$ (length of the initial exploration phase). $\delta=\exp(-\log^2(T))/(2TK)$ in the paper but set to $0.1$ here.
	\item MaxMedian: The exploration probability is set to $\epsilon_t=1/(1+t)$ as suggested in \cite{bhatt2021extreme}.
	\item QoMax-ETC: We test $q=1/2$ and $q=0.9$, $b_T=(\log T)^2$ and $n_T=\log T$ to match both the theoretical requirements of Section~\ref{sec::qomax_etc}
	and the length of the exploration phase of ExtremeETC for a fair comparison.
	\item QoMax-SDA: $f(r)=(\log r)^\frac{1}{\gamma}$ and $B(n)=n^\gamma$ for $\gamma=2/3$, which works well across all the experiments we performed.
	The quantile is either equal to $q=1/2$ or $q=0.9$.
 \end{itemize}

\subsubsection{Experiments 1-6}

We describe the setting of each experiment, that we will then refer by their number (e.g exp.1).
\begin{enumerate}
	\item (exp.1 in \cite{bhatt2021extreme}): $K=5$ Pareto distributions with tail parameters $\lambda_k \in [2.1, 2.3, 1.3, 1.1, 1.9]$.
	\item (exp.2 in \cite{bhatt2021extreme}) $K=7$ Pareto distributions with $\lambda_k \in [2.5, 2.8, 4, 3, 1.4, 1.4, 1.9]$. All arms have a scaling $C=1$ except arm $5$ with $C_5=1.1$. Hence $\nu_5$ is the dominating arm
	from a slight margin.
	\item (exp.3 in \cite{bhatt2021extreme}) $K=10$ Exponential arms with a survival function $G_k(x) = e^{-\lambda_k x}$ with parameters $\lambda_k = [2.1, 2.4, 1.9, 1.3, 1.1, 2.9, 1.5, 2.2, 2.6, 1.4]$.
	\item (exp.4 in \cite{bhatt2021extreme}) $K=20$ Gaussian arms, with same mean $\mu_k =1, \forall k$, and different variances
	$\sigma_k = [1.64, 2.29, 1.79, 2.67, 1.70, 1.36, 1.90, 2.19, 0.80, 0.12, 1.65, 1.19, 1.88, 0.89, 3.35, 1.5, 2.22, 3.03, 1.08, 0.48]$. The dominant arm has a standard deviation $3.35$.
	\item (exp.1 in \cite{carpentierExtreme}) $K=3$ Pareto distributions with $\lambda \in [5, 1.1, 2]$.
	\item (exp.2 in \cite{carpentierExtreme}) $K=3$ arms, including 2 Pareto distributions with $\lambda_k \in [1.5, 3]$, and arm $3$ is a mixture Dirac/Pareto: pull $0$ with $80\%$ probability, reward from a Pareto distribution with $\lambda=1.1$ with $20\%$ probability. Hence, the last arm dominates asymptotically.
\end{enumerate}

\paragraph{Objective of each experiment.} Before reporting the results, we explain why each experiment is interesting in our opinion for the empirical evaluation of Extreme Bandits algorithms.
Experiment 1 is quite difficult because the tail gap between arm $3$ and arm $4$ is relatively small.
Otherwise, all algorithms are supposed to have guarantees in this setting so their comparison is fair. Experiment 2 allows to consider a semi-parametric setting with a tail gap
$\delta_{5-6} =0$, hence it only holds that $\nu_5 \succ \nu_6$: we check whether the algorithms are able to (1) pull $5$ and $6$ most often, and (2) arbitrate in favor of arm $5$. Then, experiments 3 and 4 allow to test the different algorithms respectively with exponential and gaussian tails (with different variances), showing the performance of the algorithms when the tails are not polynomial. Moreover, a larger number of arms is considered in these experiments. Finally, experiment 5 is relatively easy and more of a sanity check for the performance of each algorithm (it was exp.1 in \cite{carpentierExtreme}). Experiment 6 will be interesting for discussing the limits of parameter-free approaches, as the dominant tail provides low rewards with relatively high probability.

\begin{figure*}[h]
	\centering
	\includegraphics[width=0.45\textwidth]{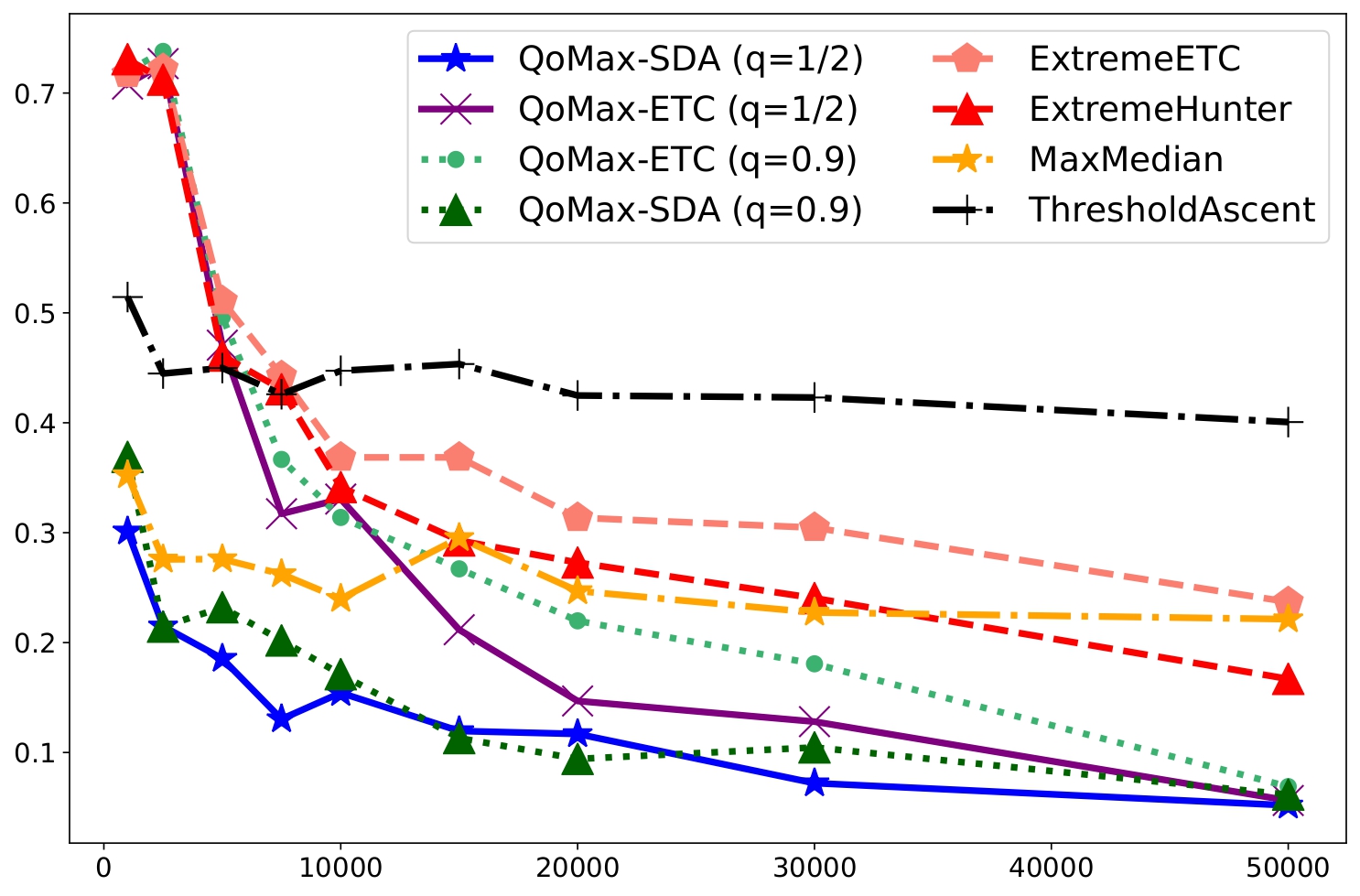} \qquad \qquad \includegraphics[width=0.45\textwidth]{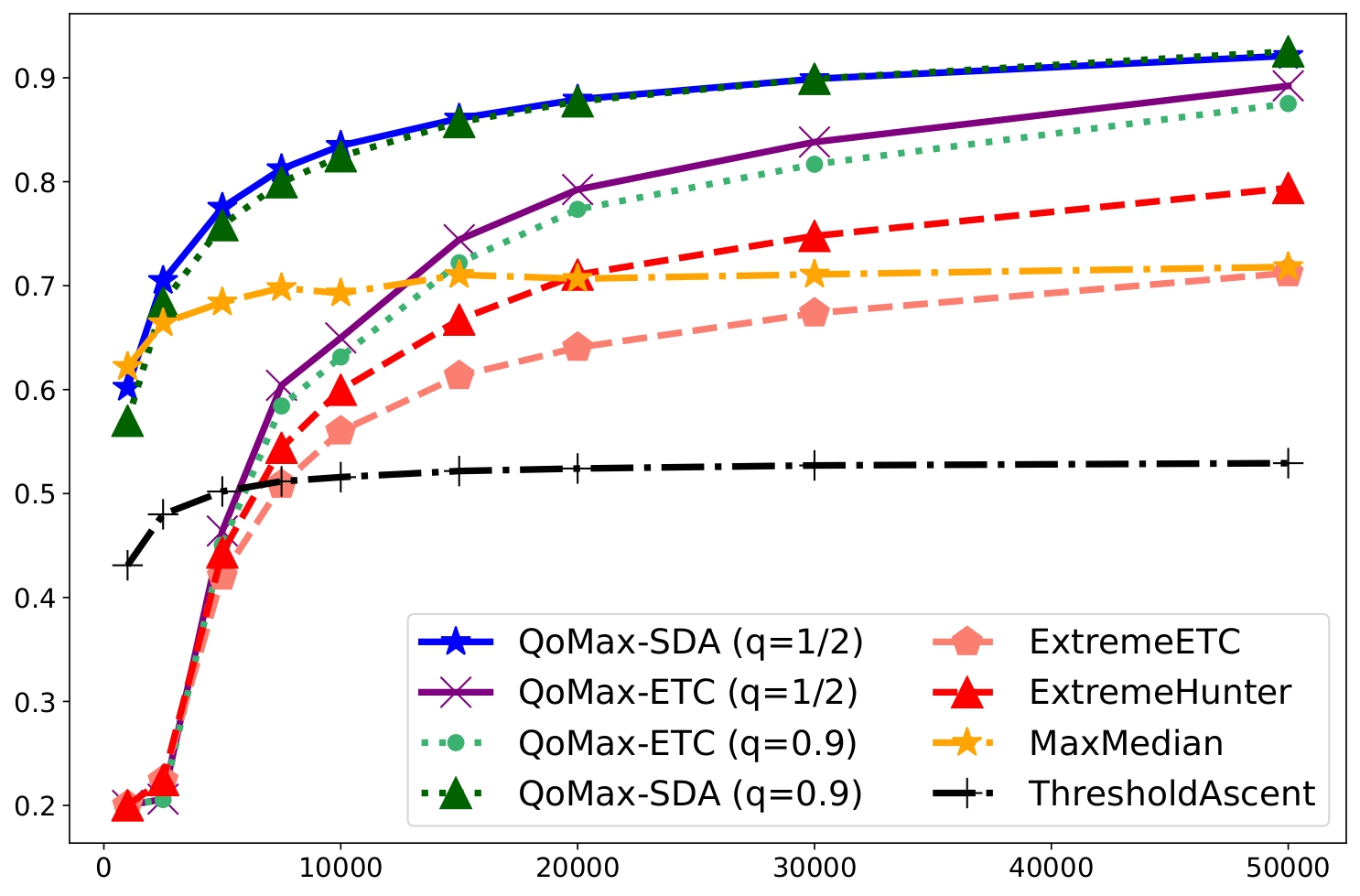}
	\caption{Experiment 1: Proxy Empirical Regret (left) and Number of pulls of the dominant arm (right), averaged over $10^4$ independent trajectories for $T \in \{10^3, 2.5\times 10^3, 5\times 10^3, 7.5\times 10^3, 9\times 10, 10^4, 1.5\times 10^4, 2\times 10^4, 3\times 10^4, 5 \times 10^4 \}$.
}
	\label{fig::xp1_app}
 \vspace{0.25cm}
\end{figure*}

\begin{figure*}[h]
	\centering
	\includegraphics[width=0.45\textwidth]{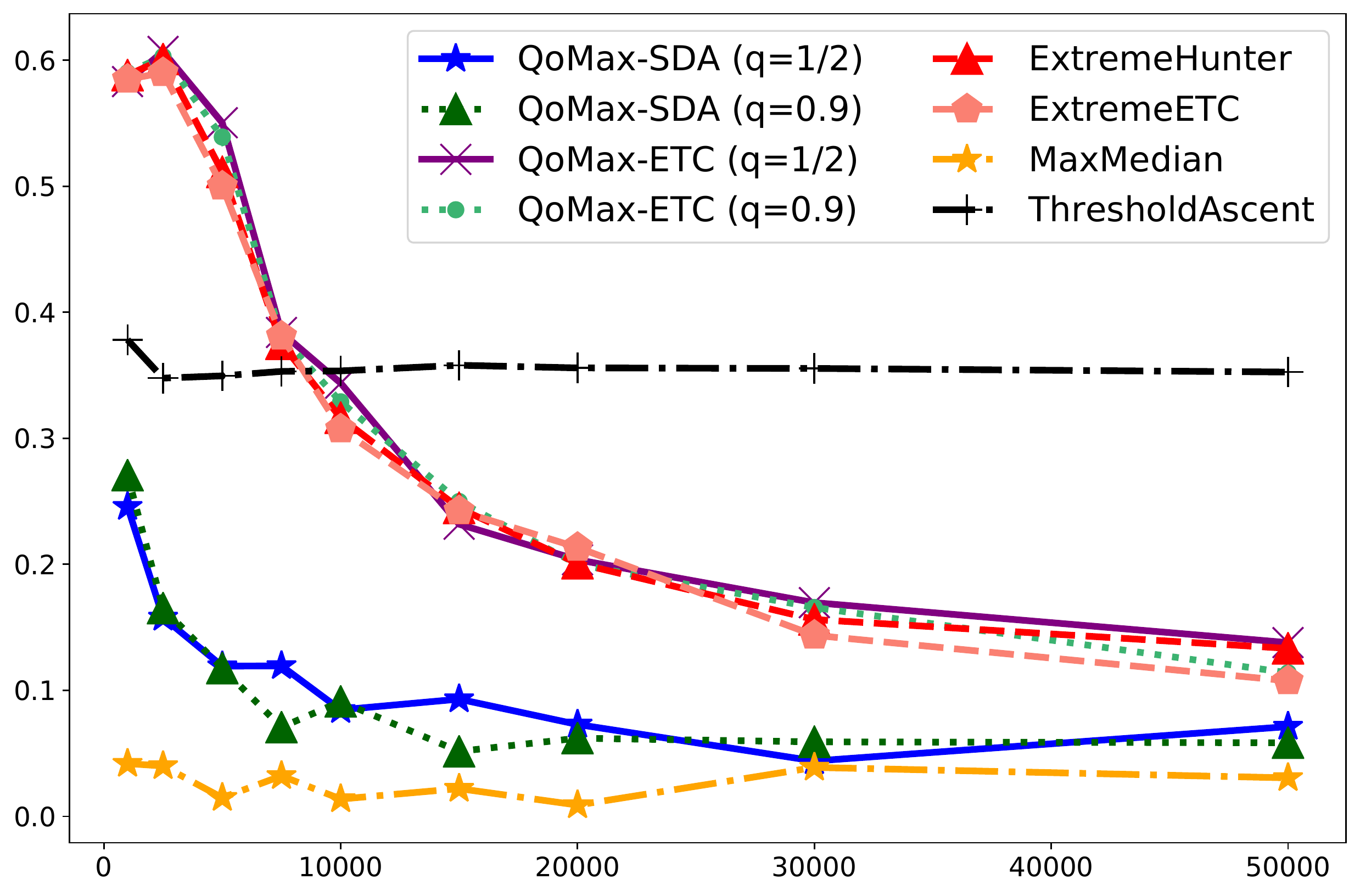} \qquad \qquad 		\includegraphics[width=0.45\textwidth]{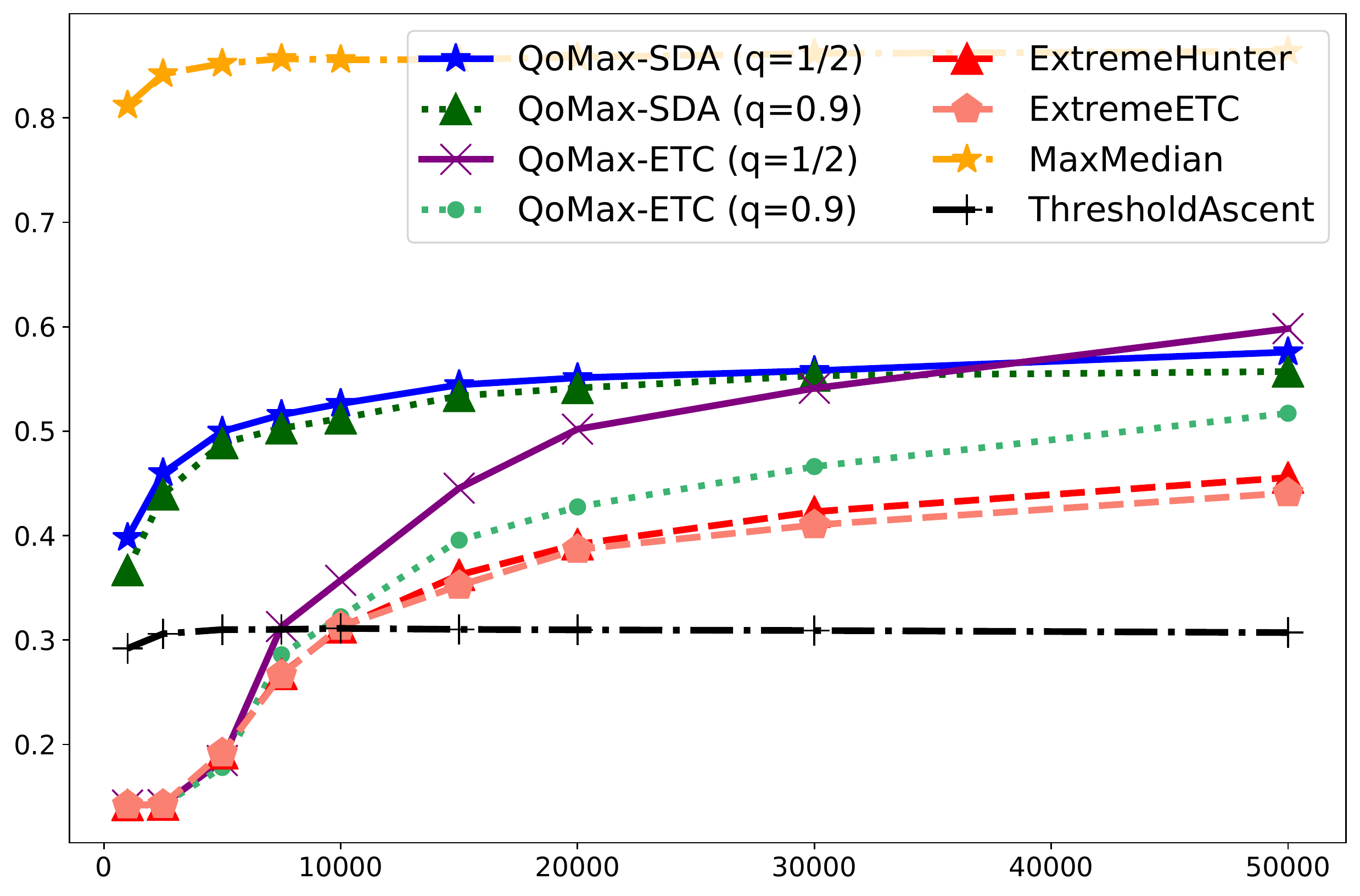}
	\caption{Experiment 2: Proxy Empirical Regret (left) and Number of pulls of the dominant arm (right), averaged over $10^4$ independent trajectories for $T \in \{10^3, 2.5\times 10^3, 5\times 10^3, 7.5\times 10^3, 9\times 10, 10^4, 1.5\times 10^4, 2\times 10^4, 3\times 10^4, 5 \times 10^4 \}$.}
	\label{fig::xp2_app}
	\vspace{0.25cm}
\end{figure*}

\begin{figure*}[h]
	\centering
\includegraphics[width=0.45\textwidth]{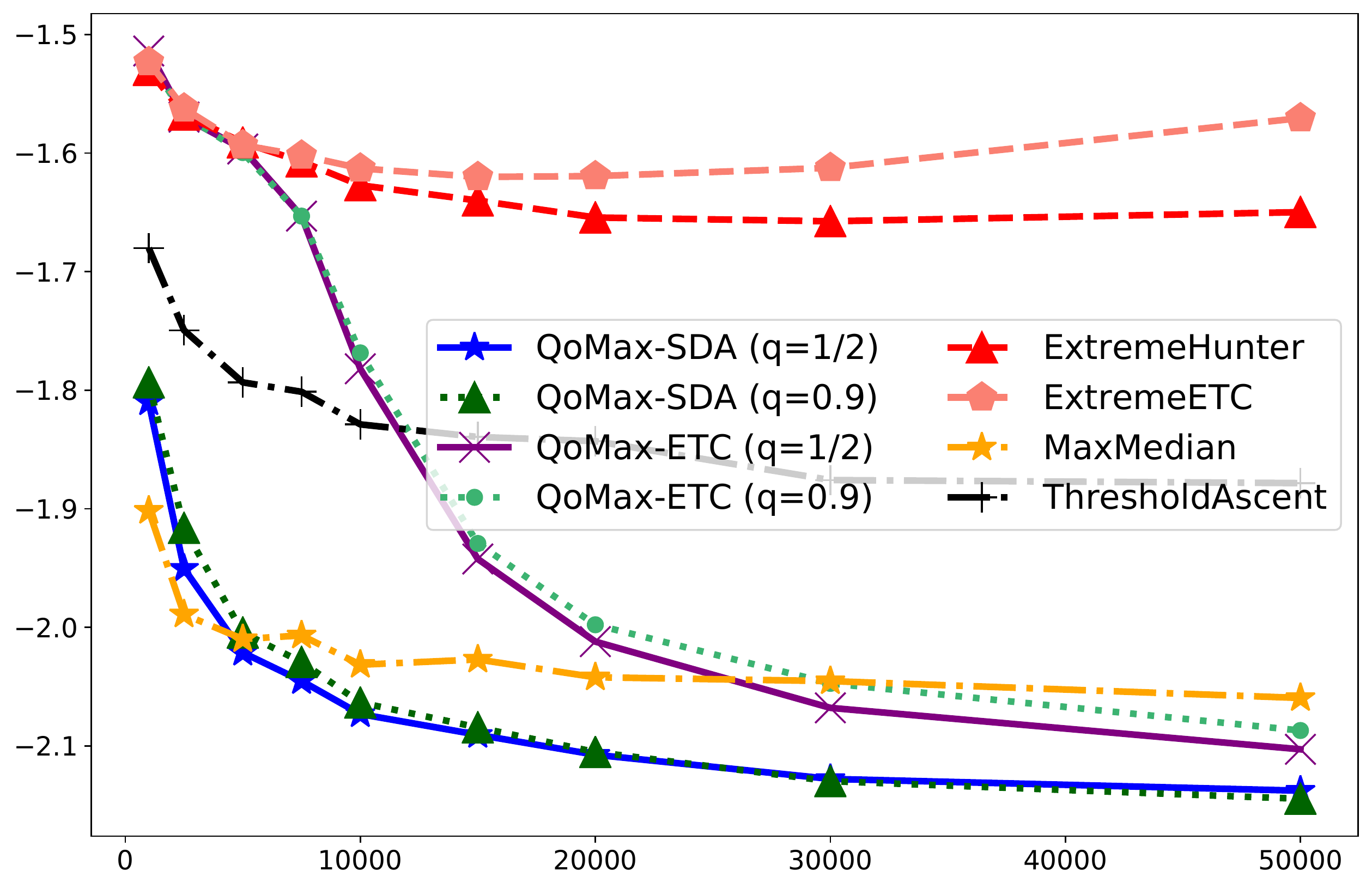} \qquad \qquad \includegraphics[width=0.45\textwidth]{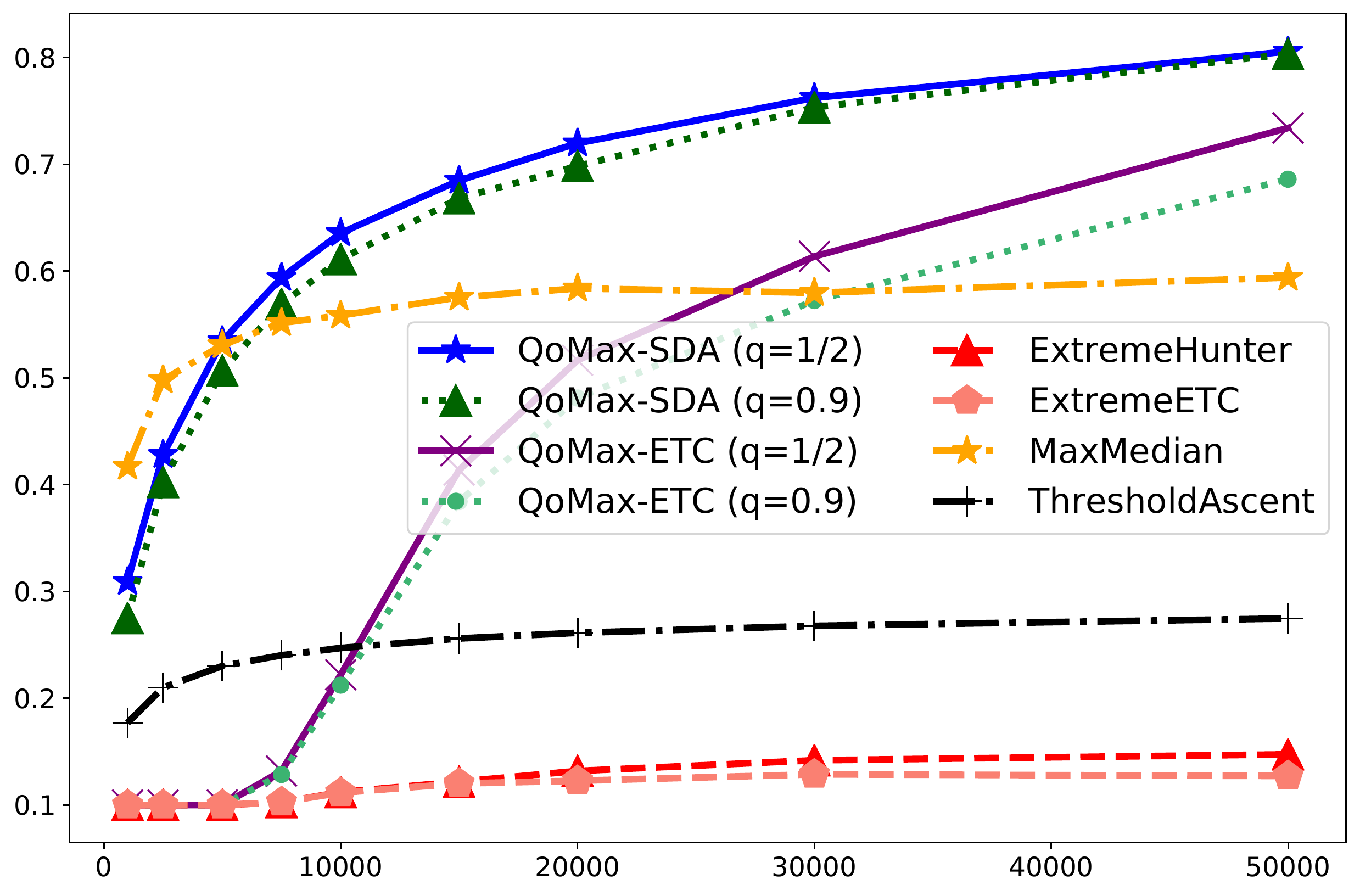}
	\caption{Experiment 3: Proxy Empirical Regret  (left) and Number of pulls of the dominant arm (right), averaged over $10^4$ independent trajectories for $T \in \{10^3, 2.5\times 10^3, 5\times 10^3, 7.5\times 10^3, 9\times 10, 10^4, 1.5\times 10^4, 2\times 10^4, 3\times 10^4, 5 \times 10^4 \}$.}
	\label{fig::xp3_app}
	\vspace{0.25cm}
\end{figure*}

\begin{figure*}[h]
	\centering
	\includegraphics[width=0.45\textwidth]{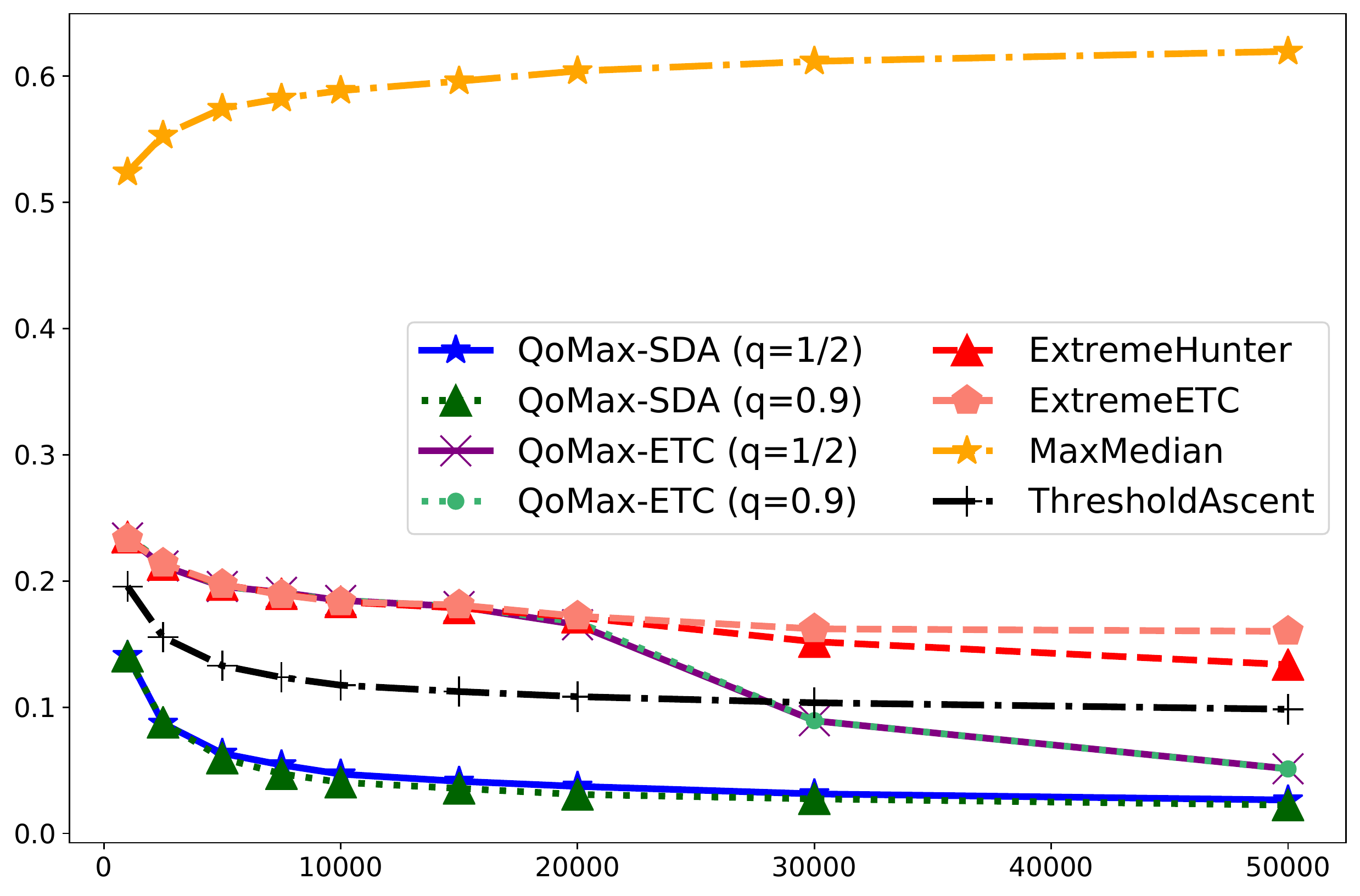} \qquad \qquad \includegraphics[width=0.45\textwidth]{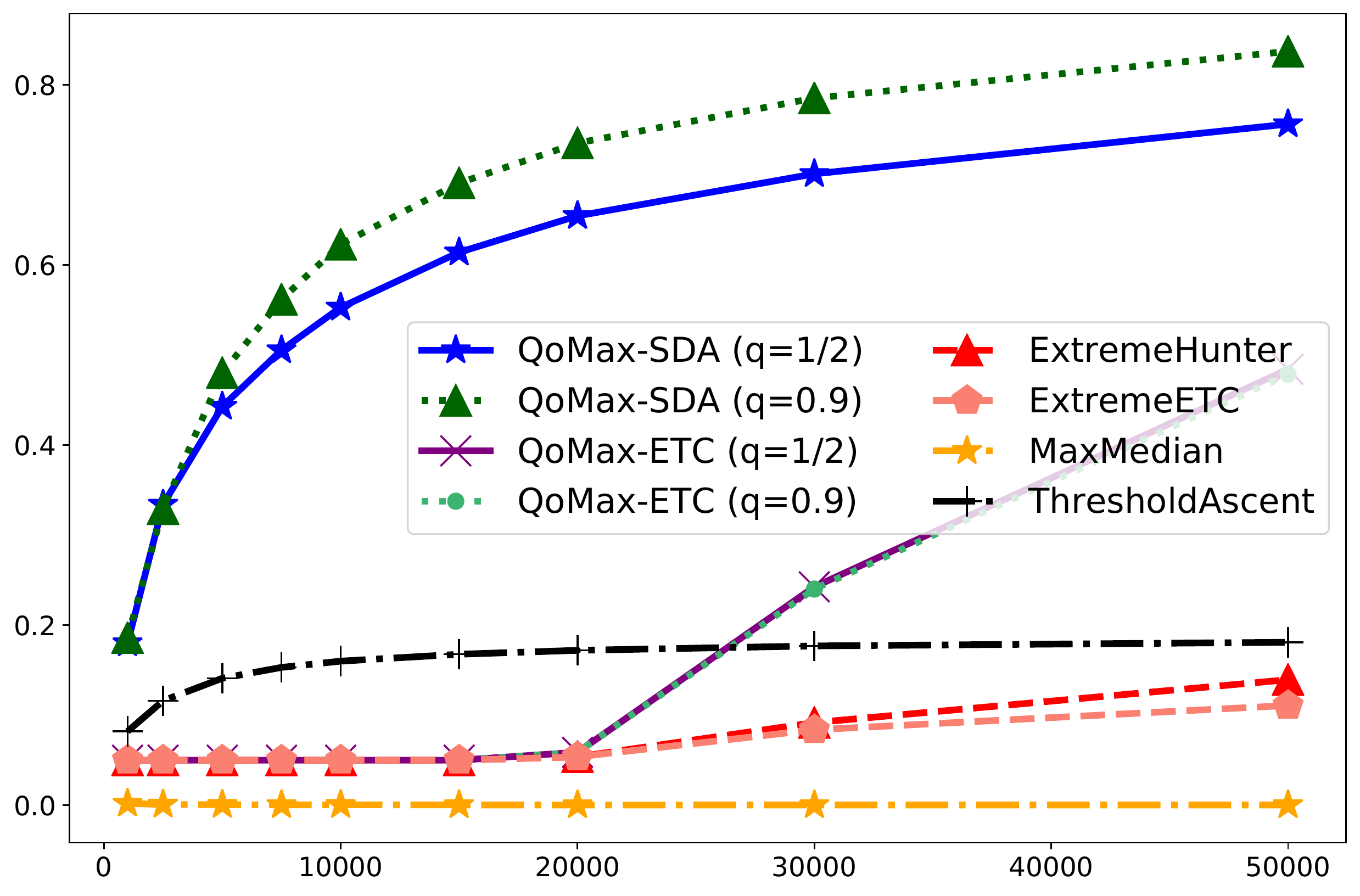}
	\caption{Experiment 4: Proxy Empirical Regret (left) and Number of pulls of the dominant arm (right), averaged over $10^4$ independent trajectories for $T \in \{10^3, 2.5\times 10^3, 5\times 10^3, 7.5\times 10^3, 9\times 10, 10^4, 1.5\times 10^4, 2\times 10^4, 3\times 10^4, 5 \times 10^4 \}$.}
	\label{fig::xp4_app}
	\vspace{0.25cm}
\end{figure*}

\begin{figure*}[h]
	\centering
	\includegraphics[width=0.45\textwidth]{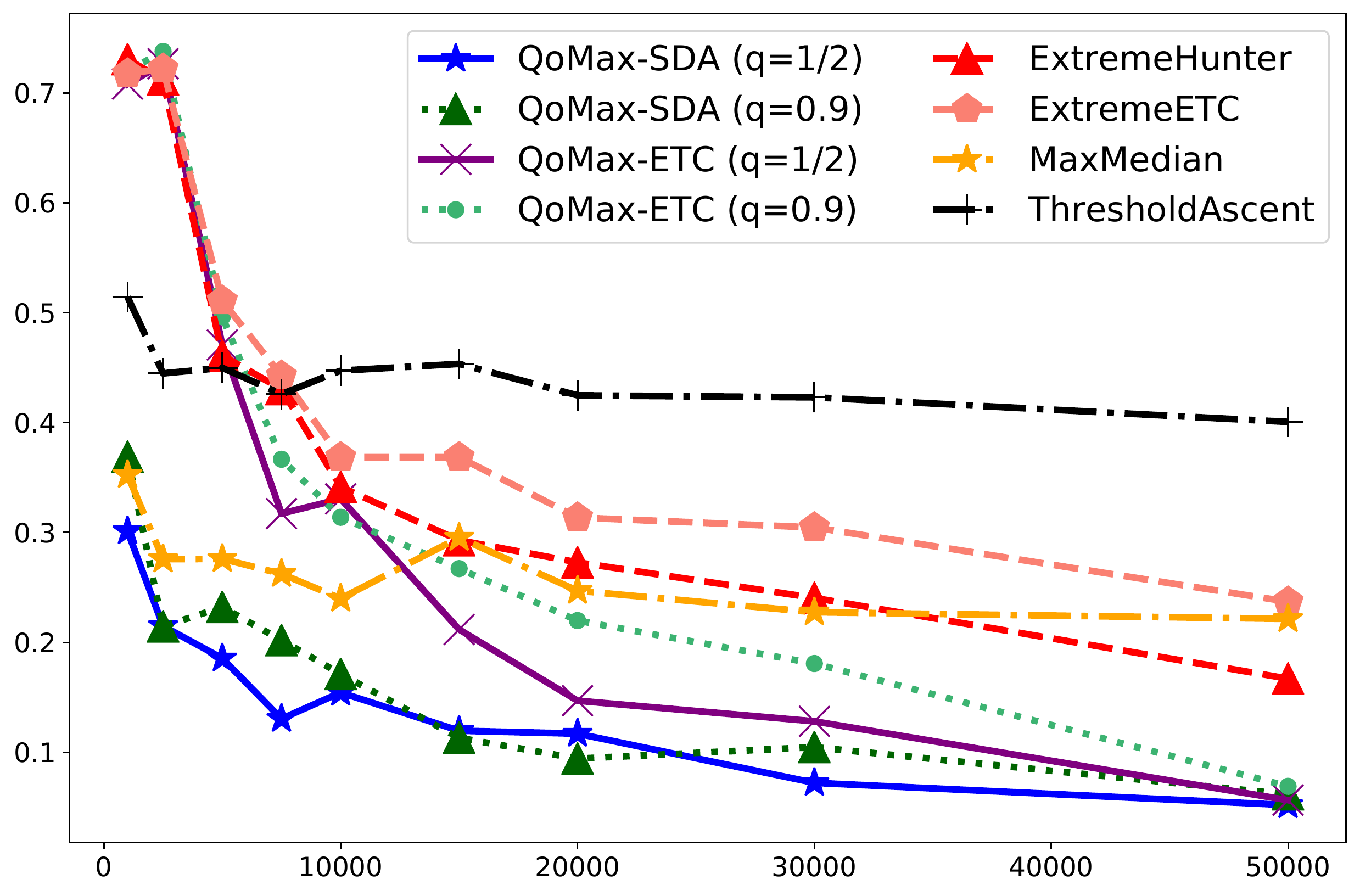} \qquad \qquad \includegraphics[width=0.45\textwidth]{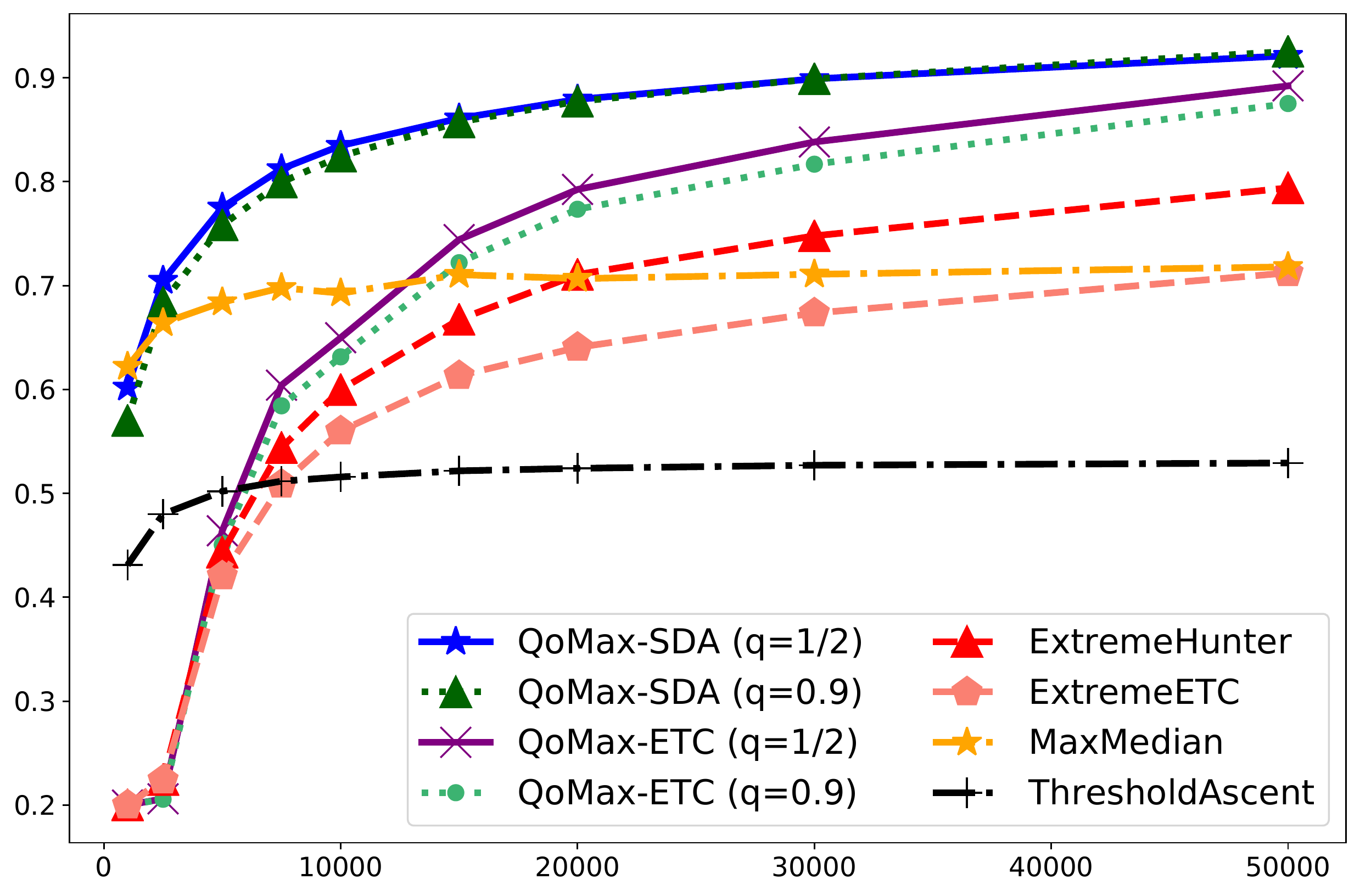}
	\caption{Experiment 5: Proxy Empirical Regret (left) and Number of pulls of the dominant arm (right), averaged over $10^4$ independent trajectories for $T \in \{10^3, 2.5\times 10^3, 5\times 10^3, 7.5\times 10^3, 9\times 10, 10^4, 1.5\times 10^4, 2\times 10^4, 3\times 10^4, 5 \times 10^4 \}$.}
	\label{fig::xp5_app}
	\vspace{0.25cm}
\end{figure*}

\begin{figure*}[h]
	\centering
	\includegraphics[width=0.45\textwidth]{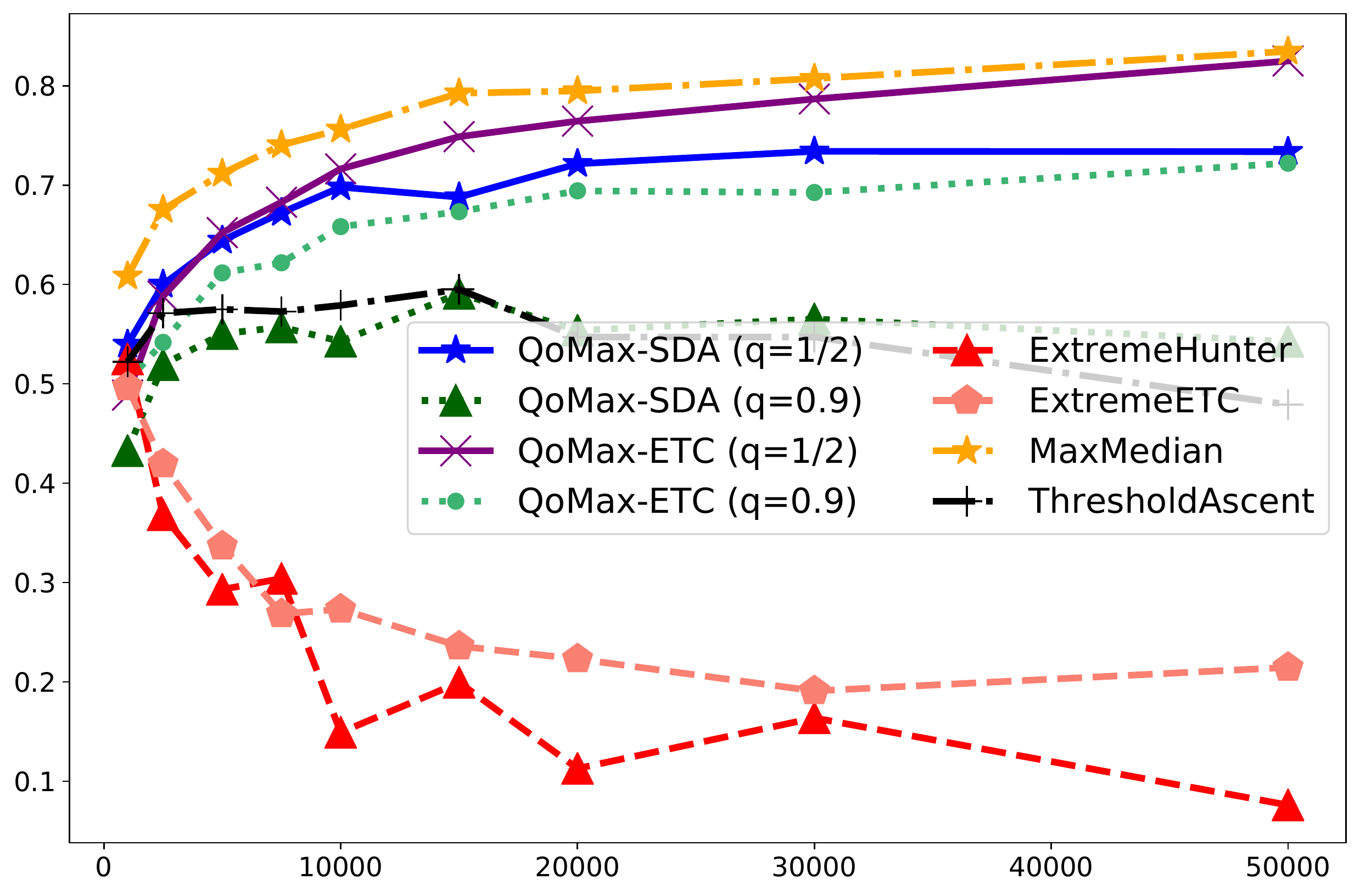} \qquad \qquad \includegraphics[width=0.45\textwidth]{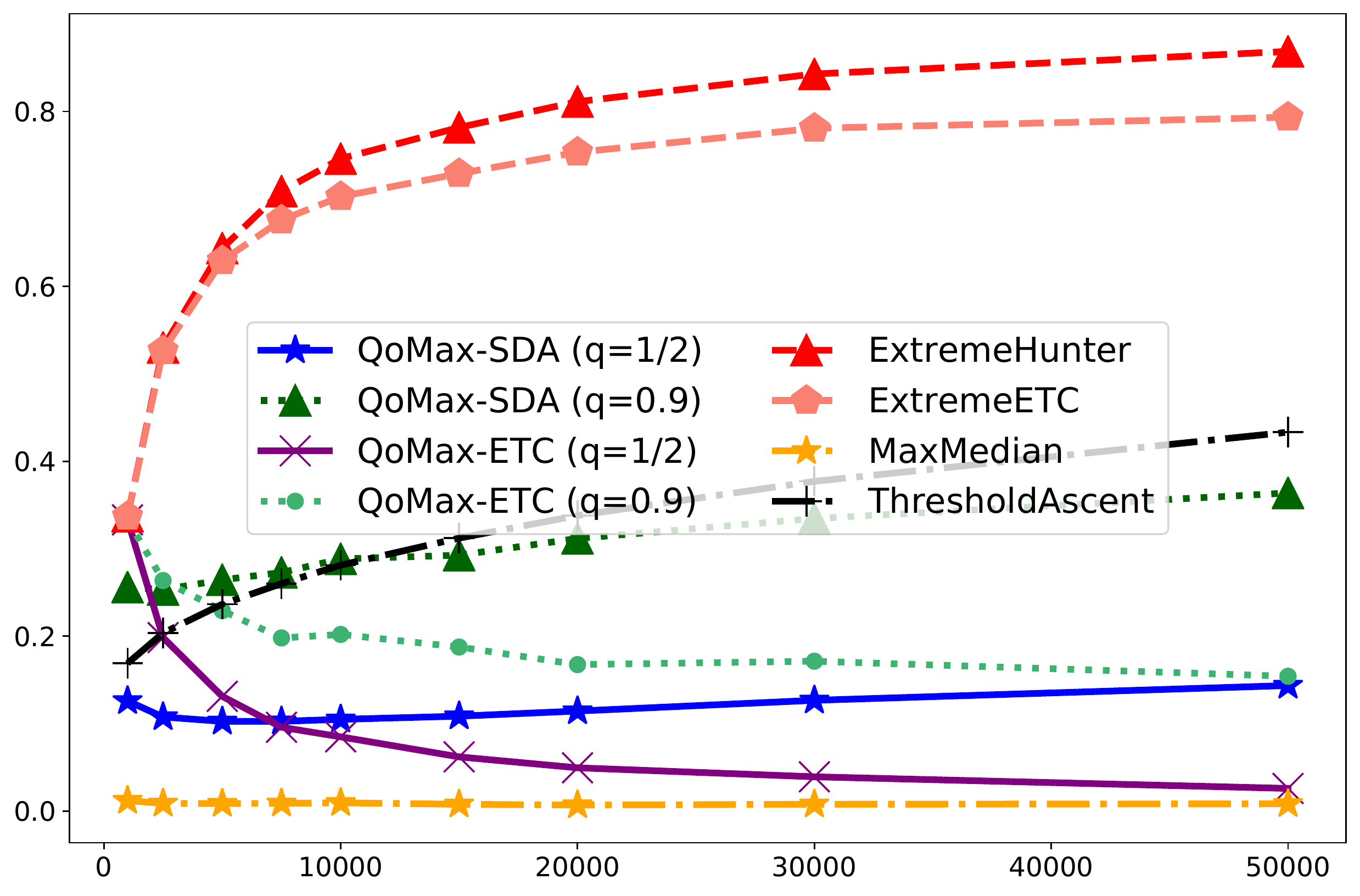}
	\caption{Experiment 6: Proxy Empirical Regret (left) and Number of pulls of the dominant arm (right), averaged over $10^4$ independent trajectories for $T \in \{10^3, 2.5\times 10^3, 5\times 10^3, 7.5\times 10^3, 9\times 10, 10^4, 1.5\times 10^4, 2\times 10^4, 3\times 10^4, 5 \times 10^4 \}$.}
	\label{fig::xp6_app}
	\vspace{0.25cm}
\end{figure*}

\paragraph{Results} For each experiment, we report the results according to the criteria \textbf{(I)-(IV)} that are introduced in Section~\ref{sec::xp}. The criteria \textbf{(I)-(II)} are reported side by side for each experiment in Figures~\ref{fig::xp1_app}-\ref{fig::xp6_app}. Tables~\ref{tab::nb_pulls_xp1}-\ref{tab::nb_pulls_xp6} associated with \textbf{(III)} report the result for the statistics on the number of pulls of the best arm on all trajectories at $T=5\times 10^4$ . Finally, Tables~\ref{tab::maxima_xp1}-\ref{tab::maxima_xp6} related to \textbf{(IV)} report the results for the statistics on the empirical distribution of the maxima on all trajectories at $T=5\times 10^4$.

We summarize our key observations on the results with the following points:
\begin{itemize}
 \item \textbf{On the non-robustness of reporting the average maximum collected.} Several examples can serve to illustrate this point. For experiment 1 (Table~\ref{tab::maxima_xp1}) if we look at the average maximum only, we would conclude that QoMax-SDA with $q=1/2$ is by far the best algorithm with an average of $1.8 \times 10^5$ ($1.1 \times 10^5$ for the second). However, we see that the quantiles of the maxima distributions are almost identical to those of other QoMax algorithms. Hence, even if $99\%$ of their distribution matches, QoMax-SDA with $q=1/2$ has a nearly $70\%$ better average caused by less than $1\%$ of the trajectories. The same thing seems to happen on different problems: the $10^4$ and $8.5 \times 10^3$ of $1/2$-QoMax-ETC and ExtremeETC are clearly over-estimated means in experiment 2 considering that they both have the same quantiles as $1/2$-QoMax-SDA (even a bit worse), which has an average of $7.5 \times 10^3$, and MaxMedian with $7.9 \times 10^3$. This variability is even more striking in Experiment 5 (see Table~\ref{tab::maxima_xp5}) where ExtremeETC has three times the average maximum of ExtremeHunter. Without surprise, this phenomenon is more present when the tails are heavier. Hence looking at the average maxima is meaningful with the statistics from Experiments 3 and 4 with lighter tails.

 \item \textbf{Quantiles.} We recall that for metric 
 \textbf{(I)} we use a quantile to estimate the expectation of the maximum, $\tilde{q} = \bP(X_T^+ \leq \bE[X_T^+]) \approx \exp(-TG(\bE[X_T^+]))$. In the experiments we plug the equivalents of $\bE[X_T^+]$ in each setting: for Pareto distribution we obtain $\tilde{q} =\exp\left(- \frac{1}{\Gamma(1-1/\lambda)^\lambda} \right)$, for exponential we obtain $\tilde{q} = e^{-1}$, and for Gaussian distributions we compute the value numerically (c.f notebook provided with the code).

	\item \textbf{QoMax Performance.} QoMax algorithms clearly outperform their competitors in Experiments \textbf{1, 3, 4} and \textbf{5} according to all criteria. As those experiments include polynomial, exponential and gaussian tails with different number of arms, this shows the generality and efficiency of the QoMax approach. QoMax-SDA seems to work better than QoMax-ETC, in particular it is competitive even for small time horizons ($T<5\times 10^3$) in most experiments. However, we see that QoMax-ETC  almost matches the performance of QoMax-SDA for $T=5\times 10^4$. For a practitioner who would be interested in larger time horizons QoMax-ETC seems to be a perfectly suitable choice.

	\item On the contrary, \textbf{ExtremeHunter} performs significantly better than \textbf{ExtremeETC} for larger horizons: the probability of mistake of the latter is still quite large, and the ability of ExtremeHunter to recover from a mistake is valuable, but we recall that the time complexity of ExtremeHunter is detrimental for the practitioner. Results from Experiments \textbf{3} and \textbf{4} show that the two algorithms are not able to handle exponential and gaussian tails.

	\item \textbf{ThresholdAscent} is never the best algorithm but has the advantage of being consistently better than the uniform strategy (according to \textbf{(II)}), as it always pulls the best arm at a frequency larger that $1/K$.	It is the most stable baseline in terms of \textbf{(III)} (it always has the narrowest range for the statistics we consider), but this is detrimental to its capacity to collect large values.

	\item We tested \textbf{MaxMedian} on larger time horizons than in the original paper, which explains the difference in some results. Indeed, we observe that in Experiments \textbf{1, 3, 5}, MaxMedian is quite competitive for shorter time horizons ($T\leq 10^4$), but almost stops improving at this step. This suggests that the algorithm does not explore enough, which is confirmed by a closer look at \textbf{(III)}: the number of pulls of the best arm are either very close to $0\%$ or to $100\%$ in most of the cases, which is a behavior specific to this algorithm and that we would like to avoid in practice. This behavior also has an impact on the statistics on the maxima distributions \textbf{(IV)}. The exploration function may be partly responsible for this : in Experiment \textbf{4} MaxMedian fails with the Gaussians\footnote{which is not what \cite{bhatt2021extreme} obtained, but we were not able to find why we could not reproduce their results.}, and actually commits to the \textit{worst} arm at an early stage. Indeed, with $20$ arms and $\epsilon_t=1/(t+1)$ we are very likely to have at least one arm that is never sampled twice, while this is the case the order statistics used is always the \textit{minimum}, favoring the arm with the \textit{lowest} variance instead of the largest. We think that a deterministic forced exploration could at least partially solve this.

	\item In \textbf{Experiment 2} MaxMedian performs very well and commits very early to the best arm for most of the trajectories. QoMax algorithms are clearly slower, but still pull the best arm more than $50\%$ of the time. Moreover, when we add the number of pulls of the second best arm we get around $90\%$ for all QoMax algorithms, which is clearly competitive. Indeed, the empirical regret of both QoMax-SDA (Figure~\ref{fig::xp2_app} (left)) is close to the one of MaxMedian, and we see in Table~\ref{tab::maxima_xp2} that their quantiles of the maxima distributions are also very close to the one of MaxMedian. Hence, we think that QoMax-SDA may have chosen more often the second best arm when it provided very large rewards, which is not a problem according to the initial objective of the algorithms.

	 \item \textbf{Experiment 6} shows that in some examples parametric algorithms can perform much better than non-parametric approaches. Indeed, the distribution of arm $3$ enters in the second-order Pareto family, and the parameter $b=1$ makes ExtremeHunter calibrate its parameters with the $\approx 5 \%$ best samples of each arm. This is enough for the algorithm to "detect" the Pareto tail of the mixture and sample it most often. Most of the other algorithms fail, including QoMax, to the exception of ThresholdAscent which still pulls the best arm $40\%$ of the time at $T=5\times 10^4$. However, this experiment also illustrates two important remarks on QoMax: the $0.9$-QoMax-SDA performs much better than the others, showing that when the tails are harder to detect choosing a larger quantile can be valuable. Furthermore, we tested another experiment imposing at least $100$ samples in each batch. This time, $0.9$-QoMax-SDA was able to pull the best arm $60\%$ of the time. Hence, this gives the practitioner the ability to increase the exploration and the quantile $q$ if very difficult tails are expected, which depends on the characteristics of the real problem at hand.
\end{itemize}

Considering all these points, we think that QoMax-ETC and QoMax-SDA are  very practical solutions in addition to their strong theoretical guarantees. They work well on most examples \textbf{with the same parameters} (avoiding painful tuning), including settings with different kind of tails (polynomial, exponential, gaussian) with different number of arms, and both easy and hard instances. We saw however with experiment 6 the limits of a distribution-free approach if we consider a hard problem. 
It also showed that in this case augmenting the quantile $q$ (and/or the forced exploration function $f$ for QoMax-SDA) used in QoMax algorithms can be beneficial.
Furthermore, we can recommend to use QoMax-ETC when the time horizon will be very large (larger than $5\times 10^4$ for instance) and QoMax-SDA for smaller time horizons, as it seems to learn faster on all examples but is more computationally demanding.

\subsubsection{Experiments with Log-Normal and Generalized Gaussian Distributions}

In this section we add two new experiments, considering two new families of distributions: (1) the log-normal distribution (2 parameters $(\mu, \sigma)$, if $X$ follows a log-normal distribution with these parameters then $\log(X) \sim \cN(\mu, \sigma)$), and (2) the generalized normal distribution (a parameter $\beta$ and a density $\sim \exp\left(-|x|^\beta\right)$).
\begin{itemize}
	\item \textbf{Experiment 7:} We consider $K=5$ log-normal arms with parameters $\mu_k \in [1, 1.5, 2, 3, 3.5]$ and $\sigma_k \in [4, 3, 2, 1, 0.5]$. When $T$ is large enough the parameter $\sigma$ determines which arm dominates (arm $1$ in our case).
	\item \textbf{Experiment 8:} We consider $K=8$ generalized gaussian arms with parameters $\beta_k \in (0.2 \times i)_{i \in \{1, \dots, 8 \}}$. Hence, the heavier tail is arm $1$.
\end{itemize}

\begin{figure*}[hbt]
	\centering
	\includegraphics[width=0.45\textwidth]{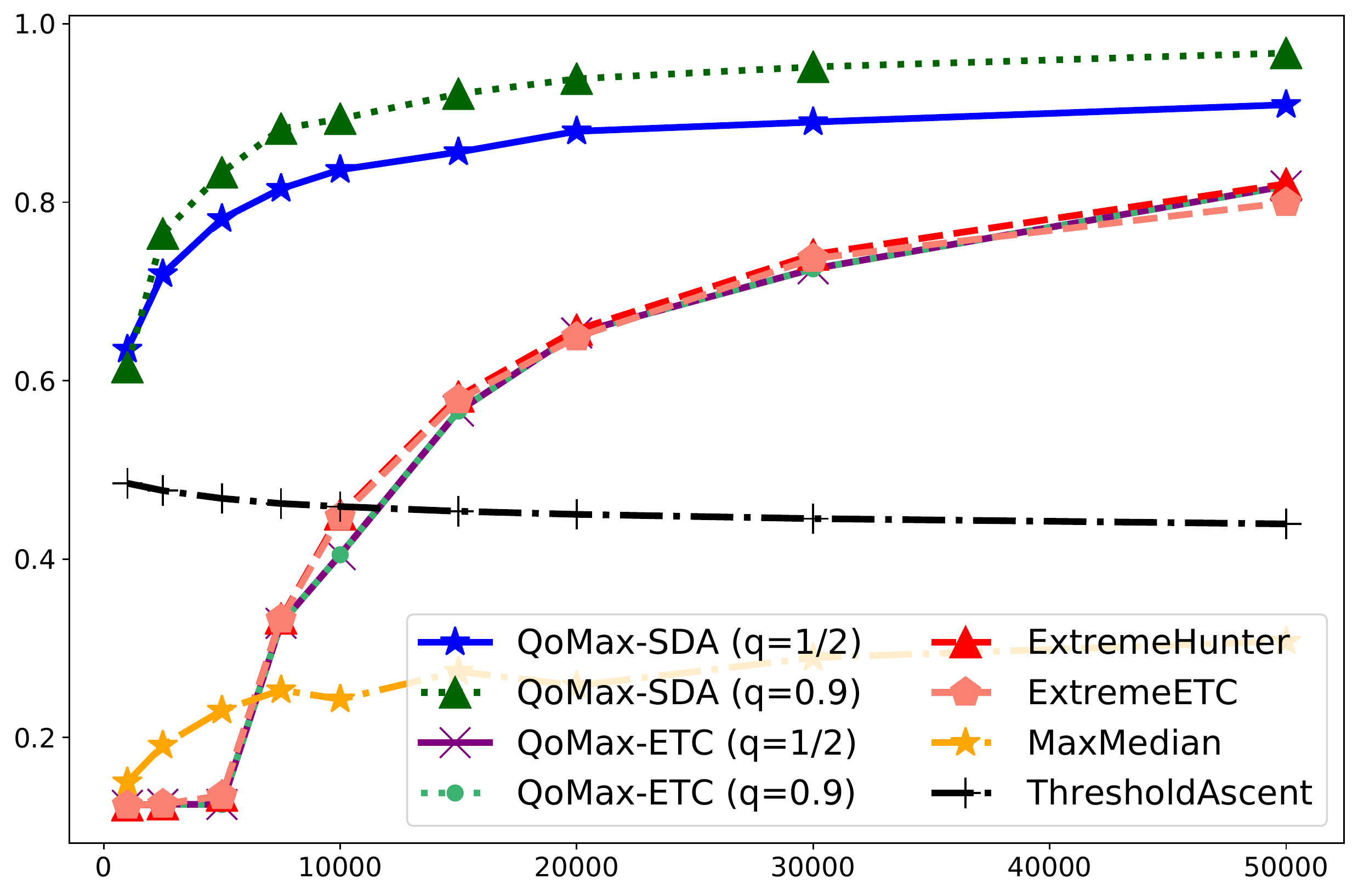}
	\caption{Experiment 7 (Log-normal arms): Number of pulls of the dominant arm, averaged over $10^4$ independent trajectories for $T \in \{10^3, 2.5\times 10^3, 5\times 10^3, 7.5\times 10^3, 9\times 10, 10^4, 1.5\times 10^4, 2\times 10^4, 3\times 10^4, 5 \times 10^4 \}$.}
	\label{fig::new_xp_log_normal}
\end{figure*}

\begin{figure*}[hbt]
	\centering
\includegraphics[width=0.45\textwidth]{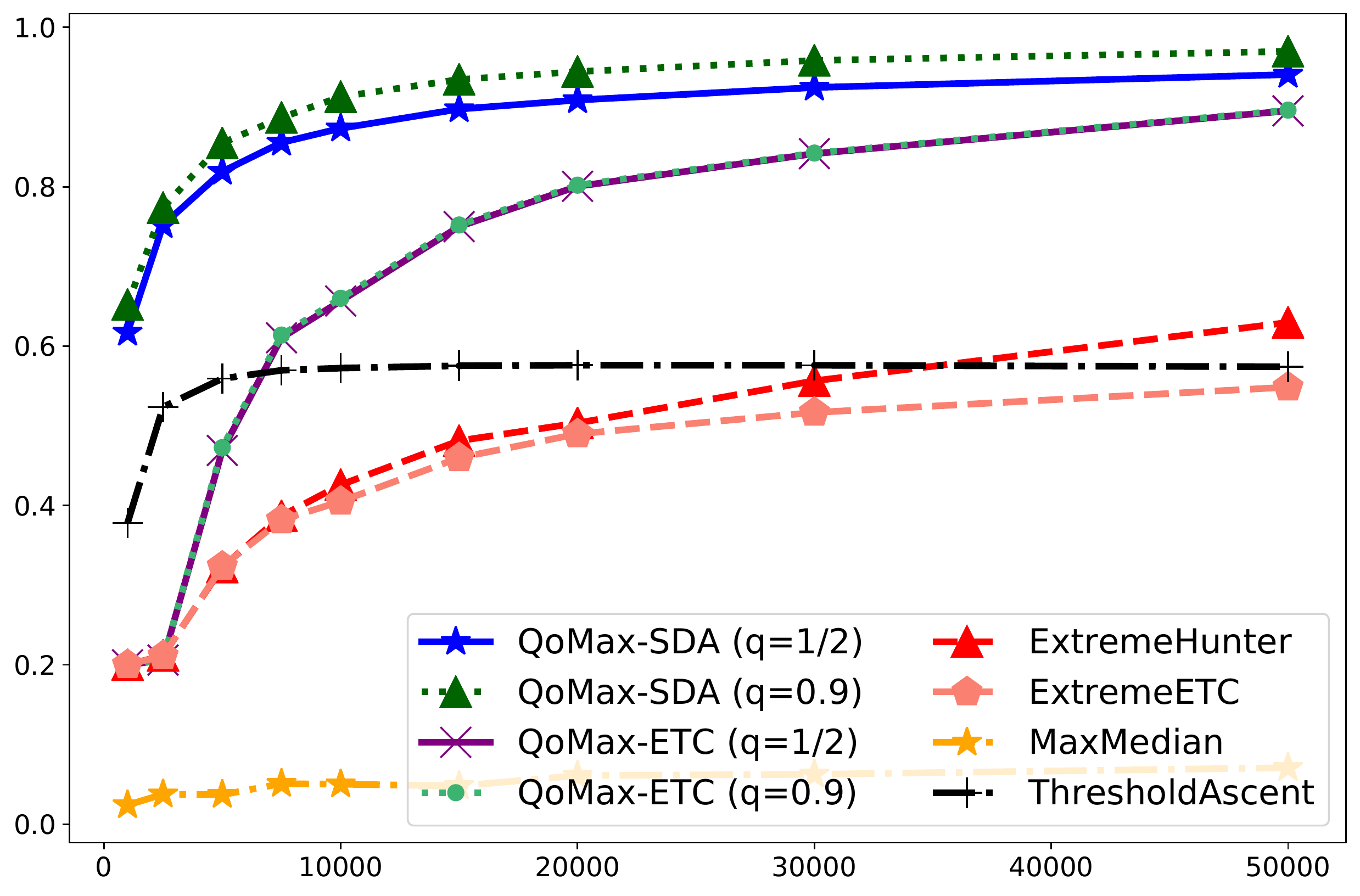}
	\caption{Experiment 8 (Generalized Gaussian arms): Number of pulls of the dominant arm, averaged over $10^4$ independent trajectories for $T \in \{10^3, 2.5\times 10^3, 5\times 10^3, 7.5\times 10^3, 9\times 10, 10^4, 1.5\times 10^4, 2\times 10^4, 3\times 10^4, 5 \times 10^4 \}$.}
	\label{fig::new_xp_generalized_gaussians}
	\vspace{0.6cm}
\end{figure*}

We run the same algorithms as for Experiments 1-6, with the exact same parameters for all of them. This time we cannot report \textbf{(I)} because we cannot compute the proxy empirical regret. Hence, we report in
Figure~\ref{fig::new_xp_log_normal} and Figure~\ref{fig::new_xp_generalized_gaussians} the number of pulls of the dominant arm for the two experiments, along with the statistics corresponding to evaluation criteria \textbf{(III)-(IV)} for these experiments. These two additional experiments further highlight
the generality and performance of QoMax algorithms compared to the other Extreme Bandits baselines.

%
%
%
\clearpage
\newpage
\subsection*{Experiment 1}

\begin{table}[hbt]
	\begin{center}
		\begin{small}
		\caption{Statistics on the number of pulls of the best arm at $T=5\times 10^4$, Experiment 1.}
			\begin{tabular}{c|c|ccccccc}
				\toprule Algorithm & Average ($\%$) & 1\% &  10\% &  25\% &  50\% &  75\% &  90\% & 99\% \\
				\midrule
				QoMax-SDA ($q=1/2$) &       92 &   42 &    90 &    93 &    94 &    95 &    95 &    95 \\
				QoMax-SDA ($q=0.9$) &       93 &   14 &    87 &    93 &    96 &    97 &    98 &    98 \\
				QoMax-ETC ($q=1/2$) &       89 &   90 &    90 &    90 &    90 &    90 &    90 &    90 \\
				QoMax-ETC ($q=0.9$) &       88 &    3 &    90 &    90 &    90 &    90 &    90 &    90 \\
				ExtremeETC          &       71 &    3 &     3 &    90 &    90 &    90 &    90 &    90 \\
				ExtremeHunter       &       79 &    3 &     5 &    89 &    90 &    90 &    90 &    90 \\
				MaxMedian           &       72 &    0 &     0 &     0 &   100 &   100 &   100 &   100 \\
				ThresholdAscent     &       53 &   46 &    50 &    52 &    53 &    55 &    56 &    57 \\
				\bottomrule
			\end{tabular}
			\label{tab::nb_pulls_xp1}
		\end{small}
\vspace{0.8cm}
\begin{small}
\caption{Statistics on the distributions of maxima at $T=5\times 10^4$, Experiment 1. Results divided by $100$ to improve readability.}
			\begin{tabular}{c|c|ccccccc}
				\toprule Algorithm & Average  & 1\% &  10\% &  25\% &  50\% &  75\% &  90\% & 99\% \\
				\midrule
				QoMax-SDA ($q=1/2$) &     1852 &   41 &    81 &   130 &   245 &   547 &  1350 & 11371 \\
				QoMax-SDA ($q=0.9$) &     1042 &   39 &    78 &   128 &   239 &   529 &  1363 & 12539 \\
				QoMax-ETC ($q=1/2$) &     1058 &   40 &    79 &   126 &   232 &   530 &  1324 & 11054 \\
				QoMax-ETC ($q=0.9$) &      919 &   34 &    75 &   122 &   230 &   511 &  1301 & 10080 \\
				ExtremeETC          &      882 &   16 &    44 &    86 &   183 &   426 &  1089 &  9515 \\
				ExtremeHunter       &     1092 &   21 &    61 &   104 &   208 &   477 &  1226 &  9799 \\
				MaxMedian           &      785 &    3 &    37 &    83 &   180 &   436 &  1126 &  9240 \\
				ThresholdAscent     &      748 &   27 &    51 &    82 &   156 &   351 &   853 &  7771 \\
				\bottomrule
			\end{tabular}
			\label{tab::maxima_xp1}
		\end{small}
	\end{center}
\end{table}

\clearpage
\subsection*{Experiment 2}

\begin{table}[hbt]
		\begin{center}
		\begin{small}
		\caption{Statistics on the number of pulls of the best arm at $T=5\times 10^4$, Experiment 2.}
			\begin{tabular}{c|c|ccccccc}
				\toprule Algorithm & Average ($\%$)& 1\% &  10\% &  25\% &  50\% &  75\% &  90\% & 99\% \\
				\midrule
				QoMax-SDA ($q=1/2$) &       58 &    2 &     6 &    23 &    72 &    88 &    91 &    92 \\
				QoMax-SDA ($q=0.9$) &       56 &    1 &     6 &    21 &    66 &    88 &    94 &    97 \\
				QoMax-ETC ($q=1/2$) &       60 &    3 &     3 &     3 &    84 &    84 &    84 &    84 \\
				QoMax-ETC ($q=0.9$) &       52 &    3 &     3 &     3 &    84 &    84 &    84 &    84 \\
				ExtremeETC          &       44 &    3 &     3 &     3 &    85 &    85 &    85 &    85 \\
				ExtremeHunter       &       46 &    3 &     3 &     3 &    71 &    85 &    85 &    85 \\
				MaxMedian           &       86 &    0 &     0 &   100 &   100 &   100 &   100 &   100 \\
				ThresholdAscent     &       31 &   20 &    25 &    28 &    31 &    34 &    36 &    40 \\
				\bottomrule
			\end{tabular}
\label{tab::nb_pulls_xp2}	
		\end{small}
	\end{center}
	\vspace{0.6cm}
\begin{center}
		\begin{small}
		\caption{Statistics on the distributions of maxima at $T=5\times 10^4$, Experiment 2. Results divided by $100$ to improve readability.}
			\begin{tabular}{c|c|ccccccc}
				\toprule Algorithm & Average & 1\% &  10\% &  25\% &  50\% &  75\% &  90\% & 99\% \\
				\midrule
				QoMax-SDA ($q=1/2$) &       75 &    8 &    13 &    18 &    30 &    56 &   112 &   657 \\
				QoMax-SDA ($q=0.9$) &       69 &    8 &    13 &    18 &    30 &    56 &   113 &   614 \\
				QoMax-ETC ($q=1/2$) &       98 &    7 &    12 &    17 &    28 &    51 &   105 &   616 \\
				QoMax-ETC ($q=0.9$) &       65 &    7 &    12 &    17 &    28 &    52 &   107 &   564 \\
				ExtremeETC          &       85 &    5 &    12 &    17 &    28 &    53 &   108 &   638 \\
				ExtremeHunter       &       61 &    7 &    12 &    17 &    28 &    52 &   100 &   522 \\
				MaxMedian           &       79 &    6 &    13 &    19 &    31 &    57 &   116 &   664 \\
				ThresholdAscent     &       46 &    5 &     9 &    13 &    21 &    38 &    77 &   418 \\
				\bottomrule
			\end{tabular}
			 \label{tab::maxima_xp2}
		\end{small}
	\end{center}
\end{table}

\clearpage
\subsection*{Experiment 3}

\begin{table}[hbt]
\begin{center}
		\begin{small}
		\caption{Statistics on the number of pulls of the best arm at $T=5\times 10^4$, Experiment 3. }
			\begin{tabular}{c|c|ccccccc}
				\toprule Algorithm & Average ($\%$)& 1\% &  10\% &  25\% &  50\% &  75\% &  90\% & 99\% \\
				\midrule
				QoMax-SDA ($q=1/2$) &       81 &    2 &    72 &    82 &    86 &    88 &    88 &    89 \\
				QoMax-SDA ($q=0.9$) &       80 &    2 &    59 &    80 &    87 &    91 &    93 &    95 \\
				QoMax-ETC ($q=1/2$) &       73 &    3 &    77 &    77 &    77 &    77 &    77 &    77 \\
				QoMax-ETC ($q=0.9$) &       69 &    3 &     3 &    77 &    77 &    77 &    77 &    77 \\
				ExtremeETC          &       13 &    3 &     3 &     3 &     3 &     3 &    77 &    77 \\
				ExtremeHunter       &       15 &    3 &     3 &     3 &     3 &     7 &    67 &    77 \\
				MaxMedian           &       59 &    0 &     0 &     0 &    98 &   100 &   100 &   100 \\
				ThresholdAscent     &       27 &   21 &    24 &    26 &    28 &    29 &    31 &    33 \\
				\bottomrule
			\end{tabular}
			\label{tab::nb_pulls_xp3}
		\end{small}
	\end{center}
	\vspace{0.6cm}
	\begin{center}
		\begin{small}
		\caption{Statistics on the distributions of maxima at $T=5\times 10^4$, Experiment 3.}
			\begin{tabular}{c|c|ccccccc}
				\toprule Algorithm & Average & 1\% &  10\% &  25\% &  50\% &  75\% &  90\% & 99\% \\
				\midrule
				QoMax-SDA ($q=1/2$) &       32 &   26 &    28 &    29 &    31 &    34 &    37 &    43 \\
				QoMax-SDA ($q=0.9$) &       32 &   25 &    28 &    29 &    31 &    34 &    37 &    44 \\
				QoMax-ETC ($q=1/2$) &       32 &   25 &    28 &    29 &    31 &    34 &    36 &    43 \\
				QoMax-ETC ($q=0.9$) &       31 &   24 &    27 &    29 &    31 &    33 &    36 &    43 \\
				ExtremeETC          &       26 &   18 &    21 &    23 &    25 &    29 &    32 &    39 \\
				ExtremeHunter       &       27 &   19 &    22 &    23 &    26 &    29 &    32 &    39 \\
				MaxMedian           &       31 &   21 &    25 &    28 &    31 &    33 &    36 &    43 \\
				ThresholdAscent     &       29 &   23 &    25 &    27 &    29 &    31 &    34 &    41 \\
				\bottomrule
			\end{tabular}
			\label{tab::maxima_xp3}
		\end{small}
	\end{center}
\end{table}

\clearpage
\subsection*{Experiment 4}

\begin{table}[h]
	\begin{center}
		\begin{small}
		\caption{Statistics on the number of pulls of the best arm at $T=5\times 10^4$, Experiment 4.}
			\begin{tabular}{c|c|ccccccc}
				\toprule Algorithm & Average ($\%$)& 1\% &  10\% &  25\% &  50\% &  75\% &  90\% & 99\% \\
				\midrule
				QoMax-SDA ($q=1/2$) &       76 &    4 &    74 &    77 &    78 &    79 &    80 &    80 \\
				QoMax-SDA ($q=0.9$) &       84 &    3 &    75 &    85 &    89 &    90 &    91 &    91 \\
				QoMax-ETC ($q=1/2$) &       48 &    3 &    51 &    51 &    51 &    51 &    51 &    51 \\
				QoMax-ETC ($q=0.9$) &       48 &    3 &    51 &    51 &    51 &    51 &    51 &    51 \\
				ExtremeETC          &       11 &    3 &     3 &     3 &     3 &     3 &    52 &    52 \\
				ExtremeHunter       &       14 &    3 &     3 &     3 &     3 &    25 &    48 &    52 \\
				MaxMedian           &        0 &    0 &     0 &     0 &     0 &     0 &     0 &     0 \\
				ThresholdAscent     &       18 &   15 &    16 &    17 &    18 &    19 &    20 &    20 \\ \bottomrule
			\end{tabular}
			\label{tab::nb_pulls_xp4}
		\end{small}
			\end{center}
		\vspace{1cm}

\begin{center}
		\begin{small}
		\caption{Statistics on the distributions of maxima at $T=5\times 10^4$, Experiment 4.}
			\begin{tabular}{c|c|ccccccc}
				\toprule Algorithm & Average & 1\% &  10\% &  25\% &  50\% &  75\% &  90\% & 99\% \\
				\midrule
				QoMax-SDA ($q=1/2$) &       14 &   13 &    13 &    14 &    14 &    15 &    16 &    17 \\
				QoMax-SDA ($q=0.9$) &       15 &   13 &    13 &    14 &    14 &    15 &    16 &    17 \\
				QoMax-ETC ($q=1/2$) &       14 &   12 &    13 &    13 &    14 &    15 &    15 &    17 \\
				QoMax-ETC ($q=0.9$) &       14 &   12 &    13 &    13 &    14 &    15 &    15 &    17 \\
				ExtremeETC          &       12 &   10 &    11 &    11 &    12 &    13 &    14 &    16 \\
				ExtremeHunter       &       13 &   10 &    11 &    12 &    13 &    14 &    14 &    16 \\
				MaxMedian           &        5 &    3 &     4 &     4 &     5 &     6 &     7 &    10 \\
				ThresholdAscent     &       13 &   12 &    12 &    13 &    13 &    14 &    15 &    16 \\
				\bottomrule
			\end{tabular}
			 \label{tab::maxima_xp4}
		\end{small}
		\end{center}
\end{table}

\clearpage
\subsection*{Experiment 5}

\begin{table}[h]
		\begin{center}
			\begin{small}
			\caption{Statistics on the number of pulls of the best arm at $T=5\times 10^4$, Experiment 5.}
				\begin{tabular}{c|c|ccccccc}
					\toprule Algorithm & Average ($\%$)& 1\% &  10\% &  25\% &  50\% &  75\% &  90\% & 99\% \\
					\midrule
					QoMax-SDA ($q=1/2$) &       97 &   97 &    97 &    97 &    97 &    97 &    97 &    97 \\
					QoMax-SDA ($q=0.9$) &       99 &   98 &    99 &    99 &    99 &    99 &    99 &    99 \\
					QoMax-ETC ($q=1/2$) &       95 &   95 &    95 &    95 &    95 &    95 &    95 &    95 \\
					QoMax-ETC ($q=0.9$) &       95 &   95 &    95 &    95 &    95 &    95 &    95 &    95 \\
					ExtremeETC        &       95 &   95 &    95 &    95 &    95 &    95 &    95 &    95 \\
					ExtremeHunter     &       95 &   95 &    95 &    95 &    95 &    95 &    95 &    95 \\
					MaxMedian         &       95 &    0 &   100 &   100 &   100 &   100 &   100 &   100 \\
					ThresholdAscent   &       74 &   73 &    73 &    74 &    74 &    74 &    74 &    74 \\
					\bottomrule
				\end{tabular}
				\label{tab::nb_pulls_xp5}
			\end{small}
	\end{center}
	\vspace{0.5cm}
	\begin{center}
			\begin{small}
			\caption{Statistics on the distributions of maxima at $T=5\times 10^4$, Experiment 5. Results divided by $100$ to improve readability.}
				\begin{tabular}{c|c|ccccccc}
					\toprule Algorithm & Average & 1\% &  10\% &  25\% &  50\% &  75\% &  90\% & 99\% \\
					\midrule
					QoMax-SDA ($q=1/2$) &     1179 &   46 &    84 &   133 &   251 &   556 &  1405 & 11239 \\
					QoMax-SDA ($q=0.9$) &     1325 &   47 &    88 &   140 &   267 &   582 &  1444 & 12836 \\
					QoMax-ETC ($q=1/2$) &     1055 &   45 &    84 &   134 &   250 &   565 &  1347 & 11434 \\
					QoMax-ETC ($q=0.9$) &      944 &   43 &    82 &   133 &   247 &   547 &  1395 & 11038 \\
					ExtremeETC          &     3428 &   42 &    83 &   132 &   245 &   542 &  1362 &  9944 \\
					ExtremeHunter       &      910 &   44 &    83 &   132 &   241 &   553 &  1386 & 11555 \\
					MaxMedian           &      939 &    2 &    70 &   124 &   240 &   548 &  1378 & 10239 \\
					ThresholdAscent     &     1096 &   35 &    67 &   107 &   200 &   445 &  1151 &  9105 \\
					\bottomrule
				\end{tabular}
				 \label{tab::maxima_xp5}
			\end{small}
		\end{center}
\end{table}

\clearpage
\subsection*{Experiment 6}

\begin{table}[h]
		\begin{center}
			\begin{small}
			\caption{Statistics on the number of pulls of the best arm at $T=5\times 10^4$, Experiment 6.}
				\begin{tabular}{c|c|ccccccc}
					\toprule Algorithm & Average ($\%$)& 1\% &  10\% &  25\% &  50\% &  75\% &  90\% & 99\% \\
					\midrule
					QoMax-SDA ($q=1/2$) &       14 &    1 &     1 &     2 &     4 &    15 &    45 &    95 \\
					QoMax-SDA ($q=0.9$) &       36 &    0 &     1 &     3 &    22 &    75 &    90 &    98 \\
					QoMax-ETC ($q=1/2$) &        3 &    3 &     3 &     3 &     3 &     3 &     3 &     3 \\
					QoMax-ETC ($q=0.9$) &       15 &    3 &     3 &     3 &     3 &     3 &    95 &    95 \\
					ExtremeETC          &       79 &    3 &     3 &    95 &    95 &    95 &    95 &    95 \\
					ExtremeHunter       &       87 &    3 &    85 &    95 &    95 &    95 &    95 &    95 \\
					MaxMedian           &        1 &    0 &     0 &     0 &     0 &     0 &     0 &     5 \\
					ThresholdAscent     &       43 &   27 &    34 &    38 &    43 &    49 &    53 &    60 \\
					\bottomrule
				\end{tabular}
				\label{tab::nb_pulls_xp6}
			\end{small}
	\end{center}
	\vspace{0.5cm}
	\begin{center}
			\begin{small}
			\caption{Statistics on the distributions of maxima at $T=5\times 10^4$, Experiment 6. Results divided by $100$ to improve readability.}
				\begin{tabular}{c|c|ccccccc}
					\toprule Algorithm & Average & 1\% &  10\% &  25\% &  50\% &  75\% &  90\% & 99\% \\
					\midrule
					QoMax-SDA ($q=1/2$) &       60 &    5 &     9 &    12 &    21 &    41 &    91 &   635 \\
					QoMax-SDA ($q=0.9$) &      120 &    6 &    10 &    15 &    28 &    64 &   155 &  1144 \\
					QoMax-ETC ($q=1/2$) &       40 &    5 &     8 &    11 &    18 &    33 &    64 &   306 \\
					QoMax-ETC ($q=0.9$) &       59 &    5 &     8 &    12 &    20 &    40 &    93 &   702 \\
					ExtremeETC          &      267 &    6 &    14 &    24 &    47 &   108 &   266 &  2687 \\
					ExtremeHunter       &      232 &    8 &    17 &    28 &    53 &   116 &   305 &  2620 \\
					MaxMedian           &       35 &    0 &     7 &    10 &    17 &    30 &    60 &   306 \\
					ThresholdAscent     &      136 &    7 &    12 &    18 &    33 &    70 &   170 &  1299 \\
					\bottomrule
				\end{tabular}
				 \label{tab::maxima_xp6}
			\end{small}
		\end{center}
\end{table}

\clearpage
\subsection*{Experiment 7}

\vspace{0.6cm}

\begin{table}[h]
	\begin{center}
		\begin{small}
		\caption{Statistics on the distributions of number of pulls of the best arm at $T=5\times 10^4$, Experiment 7.}
			\begin{tabular}{c|c|ccccccc}
				\toprule Algorithm & Average  & 1\% &  10\% &  25\% &  50\% &  75\% &  90\% & 99\% \\
				\midrule
QoMax-SDA ($q=1/2$) &       94 &   85 &    94 &    95 &    95 &    95 &    95 &    95 \\
QoMax-SDA ($q=0.9$) &       97 &   89 &    96 &    97 &    98 &    98 &    98 &    98 \\
QoMax-ETC ($q=1/2$) &       90 &   90 &    90 &    90 &    90 &    90 &    90 &    90 \\
QoMax-ETC ($q=0.9$) &       90 &   90 &    90 &    90 &    90 &    90 &    90 &    90 \\
ExtremeETC          &       55 &    3 &     3 &     3 &    90 &    90 &    90 &    90 \\
ExtremeHunter       &       63 &   13 &    40 &    45 &    53 &    90 &    90 &    90 \\
MaxMedian           &        7 &    0 &     0 &     0 &     0 &     0 &     0 &   100 \\
ThresholdAscent     &       57 &   55 &    56 &    57 &    58 &    58 &    58 &    58 \\
				\bottomrule
			\end{tabular}
			\label{tab::nb_pulls_xp7}
		\end{small}
	\end{center}

	\vspace{0.5cm}
	\begin{center}
		\begin{small}
		\caption{Statistics on the distributions of maxima at $T=5\times 10^4$, Experiment 7. Results divided by $1000$ to improve readability.}
			\begin{tabular}{c|c|ccccccc}
				\toprule Algorithm & Average & 1\% &  10\% &  25\% &  50\% &  75\% &  90\% & 99\% \\
				\midrule
QoMax-SDA ($q=1/2$) &     1393 &   73 &   151 &   257 &   488 &  1090 &  2259 & 13764 \\
QoMax-SDA ($q=0.9$) &     1401 &   79 &   163 &   260 &   524 &  1171 &  2830 & 13839 \\
QoMax-ETC ($q=1/2$) &     1337 &   77 &   154 &   245 &   430 &  1007 &  2664 & 13651 \\
QoMax-ETC ($q=0.9$) &     1459 &   84 &   150 &   251 &   461 &   987 &  2419 & 12654 \\
ExtremeETC          &      957 &    6 &    12 &    30 &   214 &   581 &  1511 &  7422 \\
ExtremeHunter       &      867 &   32 &    85 &   156 &   297 &   666 &  1569 & 10855 \\
MaxMedian           &       76 &    0 &     0 &     0 &     0 &     0 &    15 &  1678 \\
ThresholdAscent     &     1043 &   43 &    94 &   160 &   311 &   667 &  1648 & 10715 \\
				\bottomrule
			\end{tabular}
			 \label{tab::maxima_xp7}
		\end{small}
	\end{center}
\end{table}

\clearpage
\subsection*{Experiment 8}

\begin{table}[h]
	\begin{center}
	\begin{small}
	\caption{Statistics on the distributions of number of pulls of the best arm at $T=5\times 10^4$, Experiment 8.}
		\begin{tabular}{c|c|ccccccc}
							\toprule Algorithm & Average & 1\% &  10\% &  25\% &  50\% &  75\% &  90\% & 99\% \\
			\midrule
QoMax-SDA ($q=1/2$) &       91 &   91 &    91 &    91 &    91 &    91 &    91 &    91 \\
QoMax-SDA ($q=0.9$) &       97 &   97 &    97 &    97 &    97 &    97 &    97 &    97 \\
QoMax-ETC ($q=1/2$) &       82 &   82 &    82 &    82 &    82 &    82 &    82 &    82 \\
QoMax-ETC ($q=0.9$) &       82 &   82 &    82 &    82 &    82 &    82 &    82 &    82 \\
ExtremeETC          &       80 &    3 &    82 &    82 &    82 &    82 &    82 &    82 \\
ExtremeHunter       &       82 &   80 &    82 &    82 &    82 &    82 &    82 &    82 \\
MaxMedian           &       31 &    0 &     0 &     0 &     0 &    89 &   100 &   100 \\
ThresholdAscent     &       44 &   44 &    44 &    44 &    44 &    44 &    44 &    44 \\
			\bottomrule
		\end{tabular}
		\label{tab::nb_pulls_xp8}
	\end{small}
\end{center}

\vspace{0.5cm}

\begin{center}
	\begin{small}
	\caption{Statistics on the distributions of maxima at $T=5\times 10^4$, Experiment 8. Results divided by $100$ to improve readability.}
		\begin{tabular}{c|c|ccccccc}
			\toprule Algorithm & Average & 1\% &  10\% &  25\% &  50\% &  75\% &  90\% & 99\% \\
			\midrule
QoMax-SDA ($q=1/2$) &       30 &   14 &    17 &    21 &    27 &    35 &    46 &    75 \\
QoMax-SDA ($q=0.9$) &       31 &   14 &    18 &    21 &    27 &    34 &    46 &    94 \\
QoMax-ETC ($q=1/2$) &       29 &   13 &    17 &    20 &    26 &    34 &    45 &    76 \\
QoMax-ETC ($q=0.9$) &       29 &   14 &    17 &    20 &    26 &    35 &    45 &    88 \\
ExtremeETC          &       28 &    4 &    17 &    20 &    25 &    33 &    43 &    78 \\
ExtremeHunter       &       29 &   13 &    17 &    20 &    25 &    34 &    45 &    78 \\
MaxMedian           &       11 &    0 &     0 &     0 &     0 &    21 &    33 &    65 \\
ThresholdAscent     &       24 &   10 &    14 &    16 &    21 &    28 &    38 &    74 \\
			\bottomrule
		\end{tabular}
		 \label{tab::maxima_xp8}
	\end{small}
\end{center}
\end{table}

\clearpage

%
%
%

\end{document}